\documentclass[runningheads]{llncs}

\usepackage{amsmath,amsfonts,bm}

\def\eqref#1{equation~\ref{#1}}

\def\1{\bm{1}}

\DeclareMathAlphabet{\mathsfit}{\encodingdefault}{\sfdefault}{m}{sl}
\SetMathAlphabet{\mathsfit}{bold}{\encodingdefault}{\sfdefault}{bx}{n}

\newcommand{\Inf}{\text{Inf}}

\usepackage[T1]{fontenc}
\usepackage{array}
\usepackage{graphicx}
\usepackage{graphbox}
\usepackage{wrapfig}
\usepackage{enumerate}
\usepackage{bbm}

\usepackage{tabularx}
\newcolumntype{C}{>{\centering\arraybackslash}X}

\usepackage{xcolor}

\usepackage{algorithm}
\usepackage{algorithmicx}
\usepackage{algpseudocode}
\usepackage{booktabs}
\usepackage{mathtools}
\usepackage{subcaption}
\usepackage{cite} 

\usepackage{hyperref}
\usepackage{url}
\usepackage[capitalize]{cleveref}

\protected\def\subs#1{\ifmmode \text{\tiny #1} \else \textsubscript{\tiny #1}\fi}  
\newcommand{\norm}[1]{\left\lVert#1\right\rVert}

\begin{document}

\title{Leveraging Influence Functions for Resampling Data in Physics-Informed Neural Networks}
%
%
\author{
  Jonas R. Naujoks\inst{1}\textsuperscript{,$\ast$} \and
  Aleksander Krasowski\inst{1}\textsuperscript{,$\ast$}  \and
  Moritz Weckbecker\inst{1} \and\\
  Galip Ümit Yolcu\inst{1} \and
  Thomas Wiegand\inst{1,2,3} \and
  Sebastian Lapuschkin\inst{1,4}\textsuperscript{,$\dagger$} \and \\
  Wojciech Samek\inst{1,2,3}\textsuperscript{,$\dagger$} \and
  René P. Klausen\inst{1}\textsuperscript{,$\dagger$}
}
\titlerunning{Leveraging Influence Functions for Resampling Data in PINNs}
\authorrunning{Naujoks \& Krasowski et al.}
%

\institute{
  Department of Artificial Intelligence, Fraunhofer Heinrich Hertz Institute
  \and
  Department of Electrical Engineering and Computer Science, Technische Universität Berlin
  \and
  BIFOLD -- Berlin Institute for the Foundations of Learning and Data
  \and
  Centre of eXplainable Artificial Intelligence, Technological University Dublin
}

\maketitle
\let\oldthefootnote\thefootnote
\let\thefootnote\relax
\footnotetext{\itshape This article was presented at "The 3rd World Conference on eXplainable Artificial Intelligence" (2025).}
\let\thefootnote\oldthefootnote       

\begin{center}
\small
\textsuperscript{$\ast$} Equal contribution.\\
\textsuperscript{$\dagger$} corresponding: \texttt{\{rene.pascal.klausen, wojciech.samek, sebastian.lapuschkin\}@hhi.fraunhofer.de}
\end{center}
\begin{abstract}

Physics-informed neural networks (PINNs) offer a powerful approach to solving partial differential equations (PDEs), which are ubiquitous in the quantitative sciences. Applied to both forward and inverse problems across various scientific domains, PINNs have recently emerged as a valuable tool in the field of scientific machine learning. A key aspect of their training is that the data --- spatio-temporal points sampled from the PDE’s input domain --- are readily available. Influence functions, a tool from the field of explainable AI (XAI), approximate the effect of individual training points on the model, enhancing interpretability. In the present work, we explore the application of influence function-based sampling approaches for the training data. Our results indicate that such targeted resampling based on data attribution methods has the potential to enhance prediction accuracy in physics-informed neural networks, demonstrating a practical application of an XAI method in PINN training.

\keywords{Physics-Informed Neural Networks \and Data Attribution \and Adaptive Resampling \and Explainable Artificial Intelligence}
\end{abstract}

\section{Introduction}
\setcounter{footnote}{0} 
The success of machine learning, particularly deep learning approaches, in recent years has led to significant breakthroughs and rapid growth in both research and practical applications across diverse fields. In the natural sciences, deep learning applications come with the promise of providing efficient and performant solutions to otherwise classically intractable problems, such as protein structure prediction \cite{abramson_accurate_2024} or molecular dynamics \cite{schutt_schnet_2018}. 
Within the broader context of scientific machine learning, physics-informed neural networks (PINNs) \cite{raissi_physics-informed_2019} recently emerged as a promising avenue of research aimed at modeling partial differential equations (PDEs) by embedding physical constraints directly into neural network training. Here, PINNs incorporate prior domain knowledge through the use of PDE constraints. To date, they have been applied to a plethora of forward and inverse problems across diverse fields such as electromagnetics \cite{beltran-pulido_physics-informed_2022}, fluid mechanics \cite{sharma_review_2023}, medical imaging \cite{sahli_costabal_physics-informed_2020}, and geophysics \cite{rasht-behesht_physics-informed_2022}.
However, despite their potential, these methods often suffer from optimization challenges due to the composite and intricate loss functions inherent to PINNs.
This encompasses gradient pathologies \cite{wang_understanding_2020}, ill-conditioned loss terms \cite{wang_when_2022,rathore_challenges_2024,liu_config_2025}, pathological local minima \cite{rohrhofer_role_2023}, and irregularities in information propagation \cite{daw_mitigating_2023,wu_propinn_2025}, all leading to suboptimal performance. 
The literature proposes several ways to tackle these issues, such as improving the optimization algorithm, architectural modifications, loss function curation, and training data refinement \cite{hao_physics-informed_2023,wang_experts_2023}.

In this work, we approach training data refinement through the lens of explainable artificial intelligence (XAI).
Specifically, we leverage techniques originating from the field of training data attribution (TDA) to guide training data selection for PINNs. TDA quantifies the influence of individual training points by answering crucial questions such as ``Which training data points are particularly influential for a given prediction?'' or ``How would the model parameters change if a specific data point were added to or removed from the training set?'' Such methods have been widely applied in computer vision and large language models, proving useful for model debugging, identifying memorization \cite{pruthi_estimating_2020}, reweighting noisy data \cite{braun_influence_2022}, identifying vulnerabilities to adversarial attacks \cite{koh_understanding_2017}, and, in general, understanding model behavior through the lens of training data \cite{hammoudeh_training_2024}. 

Given their ability to assess the importance of individual training points, TDA methods naturally extend to studying PINNs, where training data consists of coordinate points sampled from the PDE’s input domain.
Notably, PINNs offer a straightforward setting for applying TDA methods, as their training data can be efficiently generated.
While standard PINN training typically makes use of random or structured grid-based sampling, there is an emerging body of research showing how adaptive sampling can substantially improve PINN training efficiency and solution accuracy \cite{liu_grad-rar_2022,florido_investigating_2024,gao_active_2023,wu_comprehensive_2023,nabian_efficient_2021,lau_pinnacle_2024}. 
Building on this, we investigate the promise of using XAI-based TDA techniques, specifically PINN\-fluence \cite{naujoks_pinnfluence_2024}, to adaptively select PINN training points.
More precisely, we explore the hypothesis that training points identified as \textit{influential} via TDA methods provide informative supervision, thus improving the performance of PINNs. 
\subsubsection{Our Contributions}
can be summarized as follows:
\begin{itemize}
    \item We propose a novel XAI-driven adaptive resampling approach to enhance PINN training.
    \item We empirically analyze and compare our PINN\-fluence-based resampling approach with alternative adaptive sampling strategies.
    \item We demonstrate that TDA approaches succeed in improving performance on par with state-of-the-art methods, highlighting their potential in PINN settings.
\end{itemize}

\section{Theoretical Background and Methods} \label{sec:theoretical_bg}
\subsection{Physics-Informed Neural Networks}
PINNs \cite{raissi_physics-informed_2019} are a widely used machine learning approach for solving partial differential equations. 
Let $\Omega\subset \mathbb R^n$ be an open and bounded domain, $\partial \Omega$ denote its boundary and $\overline \Omega = \Omega \cup \partial \Omega$ correspond to its closure. We consider the initial boundary value problem, defined by
\begin{align}
& \mathcal{N}[u](\boldsymbol{x}) = 0, \quad \boldsymbol x \in \Omega \label{eq:IBVPpde}\\
& \mathcal{B}_k[u](\boldsymbol{x}) = 0, \quad \boldsymbol x \in \Gamma_k \subseteq \partial \Omega \text{ for } k=1,...,K \label{eq:bc} \quad\text{.}
\end{align}
where $\mathcal{N}$ and $\mathcal{B}_k$ denote differential operators acting on the solution  $u:\overline \Omega \to \mathbb R^d$ of \cref{eq:IBVPpde,eq:bc} with $K$ initial or boundary conditions; $\boldsymbol{x}$ encompasses spatio-temporal coordinates.\footnote{Depending on the PDE, these coordinates may be purely spatial, purely temporal, both, or even any other physical quantity the PDE depends on.}
\cref{eq:bc} is formulated to include, among other types, both Dirichlet and Neumann boundary conditions.

PINNs approximate a solution $u$ of \cref{eq:IBVPpde,eq:bc} with a neural network $\phi(\boldsymbol{x}; \theta)$ with trainable network parameters $\theta \in \Theta$. The training process minimizes the empirical risk, defined by a composite loss function
\begin{align*}
    & \mathcal L (\theta) = \frac{1}{N_\text{pde}} \sum_{x\in\mathcal X_\text{pde}} L_{\text{pde}} (x;\theta) + \frac{1}{N_\text{bc}} \sum_{x\in\mathcal X_\text{bc}} L_{\text{bc}}(x;\theta) \\
    &\text{with}\ L_\text{pde}(\boldsymbol{x}; \theta) := \mathbbm 1_\Omega(\boldsymbol{x}) \norm{\mathcal N[\phi(\boldsymbol x;\theta)]}_2^2 \\
    &\text{and}\ L_\text{bc}(\boldsymbol{x}; \theta) := \sum_{k=1}^K \mathbbm 1_{\Gamma_k}(\boldsymbol x) \norm{\mathcal B_k[\phi(\boldsymbol{x};\theta)]}_2^2\quad\text{,}
\end{align*}
where $\mathbbm 1_A(\boldsymbol x) = 1$ for $x\in A$ and $0$ otherwise. 
A further data-driven regression term can be included in the loss function, but is omitted here for clarity. For the training loss, we consider in general $N_\text{pde}$ collocation points $\mathcal X_\text{pde}\subset\Omega$ sampled from the domain and $N_\text{bc}$ points $\mathcal X_\text{bc}\subseteq\partial\Omega$ from its boundary, which together constitute the training set $\mathcal{X}_{\text{train}}$.

This study focuses on the resampling of these spatio-temporal training points and proposes an influence function-based method for adaptive selection during training. In all of our experiments, we enforce boundary and initial conditions through the use of hard constraints \cite{lu_physics-informed_2021}. As a result, the loss $\mathcal{L}$ reduces to $\mathcal L_\text{pde}$. Consequently, the training data is sampled from $\Omega$ only.

\subsection{PINNfluence} Introduced in \cite{naujoks_pinnfluence_2024}, PINN\-fluence uses influence functions \cite{koh_understanding_2017} for PINNs to measure the effect of individual training points on the model's behavior to validate its prediction. This is achieved by approximating how the loss (or any other function of the model output) would change if a specific training point were to be added to or removed from the training set and the model were retrained. The effect of adding a point $\boldsymbol{x}^+$ to the training dataset on the total test loss can be approximated as \cite{naujoks_pinnfluence_2024}
\begin{equation}
    \Inf_{\mathcal{L}_\text{test}(\hat{\theta})}(\boldsymbol{x}^+)\coloneqq \nabla_{\theta}\mathcal{L}_\text{test}(\hat{\theta})\!^\top\mathcal{H}_{\hat{\theta}}^{-1}\ \nabla_{\theta}L(\boldsymbol{x}^+;\hat{\theta}),
    \label{eq:pinnfl-full}
\end{equation}
where $\hat{\theta}$ denotes the model parameters optimized on the training loss, $\mathcal{L}_\text{test}(\hat{\theta}) = \frac{1}{N_\text{test}} \sum_{\boldsymbol{x} \in \mathcal{X}_{\text{test}}} L(\boldsymbol{x}; \hat{\theta})$ is the total test loss and $\mathcal{H}_{\hat{\theta}} = \frac{1}{N_\text{train}} \nabla_{\theta}^2 \sum_{\boldsymbol{x} \in \mathcal{X}_{\text{train}}} L(\boldsymbol{x}; \hat{\theta})$ is the Hessian of the training loss with respect to $\theta$.
Due to the linearity of \cref{eq:pinnfl-full}, the effect of adding multiple training points is simply given by the sum of individual effects.
Recent work by \cite{wei_final-model-only_2024} recontextualises such training data attributions methods as an approximation of not retraining the model on a disturbed dataset but of continued fine-tuning of the already trained model on this dataset. Therefore, PINN\-fluence can help us assess the counterfactual effect of adding new points to a fine-tuning set.

\subsection{Resampling Approaches for PINNs} 
The training data of PINNs $\mathcal{X}_{\text{train}}$ consists of collocation points sampled from $\overline \Omega$. 
Usually, the data is sampled once before training and kept constant during the optimization procedure. The most straightforward sampling approaches include drawing points from a uniform distribution or an equidistant grid over $\overline\Omega$.
Due to the complex nature of loss functions in PINNs, the distribution of $\mathcal{X}_{\text{train}}$ is paramount for effective training and high performance, and is generally problem-specific.
One way of tackling this problem is adapting or constraining the weights of individual loss terms or individual data points \cite{wang_respecting_2024,liu_config_2025}.
Another effective strategy to improve the training of PINNs is to adaptively resample the collocation points. Resampling strategies such as residual-based adaptive refinement (RAR) \cite{lu_deepxde_2021} or variations of importance-based resampling \cite{wu_comprehensive_2023,florido_investigating_2024,liu_grad-rar_2022}, distribution-based approaches \cite{gao_active_2023,wu_comprehensive_2023,nabian_efficient_2021} and neural tangent kernel-based point selection \cite{lau_pinnacle_2024} have demonstrated their ability to further improve performance. Another line of research has incorporated principles such as temporal causality in the sampling process itself \cite{daw_mitigating_2023}. In essence, these methods can be subsumed as adding or replacing collocation points based on a metric that reflects the PDE residual or gradients at collocation points of interest with the goal of increasing PINN performance.

Our work aligns with the aforementioned body of research: we employ a score-based metric for gauging the importance of potential collocation points. In contrast to these prior approaches, we utilize an XAI-guided metric obtained via PINN\-fluence.

\subsection{PINNfluence-based Resampling}

In this subsection, we introduce our approach. Let $\mathcal{X}_\text{cand}\subset\overline\Omega$ denote a finite set of randomly sampled candidate points and let $\mathcal{S}: \overline{\Omega} \to \mathbb{R}_{\geq 0}$ be a scoring function. The choice of the scoring function will define the different resampling methods.

\subsubsection{Scoring Training Data based on PINN\-fluence}

Recall the influence given by \cref{eq:pinnfl-full}.
This expression gives us a way to estimate the effect of adding a potential candidate training sample $\boldsymbol{x}^+$ on the loss of a representative test set $\mathcal{X}_\text{test} \subseteq\overline{\Omega}$. Our PINN\-fluence-based scoring metric is thus defined by taking the absolute value of the influence function:
\begin{equation}
    \mathcal{S}_{\Inf}(\boldsymbol{x}^+) \coloneq |\Inf_{\mathcal{L}_\text{test}(\hat{\theta})}(\boldsymbol{x}^+)|.
\end{equation}
A positive influence corresponds to a reduction in the loss, whereas a negative influence is associated with an increase.
Hence, candidate training points $\boldsymbol{x}^+$ have large absolute influences when they have a large effect on changing the output of the loss function through the approximated change in model parameters $\hat\theta$. 
Further, our choice of using the absolute value is motivated by the observation that PINNs often become trapped in local minima as stated before. 
By incorporating points that also potentially increase the loss, we introduce perturbations that may help the model break free from undesirable minima. 
Consequently, this formulation effectively prioritizes the most impactful candidates for refinement.

\subsubsection{Distribution-based Resampling}
Scores retrieved through PINN\-fluence are often densely concentrated in a small area and therefore non-representative, which is undesirable if we wish to sample multiple points at once. Motivated by \cite{nabian_efficient_2021}, the authors of \cite{wu_comprehensive_2023} propose to sample from a probability density function proportional to the residual, which in practice can be implemented as a probability mass function.
Given a set of candidate points $\mathcal{X}_\text{cand}$ and associated importance scores $\mathcal{S}(\boldsymbol{x})$ for all $\boldsymbol{x} \in \mathcal{X}_\text{cand}$, one can define a discrete probability mass function over $\mathcal{X}_\text{cand}$ as
\begin{equation}
    p(\boldsymbol{x}) = \frac{\mathcal{S}(\boldsymbol{x})^\alpha + c}{\sum_{\boldsymbol{x}' \in \mathcal{X}_\text{cand}}\left(\mathcal{S}(\boldsymbol{x}')^\alpha + c\right)}
\label{eq:res-pmf}
\end{equation}
for each $\boldsymbol{x} \in \mathcal{X}_\text{cand}$, where $\alpha, c \in \mathbb R_{\geq0}$ are hyperparameters that determine the sharpness and uniformity of the distribution. Increasing the exponent $\alpha$ amplifies the difference between low and high scores, emphasizing high scores.  A larger value of $c$ smooths the distribution towards a uniform one, thus boosting the probability of adding low-scoring points.   
The new training set $\mathcal{X}_{\text{new}}$ is constructed by generating independent samples from $\mathcal{X}_\text{cand}$ according to $p$ without replacement.

\subsection{Resampling Based on Alternative Scoring Functions}
We compare our proposed method with several alternative scoring functions $\mathcal{S}(\boldsymbol{x})$: Residual-based adaptive refinement, a standard PINN approach, an additional TDA method based on the gradient dot product and two derivative-based ones.

\subsubsection{Residual-Based Adaptive Refinement (RAR) \cite{lu_deepxde_2021}}
RAR scores, given by
\begin{equation}
    \mathcal{S}_{\text{RAR}} = \left| \mathcal{N}[\phi(\boldsymbol{x}; \hat\theta)] \right|,
\label{eq:RAR}
\end{equation}
are assigned based on the PDE residual of the PINN output.
This method prioritizes points where the model's prediction deviates the most from satisfying the PDE. Note that in \cite{wu_comprehensive_2023}, different residual-based sampling approaches are referred to as RAR-D (iteratively adding new samples) and RAD (resampling the entire set). In our work, we use both adding and replacing strategies but refer to the method simply as RAR, focusing on the scoring of importance rather than the sampling process.
\subsubsection{Grad-Dot \cite{charpiat_input_2019}}
Grad-Dot is used as an approximation of influence functions by replacing the Hessian with an identity matrix, which yields 
\begin{equation}
\mathcal{S}_{\text{grad-dot}}(\boldsymbol{x}) = \left|\nabla_{\!\theta}\mathcal{L}_\text{test}(\hat{\theta})\!^\top \nabla_{\!\theta}L(\boldsymbol{x};\hat{\theta}) \right|.
    \label{eq:grad-dot}
\end{equation}
\subsubsection{Additional Gradient-Based Methods}
To complete our studies on different scoring functions, we introduce two heuristically motivated gradient-based measures:  
\begin{itemize}
    \item The $L^2$ norm of the prediction gradient with respect to its input:
    \begin{equation}
        \mathcal{S}_{\text{output-grad}}(\boldsymbol{x}) = \left\| \nabla_{\!\boldsymbol x} \phi(\boldsymbol{x}; \hat{\theta})\right\|_2.
    \end{equation}
    In a recent study \cite{florido_investigating_2024}, an alternative scoring function is proposed based on the second derivative of the neural network's output with respect to its inputs. Similarly, we consider the gradients of the network $\nabla_{\!\boldsymbol x} \phi(\boldsymbol{x};\theta)$.
    Given that PINNs approximate the true solution with some accuracy, we hypothesize that sampling points in regions where the function exhibits high sensitivity to input perturbations may identify informative training locations.

    \item The $L^2$ norm of the loss gradient with respect to model parameters:
    \begin{equation}
        \mathcal{S}_{\text{loss-grad}}(\boldsymbol{x}) = \left\| \nabla_{\!\theta} {L}(\boldsymbol{x}; \hat\theta) \right\|_2.
    \end{equation}
    This metric captures how strongly a given training point influences model updates, with higher values indicating points that drive larger parameter adjustments during optimization. This is different from influence function-based scores, which assess the change in the test loss or potentially any other function of the PINN output. 
\end{itemize}
\begin{algorithm}
\caption{Resampling for PINNs}
\label{alg:resampling}
\begin{algorithmic}[1]
    \Require Initial PINN model \( \phi(\boldsymbol{x}; \theta) \), PDE \( \mathcal{N}[u](\boldsymbol{x}) = 0 \), initial training dataset \( \mathcal{X}_{\text{train}} \), number of training epochs $T_\subs{tr} \geq 0$, number of fine-tuning epochs $T_\subs{ft} > 0$, number of candidate points \( N_{\text{cand}} \), number of points to sample \( k \), resampling method, perturbation method (add or replace).

    \State \textbf{Pretrain} \( \phi(\boldsymbol{x}; \theta) \) on \( \mathcal{X}_{\text{train}} \) for $T_\subs{tr}$ epochs (\textbf{optional}). 
    \State \textbf{Sample} candidate collocation points uniformly:  
    \[
    \mathcal{X}_{\text{cand}} \sim \mathcal{U}(\overline\Omega), \quad |\mathcal{X}_{\text{cand}}| = N_{\text{cand}}.
    \]
    
    \State \textbf{Compute} importance scores \( \mathcal{S}(\boldsymbol{x}) \) for all \( \boldsymbol{x} \in \mathcal{X}_{\text{cand}} \).
    \State \textbf{Select} training points by sampling from a distribution over their scores \cref{eq:res-pmf}.
    
    \State \textbf{Perturb} training dataset (\textbf{choose one}): 
    \begin{itemize}
        \item \textbf{Add}: \( \mathcal{X}_{\text{train}} \leftarrow \mathcal{X}_{\text{train}} \cup \mathcal{X}_{\text{new}} \)
        \item \textbf{Replace}: \( \mathcal{X}_{\text{train}} \leftarrow \mathcal{X}_{\text{new}} \).
        \label{alg:perturb}
    \end{itemize}

    \State \textbf{Fine-tune} \( \phi(\boldsymbol{x}; \theta) \) on the updated \( \mathcal{X}_{\text{train}} \) for $T_\subs{ft}$ epochs.
    
    \State \textbf{Repeat} steps 2–6 until a stopping criterion is met.
\end{algorithmic}
\end{algorithm}

\section{Experiments}

\subsection{Model Setup}
To compare the methods introduced in \cref{sec:theoretical_bg}, we evaluate them on five different PDEs, namely the diffusion, Burgers', Allen-Cahn, wave, and drift-diffusion equations.
Their respective mathematical descriptions and selected hyperparameters, as well as the architecture of the used fully connected neural networks, are provided in \cref{app:pdes}. We employ distinct numbers of hidden layers and neurons tailored to each specific problem, using $\operatorname{tanh}$ activations throughout all experiments. All boundary and initial conditions are enforced via hard constraints, enabling sampling exclusively from the open domain $\Omega$. All models are optimized using \texttt{Adam}, with a learning rate of $10^{-3}$,  \cite{kingma_adam_2017} and \texttt{L-BFGS} \cite{liu_limited_1989}, with default settings. To calculate PINN\-fluence \cite{naujoks_pinnfluence_2024}, we utilize a modified \texttt{ArnoldiInfluenceFunction} implementation \cite{schioppa_scaling_2021} from \texttt{captum} \cite{kokhlikyan_captum_2020}.

\subsection{Evaluation Protocol}
We consider a ground truth dataset $\mathcal{D}_{\text{gt}} = \{ (\boldsymbol{x}_m, u(\boldsymbol{x}_m)) \mid m = 1, \dots, N_{\text{gt}} \}$ consisting of input-output pairs. In the case of the diffusion, drift-diffusion, and wave equations, where the analytical solution $u$ is known, we randomly sample $\boldsymbol{x}_m$ uniformly from $\Omega$. In contrast, for the Allen-Cahn and Burgers' equations, the input points $\boldsymbol{x}_m$ are sampled from an equidistant grid over $\Omega$, with $u(\boldsymbol{x}_m)$ corresponding to a numerically obtained solution \cite{wu_comprehensive_2023}. 
To evaluate the accuracy of the PINN solution, we compute the $L^2$ relative error over $\mathcal{D}_{\text{gt}}$:
\begin{equation}
    \sqrt{\frac{\sum_{m=1}^{N_\text{gt}} \left[\phi(\boldsymbol{x}_m; \theta) - u(\boldsymbol{x}_m)\right]^2}{\sum_{m=1}^{N_\text{gt}} u(\boldsymbol{x}_m)^2}} \quad\text{.}
\end{equation}
\subsection{Resampling Strategies}
For each PDE, we conduct two experiments, differing in the perturbation strategy specified in \cref{alg:resampling}: \textbf{Adding} new samples to the training data ($\mathcal{X}_{\text{train}} \leftarrow  \mathcal{X}_{\text{train}} \cup \mathcal{X}_{\text{new}}$) or \textbf{Replacing} ($\mathcal{X}_{\text{train}} \leftarrow  \mathcal{X}_{\text{new}}$) the entire training set with newly sampled data based on the different scoring functions. In order to obtain a comparative baseline, we also include both random resampling as well as no resampling in our experiments. The number of candidate points from which we add or replace is set to $N_{\text{cand}} =10\,000$ across all experiments. For the diffusion equation, we used $|\mathcal{X_{\text{train}}}| = 30$, in both settings, and added $|\mathcal{X_{\text{new}}}| = 1$ in each perturbation step. For all remaining problems, we used $|\mathcal{X_{\text{train}}}| = 1\,000$ and $|\mathcal{X_{\text{new}}}| = 10$. In both variations, we omit the optional pretraining phase; hence the models' parameters are randomly initialized.
The training process runs for a total of $100$ cycles. Each cycle precisely follows steps 2-6 of \cref{alg:resampling}. The fine-tuning step (6) consists of $1\,000$ iterations with \texttt{Adam}, followed by $1\,000$ with \texttt{L-BFGS}.
In the additive setting, $\mathcal{X}_{\text{train}}$ consists of points drawn from a Hammersley sequence \cite{hammersley_monte_1964} and we sample from the distribution (\cref{eq:res-pmf}) with $\alpha=2$ and $c=0$, whereas for the replacement-based setting, we sample with $\alpha=1$ and $c=1$.\footnote{For the choice of hyperparameters, we refer the reader to \cite{wu_comprehensive_2023}.} Each setting is evaluated over 10 runs. 

\section{Results and Discussion}

In this section, we present the main findings of our investigation and provide an accompanying analysis of their implications and limitations.
\subsection{Results}

Our results aim to compare the performance of the various scoring strategies with an emphasis on both PINN\-fluence and RAR.
\cref{tab:add_results,tab:replace_results} summarize the relative errors for the addition- and replacement-based perturbation strategies, respectively. In both settings, for most PDEs, PINN\-fluence was competitive with RAR, whereas other methods tended to lag behind. The corresponding losses are given in \cref{tab:add_losses,tab:replace_losses}.

Additionally, a clear problem dependence could be observed: for Burgers' equation and the wave equation, the random resampling strategy as well as the static strategy resulted in significantly worse final model accuracy compared to scoring-based resampling. In contrast, for the remaining PDEs, even these settings produced reasonable results, although not surpassing RAR or PINN\-fluence. 
Solely for the diffusion and drift-diffusion equations all methods performed similarly, which we hypothesize is likely due to the relative simplicity of these problems.

The alternative scoring strategies Output-Grad, Loss-Grad, and Grad-Dot, achieved mixed performance. Especially Output-Grad produced the worst accuracies, which indicates that this scoring strategy is inadequate. Scoring based on the gradient of the loss function (Loss-Grad), on the other hand, yielded reasonable results, particularly for Burgers' equation in the additive setting. Employing Grad-Dot as a scoring method leads to similar results.

As shown in \cref{tab:add_results}, RAR, with one exception, consistently achieved the best performance in the additive setting, while PINN\-fluence-based scoring closely trails behind. In the case of replacing the training dataset depicted in \cref{tab:replace_results}, PINN\-fluence outperformed the other methods in most cases, showcasing its usefulness. Remarkably, for the drift-diffusion equation, training on an unperturbed dataset (\textit{Static}) yields the best performance, albeit not by a significant margin.

\vspace{-6pt}

\begin{table}[H]
    \centering
    \small
    \caption{Mean relative error ($L^2$) and standard deviations for sampling strategies when adding train points ($\mathcal{X}_{\text{train}} \leftarrow \mathcal{X}_{\text{train}} \cup \mathcal{X}_{\text{new}}$). The smallest error is highlighted in bold. Columns are given in the same order of magnitude (OOM).}
    \label{tab:add_results}
    \begin{tabularx}{\linewidth}{
        l | C | C | C | C | C
    }
        \toprule
         PDE & Diffusion & Allen-Cahn & Burgers' & Wave & Drift-Diffusion \\
        {[OOM]}  & $[10^{-5}]$ & $[10^{-3}]$ & $[10^{-4}]$ & $[10^{-4}]$ & $[10^{-3}]$ \\
        \midrule 
        \midrule
        PINNfluence & $4.3{\scriptstyle \pm 1.5}$ & $6.1{\scriptstyle \pm 2.2}$ & $17.2{\scriptstyle \pm 5.5}$ & $4.2{\scriptstyle \pm 0.7}$ & $3.0{\scriptstyle \pm 0.7}$ \\
        RAR & $\mathbf{2.9{\scriptstyle \pm 0.5}}$ & $\mathbf{2.5{\scriptstyle \pm 1.0}}$ & $\mathbf{6.5{\scriptstyle \pm 3.0}}$ & $\mathbf{3.7{\scriptstyle \pm 0.9}}$ & $2.7{\scriptstyle \pm 0.5}$ \\
        Grad-Dot & $4.5{\scriptstyle \pm 1.0}$ & $13.9{\scriptstyle \pm 8.5}$ & $44.2{\scriptstyle \pm 30.1}$ & $5.3{\scriptstyle \pm 1.1}$ & $\mathbf{2.6{\scriptstyle \pm 0.2}}$ \\
        Out-Grad & $6.3{\scriptstyle \pm 1.3}$ & $11.5{\scriptstyle \pm 3.9}$ & $801{\scriptstyle \pm 1471}$ & $246{\scriptstyle \pm 149}$ & $3.5{\scriptstyle \pm 0.6}$ \\
        Loss-Grad & $5.2{\scriptstyle \pm 0.9}$ & $11.5{\scriptstyle \pm 3.3}$ & $32.2{\scriptstyle \pm 13.6}$ & $5.3{\scriptstyle \pm 1.7}$ & $2.6{\scriptstyle \pm 0.4}$ \\
        Random & $4.7{\scriptstyle \pm 1.5}$ & $7.5{\scriptstyle \pm 2.7}$ & $771{\scriptstyle \pm 649}$ & $723{\scriptstyle \pm 1004}$ & $3.3{\scriptstyle \pm 0.4}$ \\
        Static & $62.5{\scriptstyle \pm 36.7}$ & $7.6{\scriptstyle \pm 3.3}$ & $811{\scriptstyle \pm 475}$ & $1317{\scriptstyle \pm 754}$ & $3.3{\scriptstyle \pm 0.7}$ \\
        \bottomrule
    \end{tabularx}
\end{table}

\vspace{-1cm}

\begin{table}[!th]
    \centering
    \small
    \caption{Mean relative error ($L^2$) and standard deviations for sampling strategies when replacing the whole training set ($\mathcal{X}_{\text{train}} \leftarrow \mathcal{X}_{\text{new}}$). The smallest error is highlighted in bold. Columns are given in the same order of magnitude (OOM).}
    \label{tab:replace_results}
    \begin{tabularx}{\linewidth}{
        l | C | C | C | C | C
    }
        \toprule
         PDE 
         & Diffusion
         & Allen-Cahn 
         & Burgers' 
         & Wave 
         & Drift-Diffusion\\ 
        {[OOM]}  & $[10^{-4}]$ & $[10^{-3}]$ & $[10^{-3}]$ & $[10^{-2}]$ & $[10^{-3}]$ \\
        \midrule 
        \midrule
        PINNfluence & $7.2{\scriptstyle \pm 5.0}$ & $\mathbf{6.4{\scriptstyle \pm 1.9}}$ & $\mathbf{1.2{\scriptstyle \pm 0.4}}$ & $\mathbf{2.0{\scriptstyle \pm 2.1}}$ & $3.9{\scriptstyle \pm 0.7}$ \\
        RAR & $5.8{\scriptstyle \pm 5.3}$ & $6.7{\scriptstyle \pm 6.5}$ & $1.6{\scriptstyle \pm 0.5}$ & $27.6{\scriptstyle \pm 39.7}$ & $4.2{\scriptstyle \pm 0.9}$ \\
        Grad-Dot & $8.6{\scriptstyle \pm 12.5}$ & $9.9{\scriptstyle \pm 2.3}$ & $1.8{\scriptstyle \pm 0.7}$ & $34.2{\scriptstyle \pm 26.4}$ & $4.5{\scriptstyle \pm 1.5}$ \\
        Out-Grad & $\mathbf{3.1{\scriptstyle \pm 1.1}}$ & $219{\scriptstyle \pm 217}$ & $2.4{\scriptstyle \pm 1.2}$ & $58.2{\scriptstyle \pm 32.1}$ & $5.3{\scriptstyle \pm 1.5}$ \\
        Loss-Grad & $4.0{\scriptstyle \pm 4.1}$ & $36.6{\scriptstyle \pm 85.9}$ & $2.3{\scriptstyle \pm 2.1}$ & $28.8{\scriptstyle \pm 27.8}$ & $4.8{\scriptstyle \pm 0.9}$ \\
        Random & $3.5{\scriptstyle \pm 2.3}$ & $49.2{\scriptstyle \pm 81.9}$ & $197{\scriptstyle \pm 128}$ & $78.3{\scriptstyle \pm 25.7}$ & $4.6{\scriptstyle \pm 1.5}$ \\
        Static & $6.2{\scriptstyle \pm 3.7}$ & $7.6{\scriptstyle \pm 3.3}$ & $81.1{\scriptstyle \pm 47.5}$ & $13.2{\scriptstyle \pm 7.5}$ & $\mathbf{3.3{\scriptstyle \pm 0.7}}$ \\
        \bottomrule
    \end{tabularx}
\end{table}

\begin{figure}[h!]
    \centering
    \begin{subfigure}[b]{0.9\textwidth}
        \centering
        {\includegraphics[trim={0.42cm 0.5cm 0.4cm 0.5cm},clip,width=\linewidth]{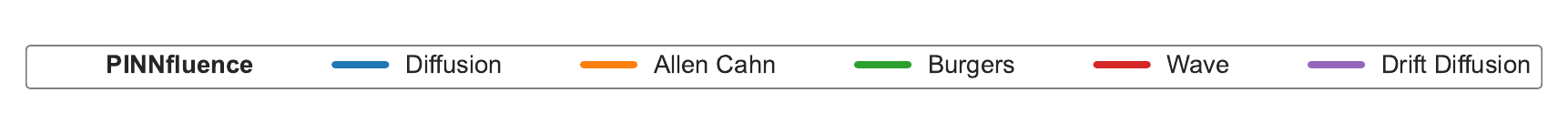}}
    \end{subfigure}
    \begin{subfigure}[b]{0.9\textwidth}
        \centering
        {\includegraphics[trim={0.42cm 0.5cm 0.4cm 0.5cm},clip,width=\linewidth]{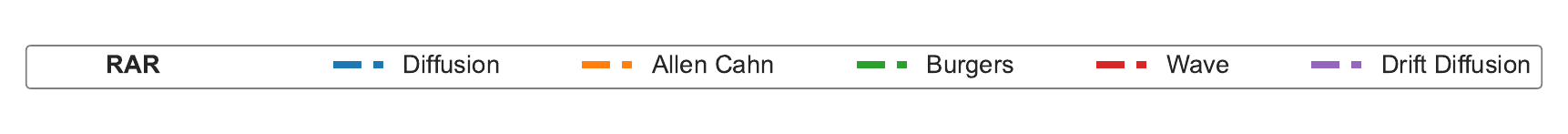}}
    \end{subfigure}
    
    \begin{subfigure}[b]{0.48\textwidth}
        {\includegraphics[width=\linewidth,height=\linewidth]{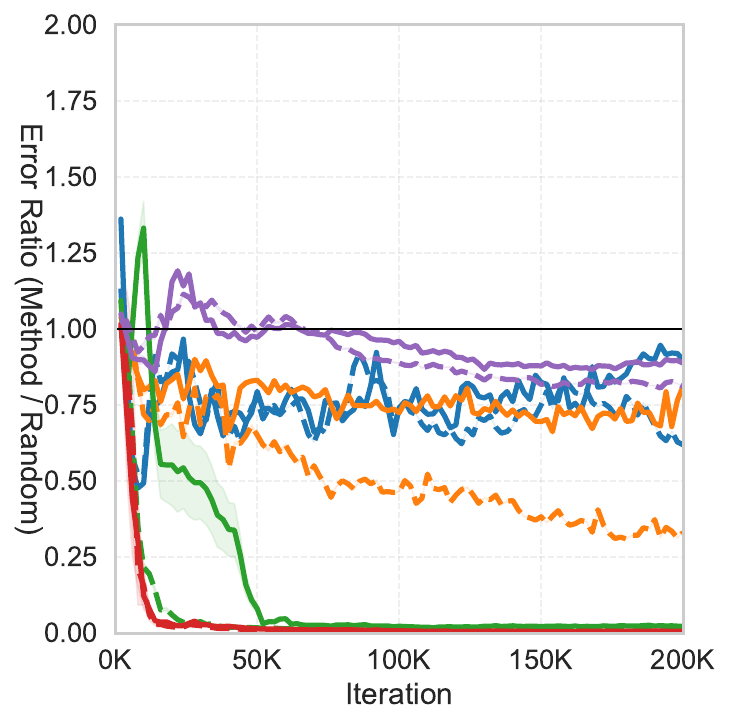}}
        \caption{Adding points: $\mathcal{X}_{\text{train}} \leftarrow  \mathcal{X}_{\text{train}} \cup \mathcal{X}_{\text{new}}$}
    \end{subfigure}
    \hfill 
    \vspace{.01cm}
    \begin{subfigure}[b]{0.48\textwidth}
        {\includegraphics[width=\linewidth,height=1.015\linewidth]{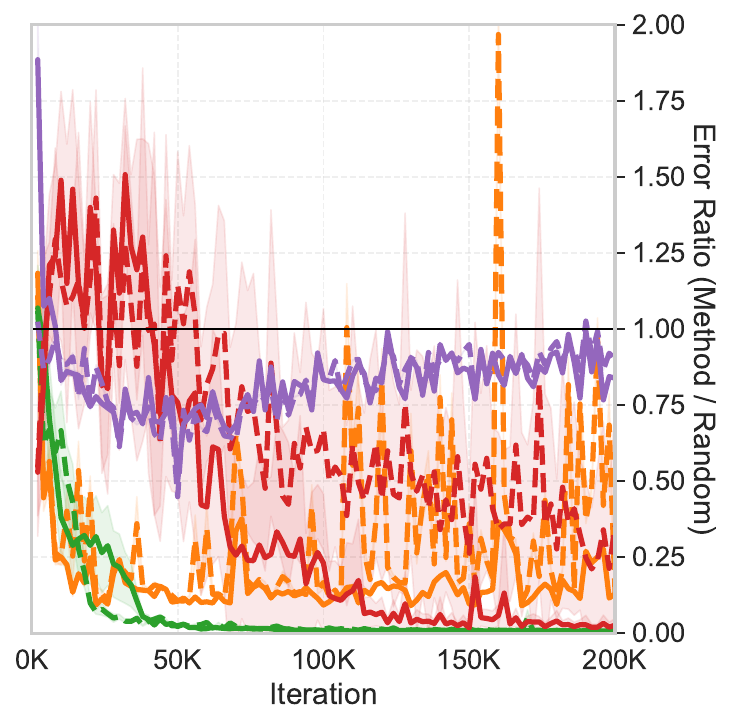}}
        \caption{Replacing points: $\mathcal{X}_{\text{train}} \leftarrow  \mathcal{X}_{\text{new}}$}
        \label{fig:pinnfluence_vs_rar_replace}
    \end{subfigure}

    \caption[Performance comparison between PINN\-fluence]{Performance comparison between PINN\-fluence (\textit{solid lines}) and RAR (\textit{dashed lines}) resampling strategies across five PDEs (\textit{different colors}), with results normalized to uniform random sampling. Lines represent the ratio of $L^2$ relative error (method to random sampling) averaged over 10 runs, with shaded regions indicating standard deviation. Lower values indicate better performance relative to random sampling.\footnotemark}
    
    \label{fig:pinnfluence_vs_rar}
\end{figure}

To enable a more fine-grained comparison of the two top-performing scoring methods, \cref{fig:pinnfluence_vs_rar} displays the training performance of PINN\-fluence and RAR. Here, the error ratio relative to the random sampling baseline is shown to provide a clear perspective on the effectiveness of each method. We omit the other methods here for clarity, as they did not perform on par. 
Across all studied PDEs, both PINN\-fluence and RAR displayed similar performance -- even during training -- highlighting the usefulness of our XAI-driven approach. The biggest discrepancies in the addition-based scenario were given by the Allen-Cahn and Burgers' equations, where RAR achieved superior performance. 

When replacing points (\cref{fig:pinnfluence_vs_rar_replace}), the largest discrepancy is apparent in the wave equation, where PINN\-fluence surpassed RAR. Here, it should be noted that in this setting, the training performance across PDEs was noisier and less stable. This can be attributed to the close performance of the random sampling baseline and the intrinsic variability of the replacement-based setting.

\footnotetext{Note that for \cref{fig:pinnfluence_vs_rar_replace}, we have omitted the diffusion equation, as the methods' performance closely matched the random sampling baseline and the resulting plot lacked clarity.}

This trend is further illustrated in \cref{fig:results-l2_all_pdes_add,fig:results-l2_all_pdes_replace}, which depict the $L^2$ relative errors for all approaches. In the replacement-based setting, for all PDEs with the exception of Burgers' equation, the baseline matched the performance of the other methods closer.
The figure also provides a more detailed comparison for each problem, depicting all sampling methods.

\def\mysfigwidth{.405\textwidth}
\def\myfigheight{.18\textheight}
\def\mysubfigskip{\vspace{-6pt}}

\begin{figure}[h!]
    \centering
    \begin{subfigure}{0.9\textwidth}
        \centering 
        \includegraphics[trim={0 0 0 0},clip,width=\linewidth]{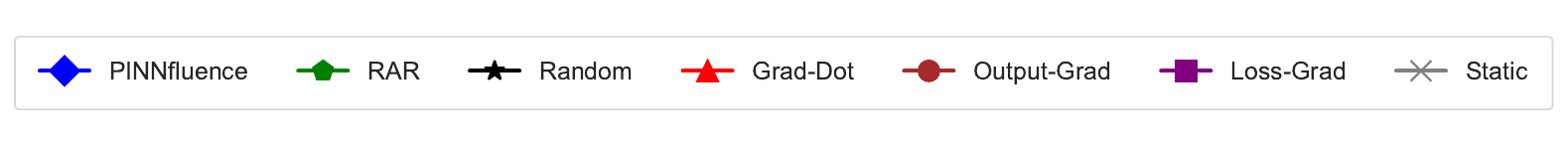}
    \end{subfigure}

    \begin{subfigure}{\mysfigwidth}
        \centering
        \includegraphics[width=\linewidth]{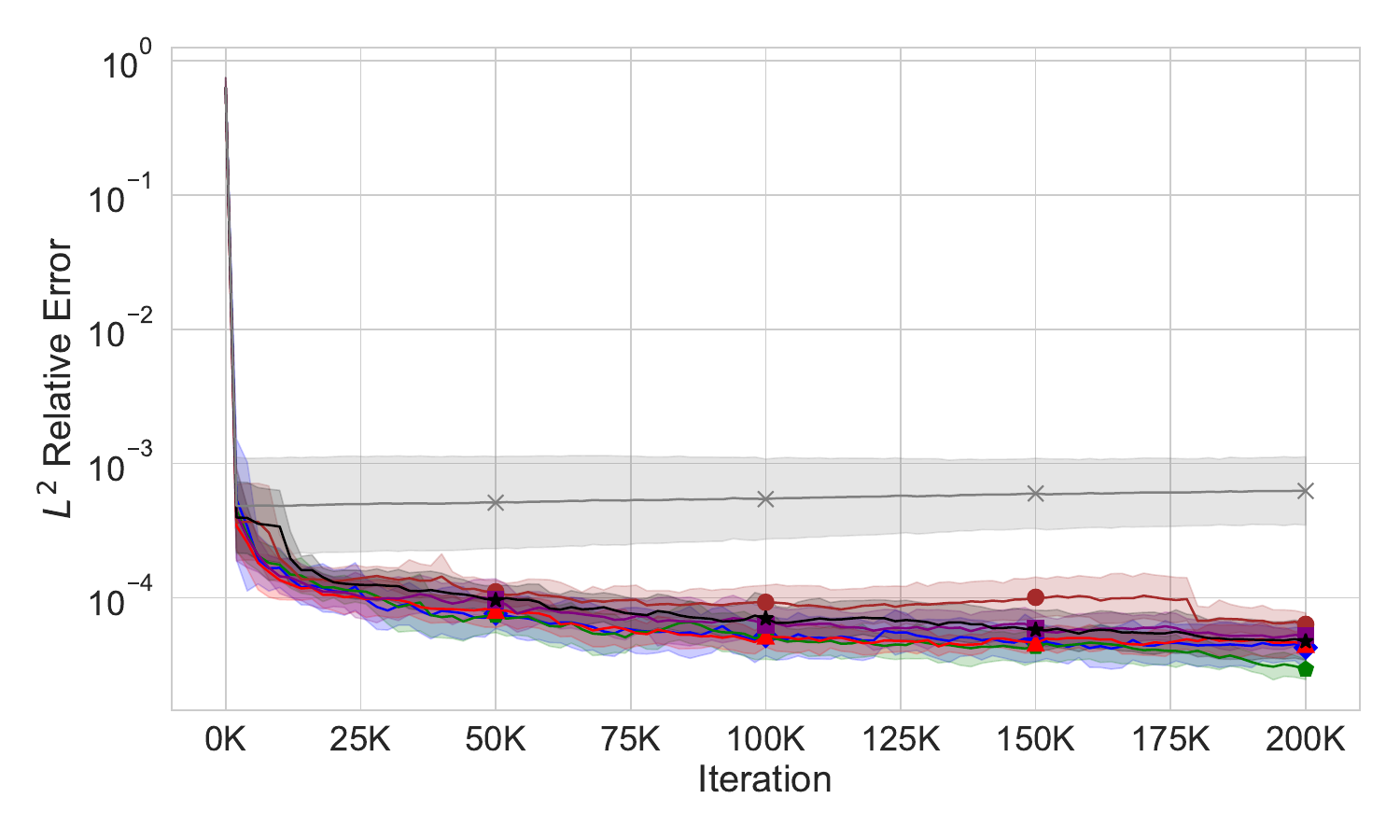}\mysubfigskip
        \caption{Diffusion}
    \end{subfigure}
    \begin{subfigure}{\mysfigwidth}\mysubfigskip
        \centering
        \includegraphics[width=\linewidth]{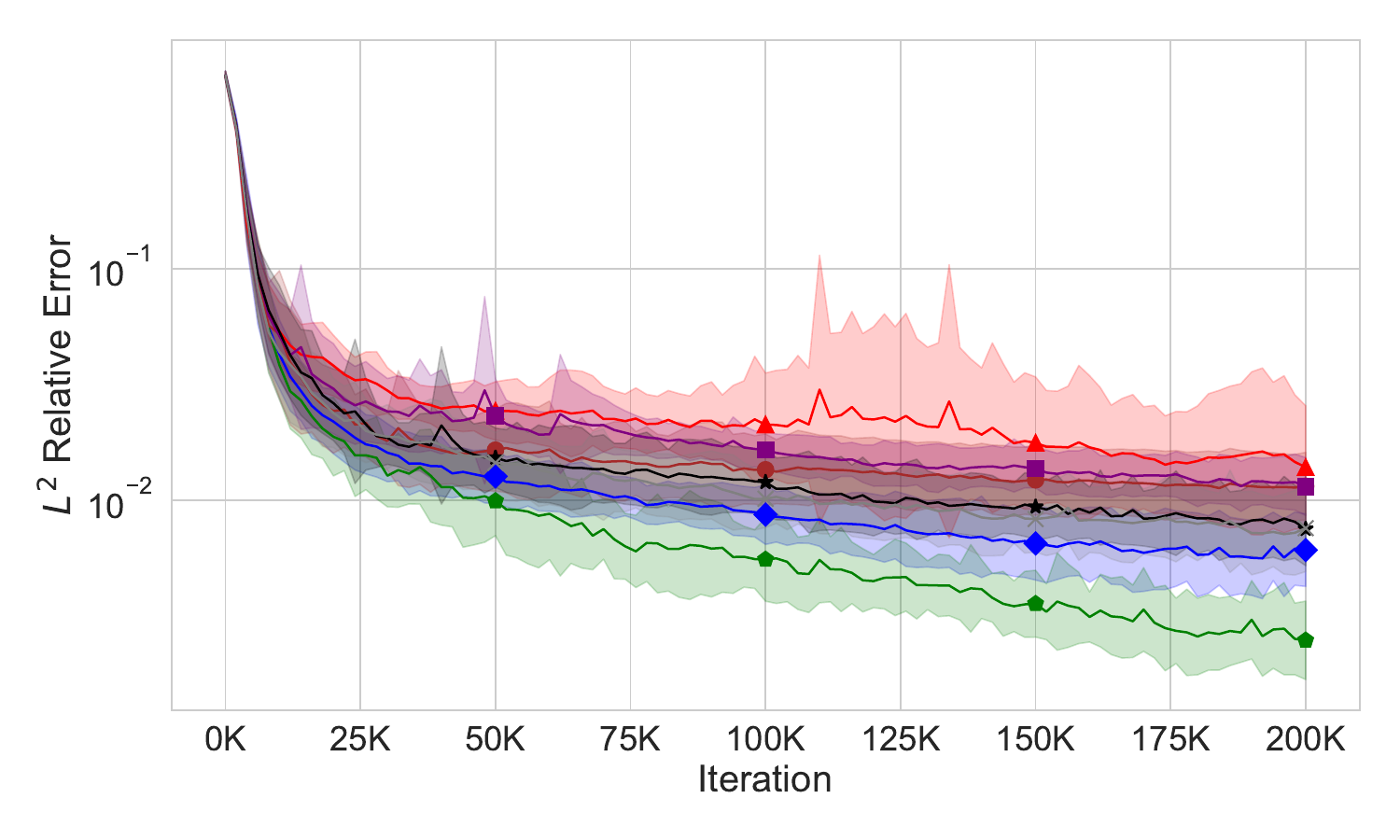}\mysubfigskip
        \caption{Allen-Cahn}
    \end{subfigure}
    
    \begin{subfigure}{\mysfigwidth}
        \centering
        \includegraphics[width=\linewidth]{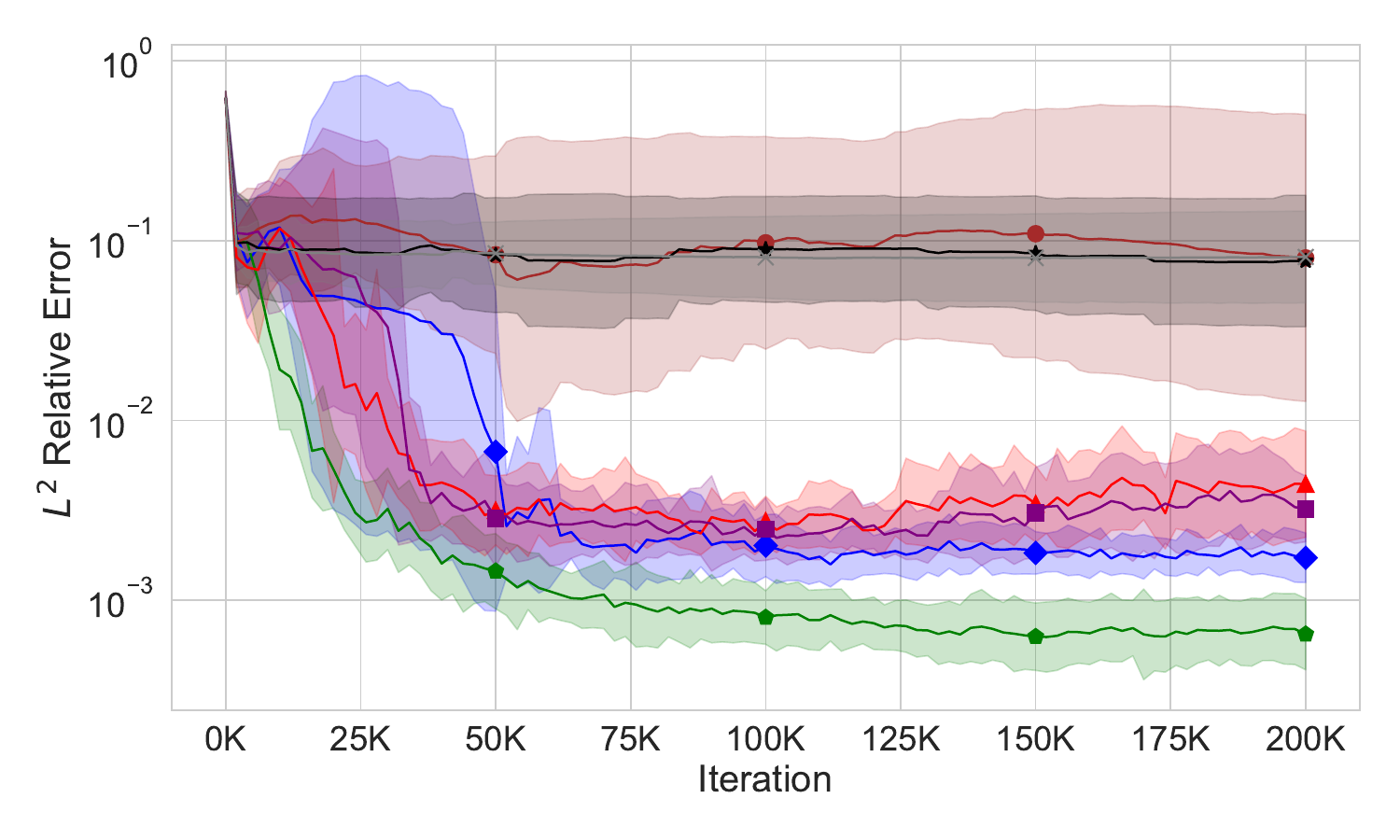}\mysubfigskip
        \caption{Burgers'}
    \end{subfigure} 
    \begin{subfigure}{\mysfigwidth}\mysubfigskip
        \centering
        \includegraphics[width=\linewidth]{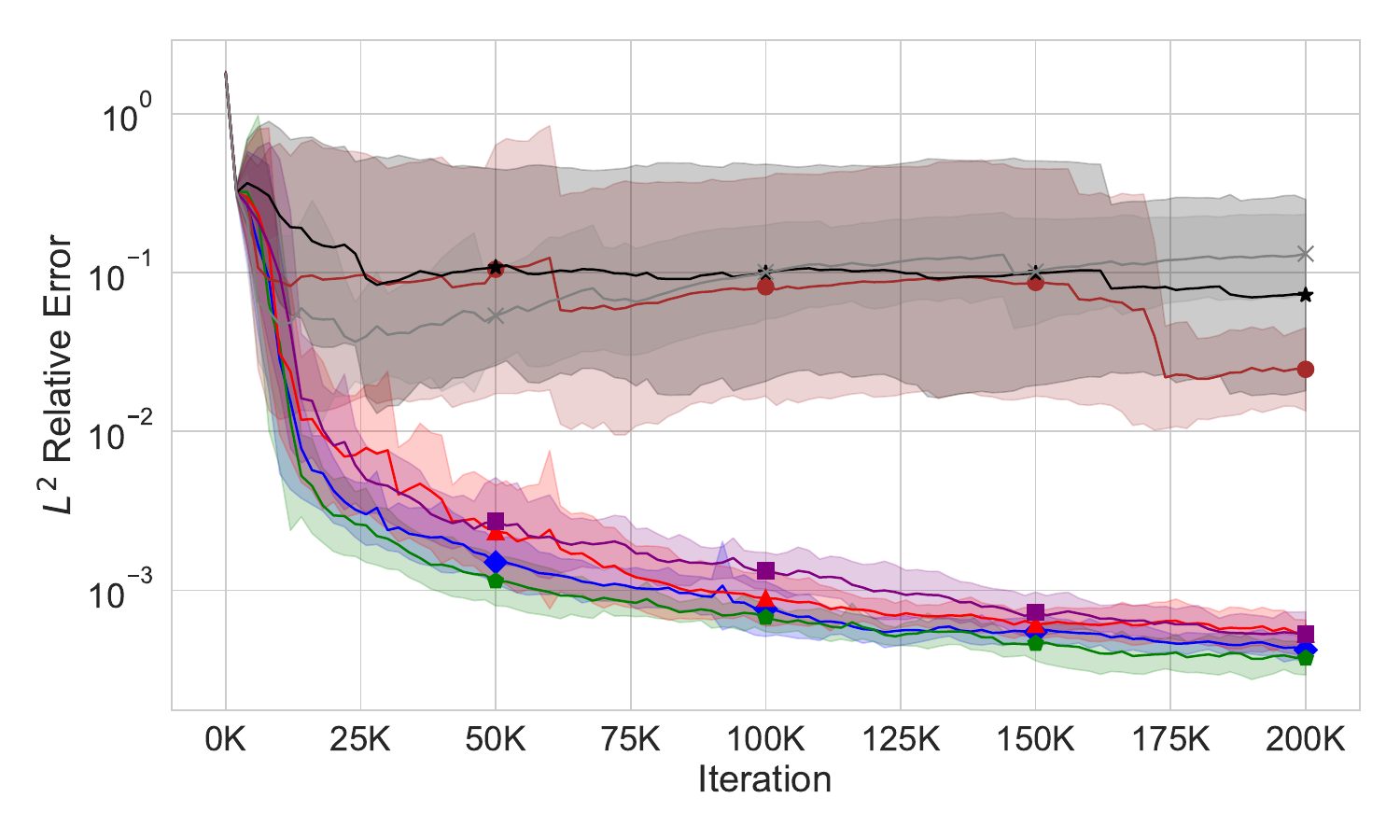}\mysubfigskip
        \caption{Wave}
    \end{subfigure}
    
    \begin{subfigure}{\mysfigwidth}
        \centering
        \includegraphics[width=\linewidth]{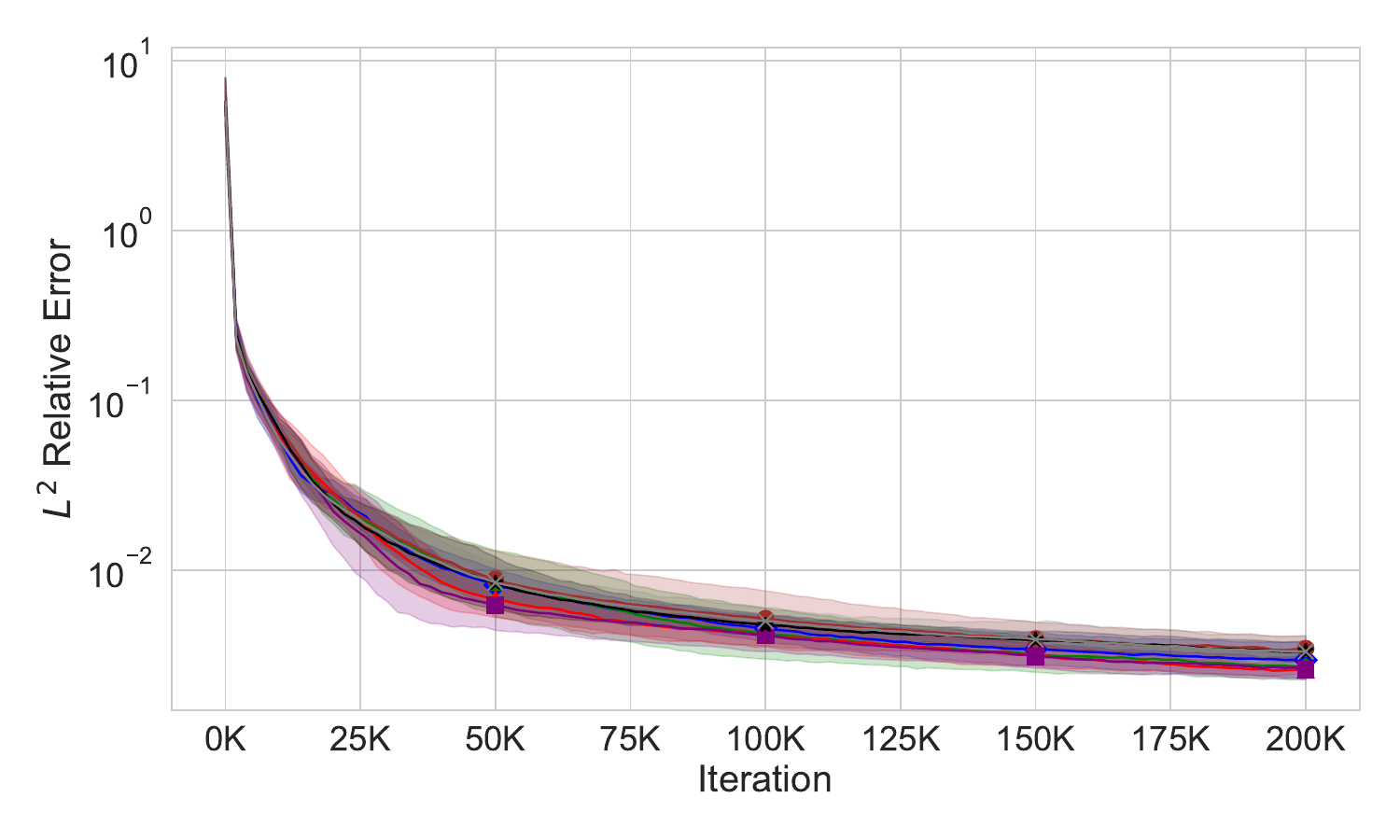}\mysubfigskip
        \caption{Drift-Diffusion}
    \end{subfigure}

    \caption{Comparison of different scoring functions for resampling on five PDEs for iteratively \textbf{adding} new data points. Lines represent $L^2$ relative error averaged over 10 runs, with each point representing the model state after one cycle of \texttt{Adam+LBFGS}, and shaded regions indicating log-scale standard deviations.
    }
    \label{fig:results-l2_all_pdes_add}
\end{figure}

\def\mysfigwidth{.405\textwidth}
\begin{figure}[h!]
    \centering
    \begin{subfigure}{0.9\textwidth}
        \centering 
        \includegraphics[trim={0 0 0 0},clip,width=\linewidth]{figures/lbfgs/replace/legend_horizontal_seperate_plots.pdf}
    \end{subfigure}

    \begin{subfigure}{\mysfigwidth}
        \centering
        \includegraphics[width=\linewidth]{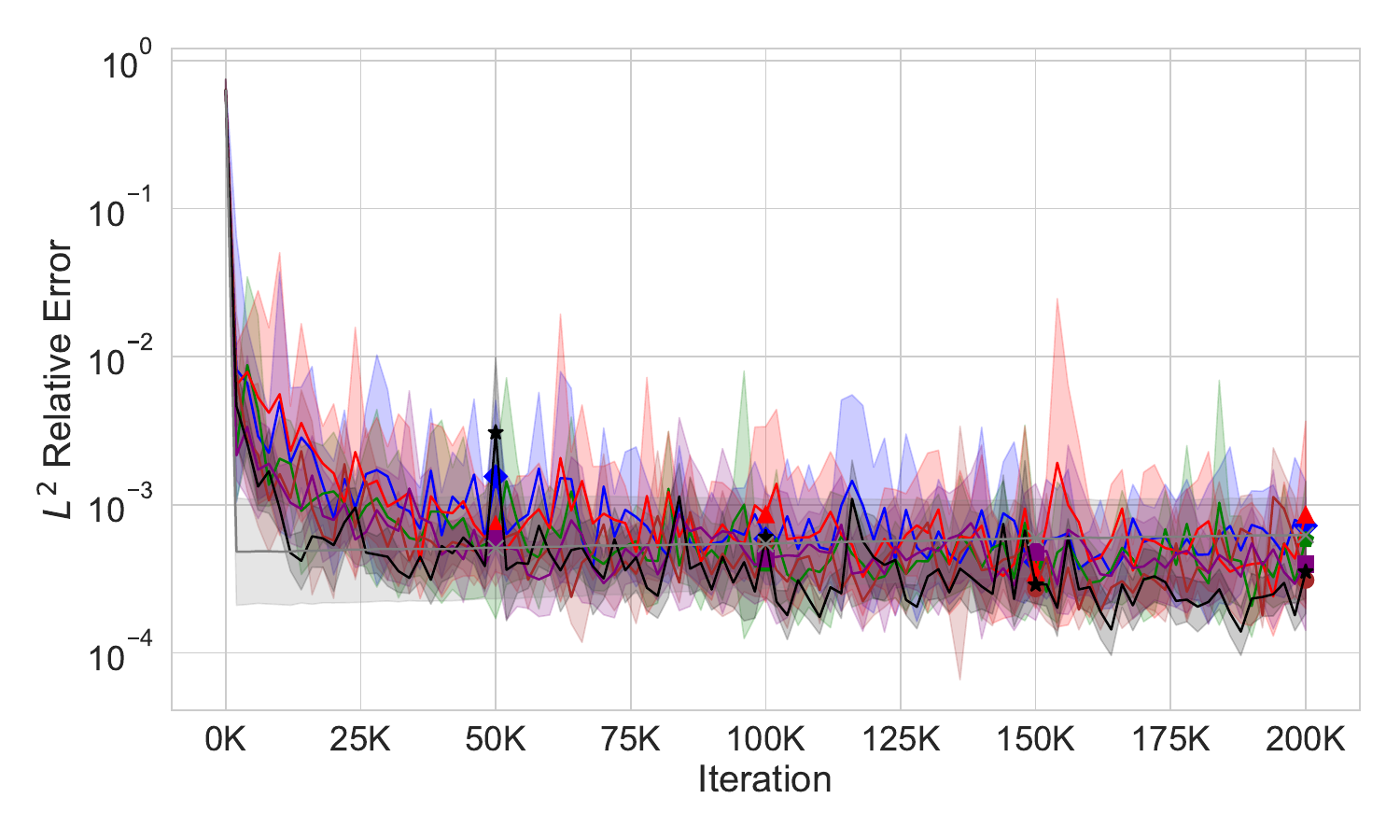}\mysubfigskip
        \caption{Diffusion}
    \end{subfigure}
    \begin{subfigure}{\mysfigwidth}\mysubfigskip
        \centering
        \includegraphics[width=\linewidth]{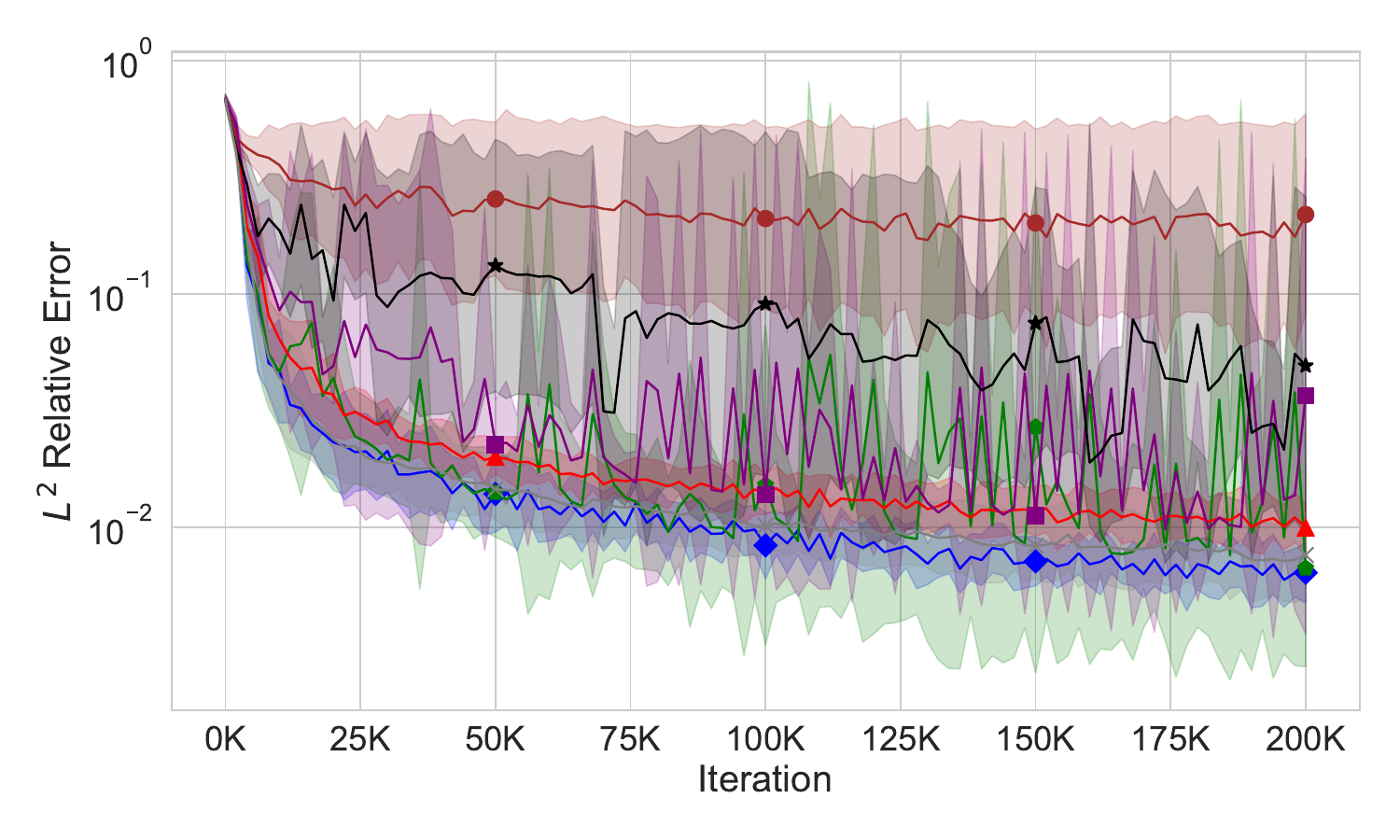}\mysubfigskip
        \caption{Allen-Cahn}
    \end{subfigure}
    
    \begin{subfigure}{\mysfigwidth}
        \centering
        \includegraphics[width=\linewidth]{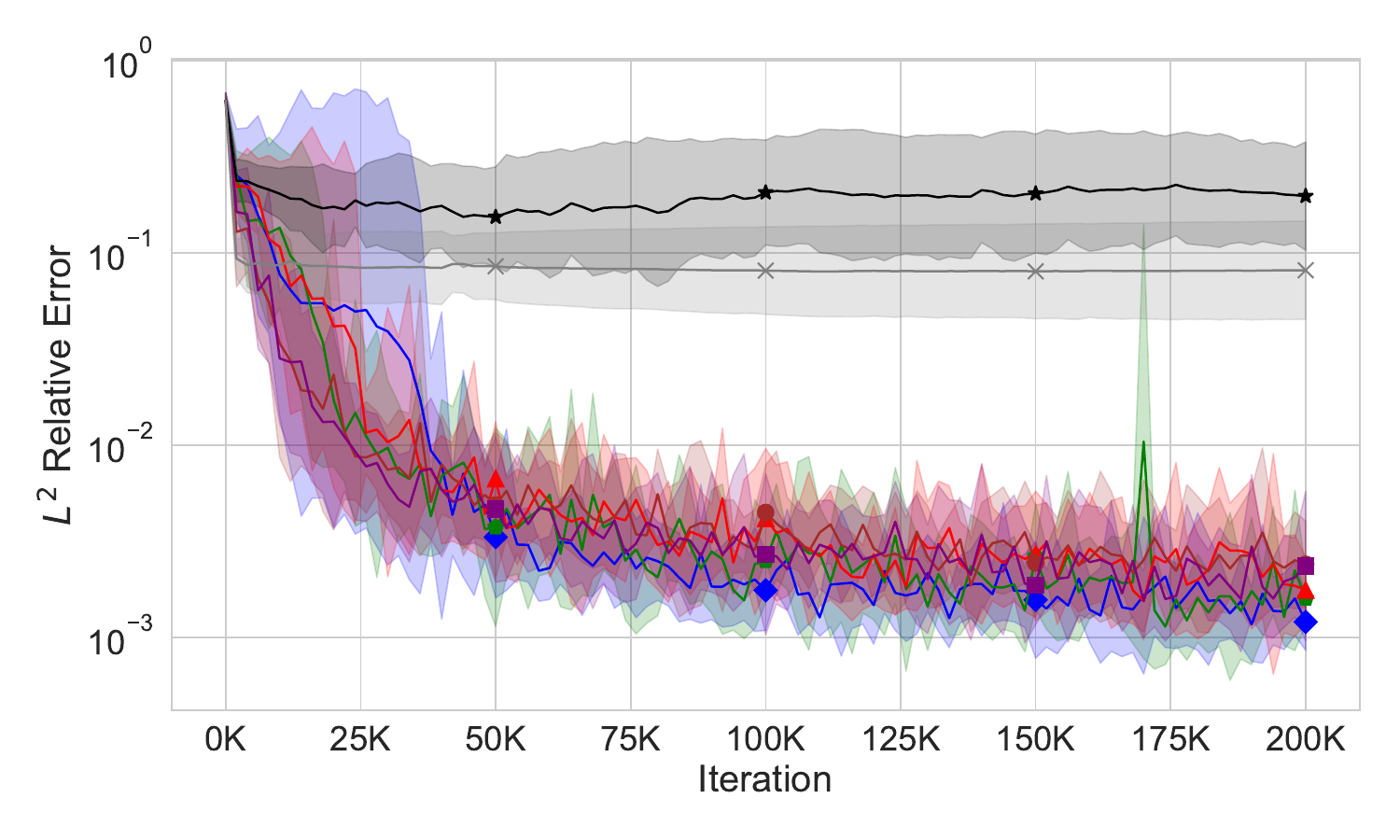}\mysubfigskip
        \caption{Burgers'}
    \end{subfigure}    
    \begin{subfigure}{\mysfigwidth}\mysubfigskip
        \centering
        \includegraphics[width=\linewidth]{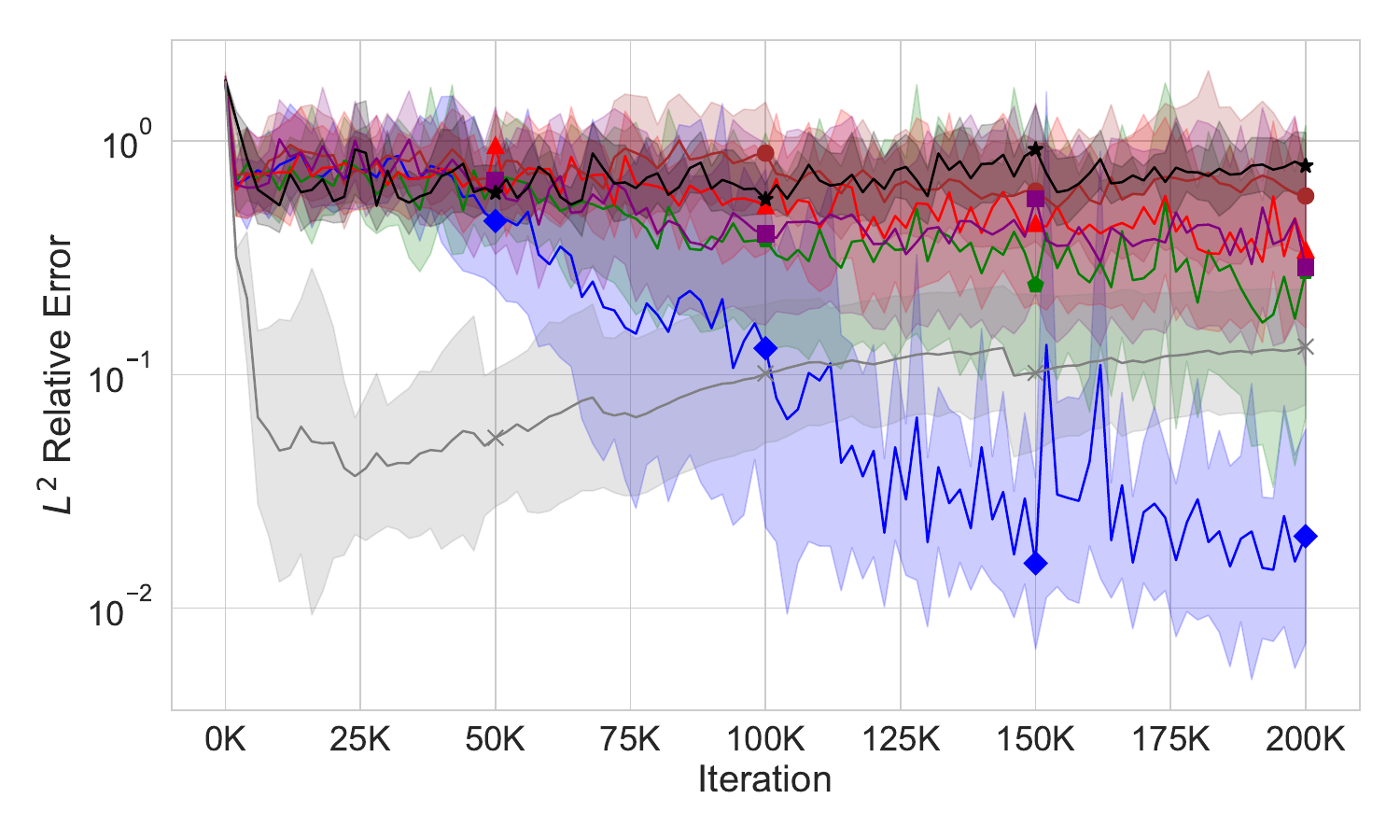}\mysubfigskip
        \caption{Wave}
    \end{subfigure}

    \begin{subfigure}{\mysfigwidth}
        \centering
        \includegraphics[width=\linewidth]{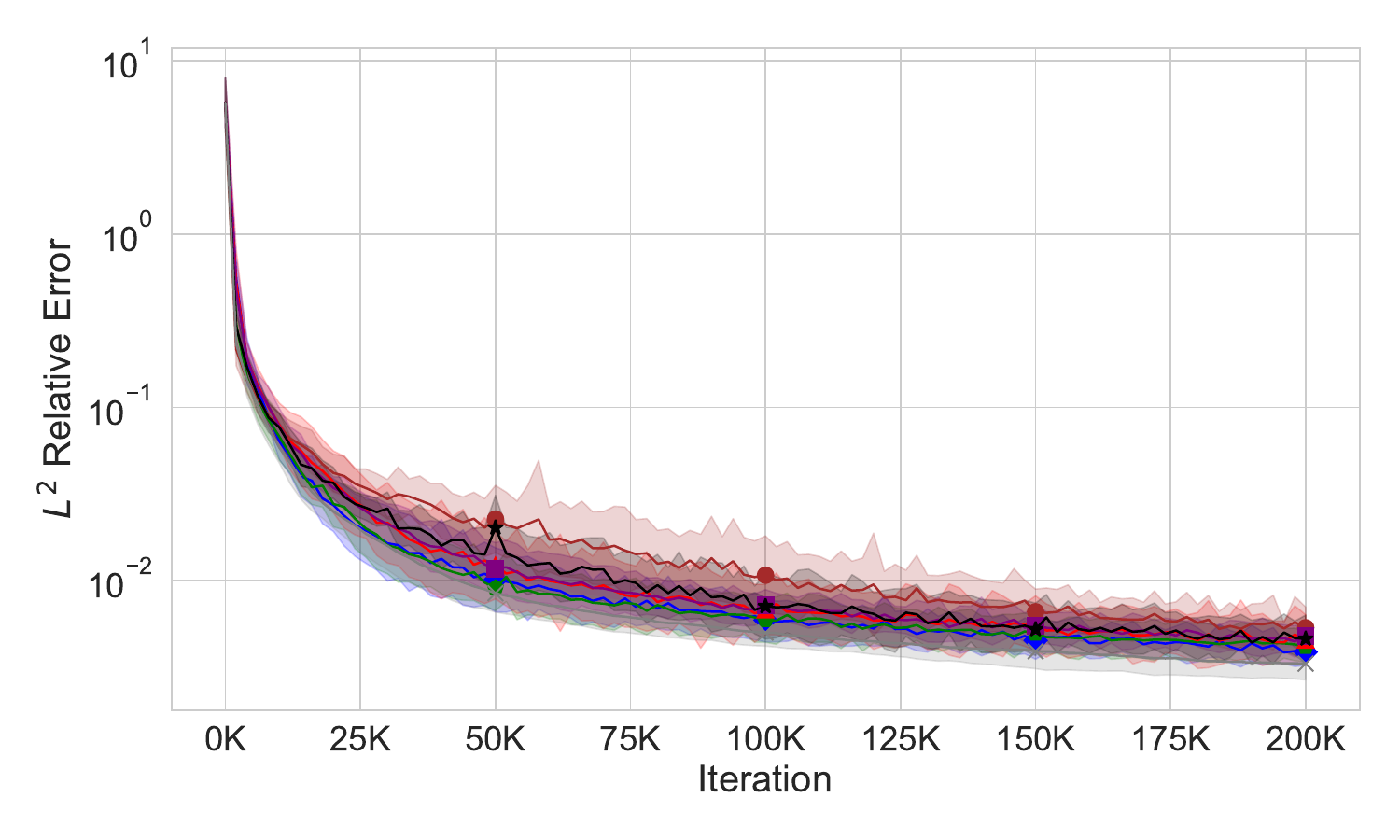}\mysubfigskip
        \caption{Drift-Diffusion}
    \end{subfigure}

    \caption{Comparison of different scoring functions for resampling on five PDEs for iteratively \textbf{replacing} the whole training set. Lines represent $L^2$ relative error averaged over 10 runs, with each point representing the model state after one cycle of \texttt{Adam+LBFGS}, and shaded regions indicating log-scale standard deviations.
    }
    \label{fig:results-l2_all_pdes_replace}
\end{figure}

For a more detailed illustration of how the training datasets evolve during training, we refer the reader to \cref{fig:diffusion_adding,fig:diffusion_replacing,fig:allen_cahn_adding,fig:allen_cahn_replacing,fig:burgers_adding,fig:burgers_replacing,fig:wave_adding,fig:wave_replacing,fig:drift_diffusion_sine_adding,fig:drift_diffusion_sine_replacing} in the Appendix. Furthermore, \cref{fig:boxplots_l2,fig:boxplots_test_loss}, present data included in the tables in the form of boxplots.

\subsection{Discussion}
We hypothesized that sampling training points based on their influence via an XAI-based training data attribution method would enhance the performance of PINNs. Our experimental results demonstrate that, in fact, PINN\-fluence is able to accurately detect influential samples and improve prediction accuracy by resampling, thereby validating the applicability of TDA methods in PINNs.
Using PINN\-fluence as a scoring function, we matched the performance of RAR and outperformed the other investigated approaches, especially in the replacement-based setting. This constitutes the first time, to our knowledge, that training data attribution methods from explainable AI have been applied to guide the training and improve the accuracy of physics-informed neural networks. 

Furthermore, compared to another influence-based method, Grad-Dot, the additional Hessian computation in PINN\-fluence generally proved beneficial to the performance and also showed different patterns in score assignment, visible in \cref{app:training-data-dynamics}. 
Output-grad produced the most distinctive patterns, as it highlights areas where predictions vary the most, in contrast to all other methods which are based on the loss. We hypothesized that for the Allen-Cahn and Burgers' equations exactly these regions would be most influential. While this is in fact the case for the Burgers' equation, Output-Grad still underperformed in comparison. 
Grad-Dot approximates the influence function by replacing the inverse Hessian with an identity matrix, effectively capturing the gradient alignment between $L_{\text{test}}$ and the loss of a candidate point. Loss-Grad, on the other hand, measures the magnitude of the loss gradient with respect to the model parameters, emphasizing points that induce high parameter changes. Although conceptually different, both methods assigned similar scores over most PDEs.

Overall, we observe that among the approaches we have investigated, methods other than PINN\-fluence-based scoring and RAR failed to provide reliable performance improvements. Furthermore, the effectiveness of any resampling approach also depends on the specific PDE one is trying to solve. In some of our experiments, notably for the diffusion and the drift-diffusion equations, score-based resampling offered little to no advantage over the static or random sampling. Nevertheless, we did not observe any instances where the successful resampling methods significantly degrade accuracy compared to these baselines.

The primary limitation of our resampling approach is its computational cost. Residual-based resampling methods \cite{lu_deepxde_2021,wu_comprehensive_2023} evaluate the loss directly, requiring only a single forward pass through the PINN, with the exception of applying the PDE operator. 
The computation of PINN\-fluence as given in \cref{eq:pinnfl-full} presupposes the approximation of the Hessian of the loss function, the gradients of the loss of the test set and the candidate point to be added. All of these operations require (repeated) backward-passes through the network and thus require considerably more computational effort -- in addition to computing their inner product.
Thus, while PINN\-fluence provides a more theoretically motivated selection of training points, its scalability remains a challenge.
Further, our experiments were limited to a small set of forward problems, restricting generalizability.
We also focused solely on hard-constrained problems, omitting soft constraints.

\section{Conclusions}
In this work, we have explored the application of influence function-based resampling in the training of PINNs, demonstrating that our XAI-guided PINN\-fluence method effectively identifies influential training points. Notably, its performance matches that of RAR, currently the most widely adopted approach 
\cite{lu_deepxde_2021,wu_comprehensive_2023}. This showcases the potential of XAI-driven techniques in scientific machine learning, as such tailored sampling strategies can help address challenges in PINN training and enhance performance. Furthermore, our experiments, although limited in scope, showed that targeted resampling using PINN\-fluence or RAR is beneficial and does not incur performance penalties. Thus, we generally recommend incorporating this kind of resampling when training PINNs.
Additionally, we also observed that while some PDEs were relatively straightforward to learn for PINNs, others posed more of a challenge. 
This underscores the fact that the learning behavior of PINNs is fundamentally dependent on the PDE operator itself \cite{ryck_operator_2024}.
Future work could focus on exploring more efficient computations of influence functions to enhance their practicality. Additionally, combining influence functions with the neural tangent kernel framework presents itself as a promising direction, as such approaches have successfully been applied to improve PINN performance \cite{wang_when_2022,rathore_challenges_2024}.
\begin{credits}
\subsubsection{\ackname} This work was supported by the Fraunhofer Internal Programs under Grant No. PREPARE 40-08394.

\subsubsection{\discintname}
The authors declare that they have no competing interests.
\end{credits}
\subsubsection*{Code Availability}
The code is available at\\ \url{https://github.com/aleks-krasowski/PINNfluence_resampling}.

\bibliographystyle{splncs04}
\bibliography{references}

\clearpage

\appendix
\section{Appendix}

\subsection{Description of the Partial Differential Equations}\label{app:pdes}

In the following we describe the used PDEs with corresponding initial and boundary conditions, as well as the used neural network architectures. Except for the wave and drift-diffusion equation, the domain was set to $\Omega=(-1,1)\times(0,1)$ with the solutions $u:\overline\Omega\to\mathbb R$ where $u: (x,t)\mapsto u(x,t)$.

\begin{enumerate}
    \itemsep1.5em
    \item \textbf{Diffusion Equation}
    \begin{equation*}
        \frac{\partial u}{\partial t} = \frac{\partial^2 u}{\partial x^2} + (\pi^2-1) \sin(\pi x)\ e^{-t}
    \end{equation*}
    with initial and boundary conditions:
    \begin{align*}
        u(x, 0) &= \sin (\pi x), \\
        u(-1, t) &= u(1, t) = 0.
    \end{align*}
    Having the closed form solution
    \begin{align*}
        u(x,t) = \operatorname{sin}\left( \pi x \right) e^{-t}.
    \end{align*}
    {\itshape Neural network architecture:} 3 hidden layers with 32 neurons each.

    \item \textbf{Burgers' Equation}
    \begin{equation*}
        \frac{\partial u}{\partial t} + u\, \frac{\partial u}{\partial x} = \nu\, \frac{\partial^2 u}{\partial x^2}, \qquad \text{with } \nu = \frac{100}{\pi}
    \end{equation*}
    with initial and boundary conditions:
    \begin{align*}
        u(x, 0) &= -\sin (\pi x), \\
        u(-1, t) &= u(1, t) = 0.
    \end{align*}
    Ground truth values obtained from \cite{wu_comprehensive_2023}.\\
    {\itshape Neural network architecture:} 3 hidden layers with 32 neurons each.

    \item \textbf{Allen-Cahn Equation}
    \begin{equation*}
        \frac{\partial u}{\partial t} = D\, \frac{\partial^2 u}{\partial x^2} - 5(u - u^3), \qquad \text{with } D=\frac{1}{1000}
    \end{equation*}
    with initial and boundary conditions:
    \begin{align*}
        u(x, 0) &= x^2 \cos (\pi x), \\
        u(-1, t) &= u(1, t) = -1.
    \end{align*}
    Ground truth values obtained from \cite{wu_comprehensive_2023}.\\
    {\itshape Neural network architecture:} 3 hidden layers with 64 neurons each.

    \item \textbf{Wave Equation}
    The domain for the wave equation was set to $\Omega=(0,1)\times(0,1)$.
    \begin{equation*}
        \frac{\partial^2 u}{\partial t^2} - c^2 \frac{\partial^2 u}{\partial x^2} = 0 \qquad\text{with}\ c=2
    \end{equation*}
    with initial and boundary conditions:
    \begin{align*}
        u(x, 0) &= \sin (\pi x) + \frac{1}{2} \sin (4 \pi x), \\
        \frac{\partial u}{\partial t}(x, 0) &= 0, \\
        u(0, t) &= u(1, t) = 0.
    \end{align*}
    Having the closed form solution
    \begin{align*}
        u(x,t) = \sin(\pi x)\cos(\pi c t) + \frac{1}{2}\sin(4\pi x)\cos(4\pi c t).
    \end{align*}
    {\itshape Neural network architecture:} 5 hidden layers with 100 neurons each.

    \item \textbf{Drift-Diffusion Equation}
    The domain for the drift-diffusion equation was set to $\Omega=(0,2\pi)\times(0,1)$.
    \begin{equation*}
        \frac{\partial u}{\partial t} = \alpha\frac{\partial^2 u}{\partial x^2} - \beta \frac{\partial u}{\partial x}, \qquad \text{with } \alpha=1, \beta=20
    \end{equation*}
    and initial and boundary conditions:
    \begin{align*}
        u(x, 0) &= \operatorname{sin}\!\left(2x + \frac{\pi}{4}\right) , \\
        u(0,t) &= \operatorname{sin}\left(\frac{\pi}{4} - 2 \cdot \beta  t\right)\ e^{-4 \alpha t}, \\
        u(2\pi,t) &= \operatorname{sin}\left(\frac{17\pi}{4} - 2 \beta t\right)\  e^{-4 \alpha t}.
    \end{align*}
    Having the closed form solution
    \begin{align*}
        u(x,t) = \sin\!\left(2x-2\beta t+\frac{\pi}{4}\right)\ e^{-4\alpha t}\ .
    \end{align*}
    {\itshape Neural network architecture:} 3 hidden layers with 64 neurons each.
\end{enumerate}

\clearpage

\subsection{Supplementary Tables and Figures}\label{app:supp-stuff}

\setcounter{table}{0}  
\setcounter{figure}{0}  
\counterwithin{table}{section}
\counterwithin{figure}{section}

\begin{table}
    \centering
    \small
    \caption{Mean loss and standard deviation for sampling strategies when adding train points ($\mathcal{X}_{\text{train}} \leftarrow \mathcal{X}_{\text{train}} \cup \mathcal{X}_{\text{new}}$). The smallest loss is highlighted in bold. Columns are given in the same order of magnitude (OOM).}
    \label{tab:add_losses}
    \begin{tabular}{
        l | c | c | c | c | c
    }
        \toprule
        PDE & Diffusion & Allen-Cahn & Burgers' & Wave & Drift-Diffusion \\ 
        {[OOM]}  & $[10^{-7}]$ & $[10^{-6}]$ & $[10^{-6}]$ & $[10^{-2}]$ & $[10^{-4}]$ \\
        \midrule 
        \midrule
        PINNfluence & $2.0{\scriptstyle \pm 0.7}$ & $2.8{\scriptstyle \pm 0.8}$ & $34.9{\scriptstyle \pm 9.9}$ & $2.0{\scriptstyle \pm 0.2}$ & $\mathbf{7.3{\scriptstyle \pm 1.9}}$ \\
        RAR & $\mathbf{1.8{\scriptstyle \pm 0.5}}$ & $\mathbf{1.9{\scriptstyle \pm 0.5}}$ & $\mathbf{7.3{\scriptstyle \pm 3.2}}$ & $\mathbf{1.5{\scriptstyle \pm 0.2}}$ & $8.1{\scriptstyle \pm 2.7}$ \\
        Grad-Dot & $2.6{\scriptstyle \pm 0.9}$ & $7.2{\scriptstyle \pm 5.7}$ & $84.2{\scriptstyle \pm 21.8}$ & $2.6{\scriptstyle \pm 0.8}$ & $11.6{\scriptstyle \pm 5.6}$ \\
        Out-Grad & $31.6{\scriptstyle \pm 18.5}$ & $5.2{\scriptstyle \pm 1.2}$ & $473477{\scriptstyle \pm 1007204}$ & $2704{\scriptstyle \pm 3045}$ & $257{\scriptstyle \pm 776}$ \\
        Loss-Grad & $3.4{\scriptstyle \pm 0.7}$ & $4.5{\scriptstyle \pm 1.0}$ & $76.0{\scriptstyle \pm 26.7}$ & $3.1{\scriptstyle \pm 1.1}$ & $7.7{\scriptstyle \pm 2.1}$ \\
        Random & $10.4{\scriptstyle \pm 5.9}$ & $5.6{\scriptstyle \pm 2.2}$ & $143808{\scriptstyle \pm 219160}$ & $13339{\scriptstyle \pm 29940}$ & $28.6{\scriptstyle \pm 13.9}$ \\
        Static & $7020{\scriptstyle \pm 7562}$ & $16.2{\scriptstyle \pm 7.6}$ & $200653{\scriptstyle \pm 142082}$ & $22589{\scriptstyle \pm 24609}$ & $52.2{\scriptstyle \pm 79.5}$ \\
        \bottomrule
    \end{tabular}
\end{table}

\begin{table}
    \centering
    \small
    \caption{Mean loss and standard deviation for sampling strategies when replacing the whole training set ($\mathcal{X}_{\text{train}} \leftarrow \mathcal{X}_{\text{new}}$). The smallest loss is highlighted in bold. Columns are given in the same order of magnitude (OOM).}
    \label{tab:replace_losses}
    \begin{tabular}{
        l | c | c | c | c | c
    }
        \toprule
        PDE & Diffusion & Allen-Cahn & Burgers' & Wave & Drift-Diffusion \\ 
        {[OOM]}  & $[10^{-6}]$ & $[10^{-6}]$ & $[10^{-5}]$ & $[10^{1}]$ & $[10^{-3}]$ \\
        \midrule 
        \midrule
        PINNfluence & $37.1{\scriptstyle \pm 41.0}$ & $\mathbf{3.1{\scriptstyle \pm 0.9}}$ & $\mathbf{2.1{\scriptstyle \pm 0.9}}$ & $\mathbf{1.5{\scriptstyle \pm 2.9}}$ & $\mathbf{1.3{\scriptstyle \pm 0.5}}$ \\
        RAR & $19.1{\scriptstyle \pm 21.5}$ & $8.9{\scriptstyle \pm 14.9}$ & $4.7{\scriptstyle \pm 2.4}$ & $138{\scriptstyle \pm 246}$ & $1.4{\scriptstyle \pm 0.4}$ \\
        Grad-Dot & $26.0{\scriptstyle \pm 57.6}$ & $5.9{\scriptstyle \pm 1.8}$ & $3.5{\scriptstyle \pm 1.2}$ & $334{\scriptstyle \pm 323}$ & $2.2{\scriptstyle \pm 1.3}$ \\
        Out-Grad & $\mathbf{7.4{\scriptstyle \pm 5.7}}$ & $209198{\scriptstyle \pm 303219}$ & $6.0{\scriptstyle \pm 3.9}$ & $977{\scriptstyle \pm 801}$ & $2.8{\scriptstyle \pm 2.2}$ \\
        Loss-Grad & $7.8{\scriptstyle \pm 10.5}$ & $2715{\scriptstyle \pm 8571}$ & $3.6{\scriptstyle \pm 2.5}$ & $233{\scriptstyle \pm 250}$ & $2.5{\scriptstyle \pm 1.4}$ \\
        Random & $8.5{\scriptstyle \pm 9.7}$ & $24544{\scriptstyle \pm 58117}$ & $234971{\scriptstyle \pm 301946}$ & $1437{\scriptstyle \pm 988}$ & $2.8{\scriptstyle \pm 1.3}$ \\
        Static & $702{\scriptstyle \pm 756}$ & $16.2{\scriptstyle \pm 7.6}$ & $20065{\scriptstyle \pm 14208}$ & $22.6{\scriptstyle \pm 24.6}$ & $5.2{\scriptstyle \pm 7.9}$ \\
        \bottomrule
    \end{tabular}
\end{table}

\clearpage

\subsection{Training Point Selection Dynamics}\label{app:training-data-dynamics}

The following figures illustrate the training point selection dynamics for each PDE and sampling method. We present results for both \textit{adding} new points to the training set and \textit{replacing} existing ones. A general observation is the distinct behavior of Output-Grad, which consistently highlights areas of high prediction gradients across all test cases. In contrast, for all other methods the area of highest importance evolves throughout training over different areas, a behavior that is particularly visible in the wave and drift-diffusion equation examples.

Note that as we do not utilize pretraining, i.e.\ $\mathcal{X}_{\text{train}}$ is drawn directly at the beginning of training utilizing a scoring method in the \textit{replacing} setting. Additionally, recall that we sample from a discrete probability mass function defined by \cref{eq:res-pmf} with different parameters between the \textit{adding} and \textit{replacing} settings.

\newlength{\methodcolwidth}
\setlength{\methodcolwidth}{2cm} 

\begin{figure}
    \centering
    \begin{tabular}{>{\centering\arraybackslash}m{\methodcolwidth} >{\centering\arraybackslash}p{0.9\textwidth - \methodcolwidth - 0.5cm}} 
        PINN\-fluence & \includegraphics[align=c,trim={0 0 0 0},clip,width=\linewidth]{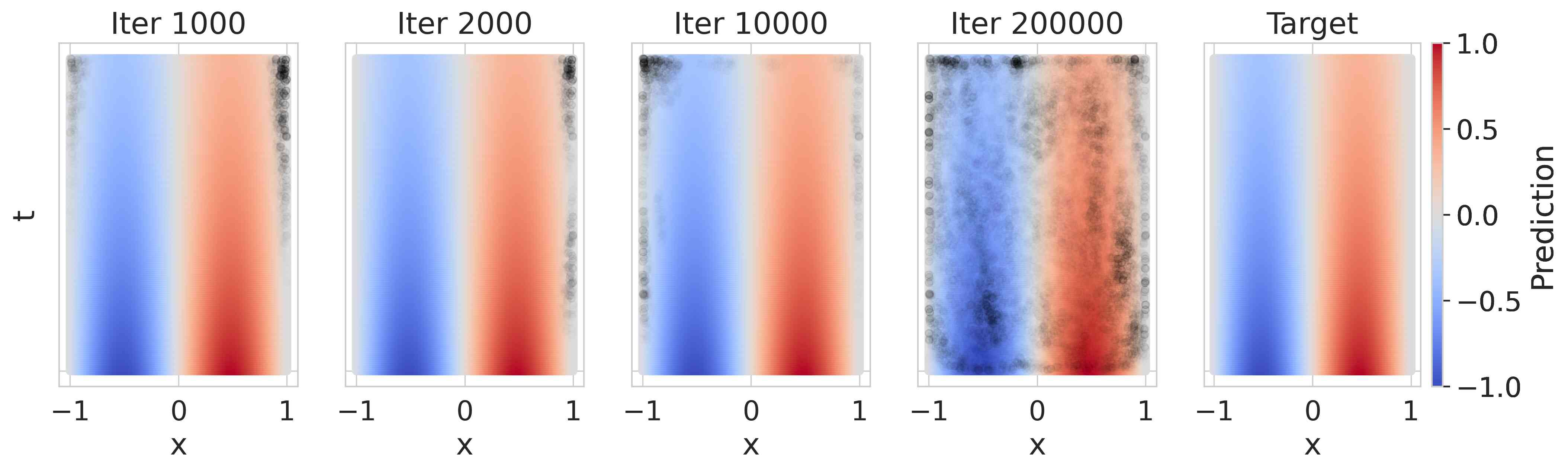} \\
        \addlinespace
        RAR&  \includegraphics[align=c,trim={0 0 0 0},clip,width=\linewidth]{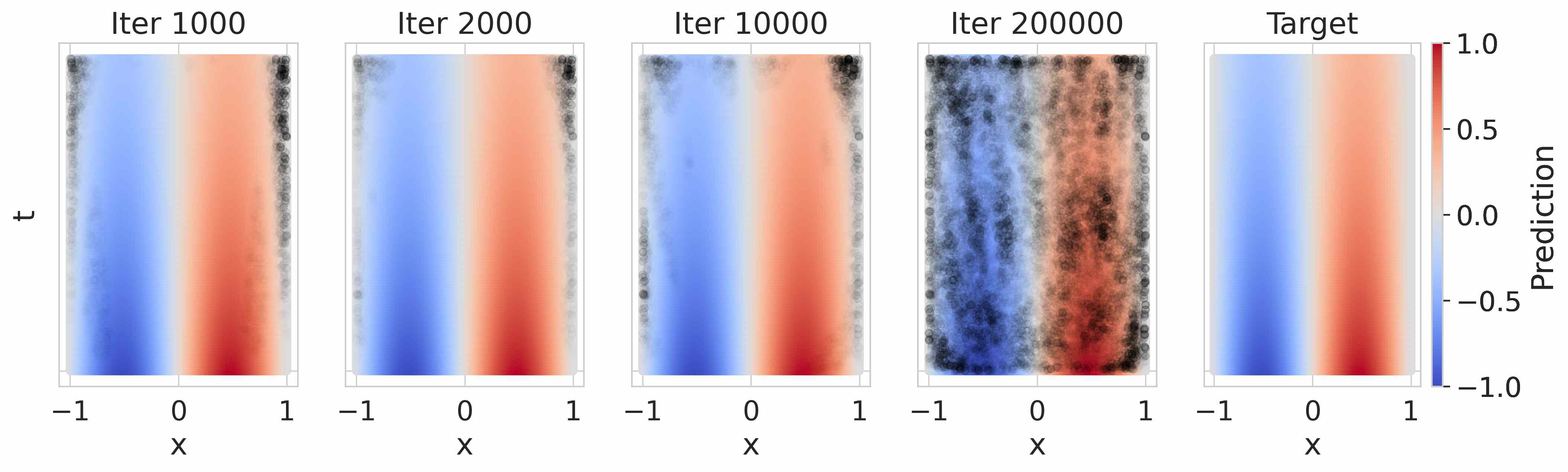} \\
        \addlinespace
        Grad-Dot&  \includegraphics[align=c,trim={0 0 0 0},clip,width=\linewidth]{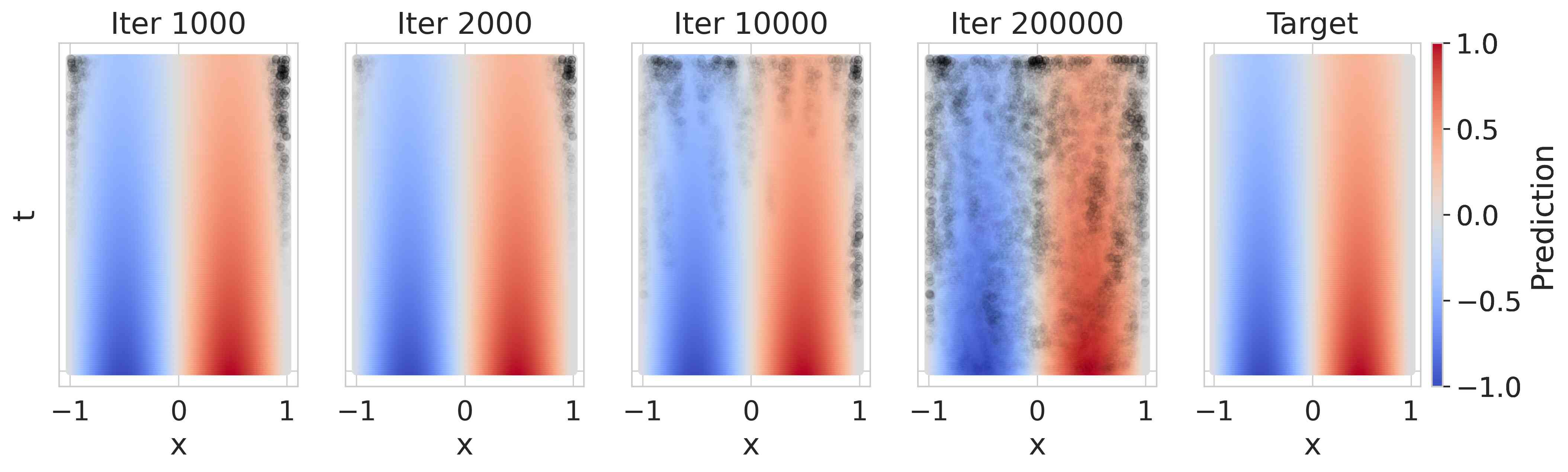} \\
        \addlinespace
        Output-Grad & \includegraphics[align=c,trim={0 0 0 0},clip,width=\linewidth]{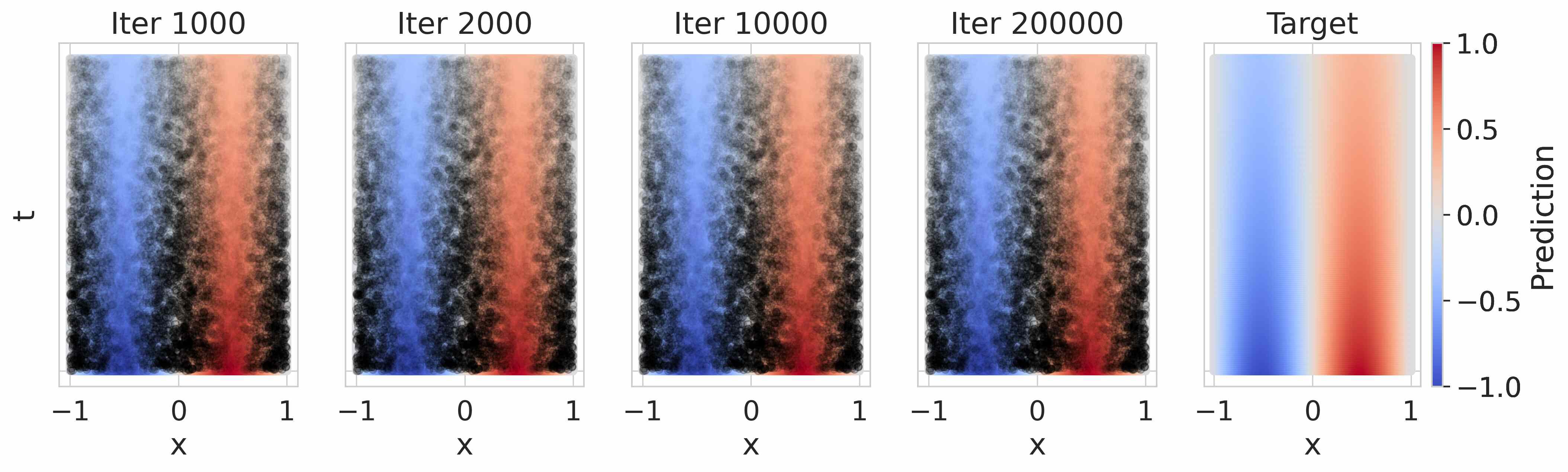} \\
        \addlinespace
        Loss-Grad & \includegraphics[align=c,trim={0 0 0 0},clip,width=\linewidth]{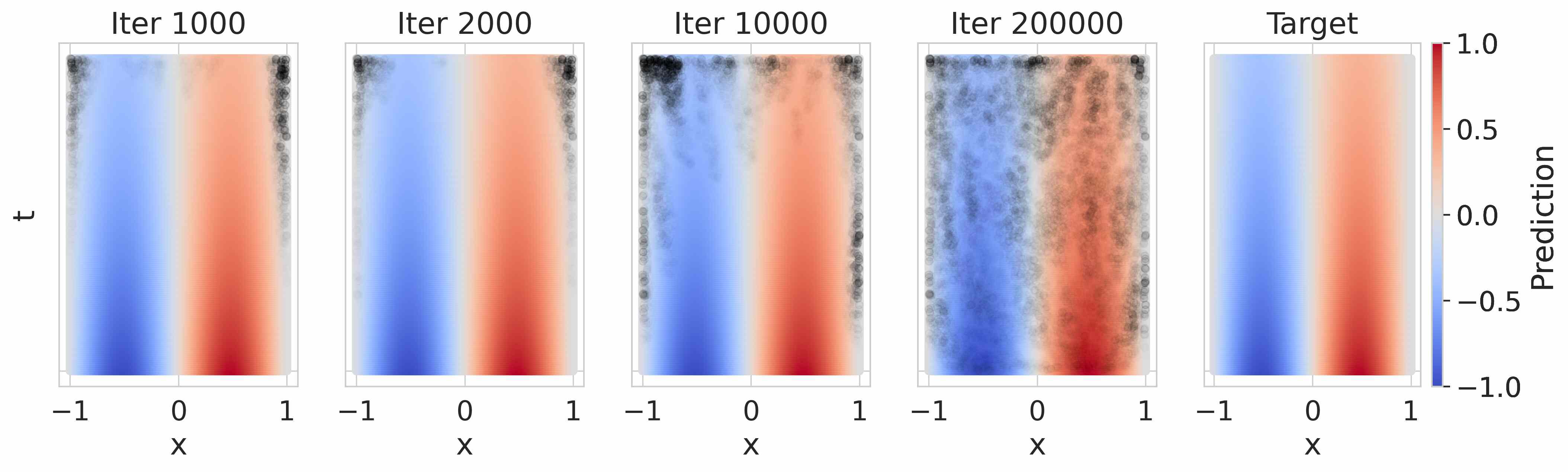} \\
    \end{tabular}
    \caption{\textbf{Diffusion \emph{(Adding)}}: Training point selection dynamics throughout training across respective scoring methods, averaged over 10 runs. Black points represent sampling probability determined by \cref{eq:res-pmf} with $\alpha=2$ and $c=0$. High transparency indicates low sampling probability. Values are clipped to $[-1,1]$.}
    \label{fig:diffusion_adding}
\end{figure}

\begin{figure}
    \centering
    \begin{tabular}{>{\centering\arraybackslash}m{\methodcolwidth} >{\centering\arraybackslash}p{0.9\textwidth - \methodcolwidth - 0.5cm}}
        PINN\-fluence&  \includegraphics[align=c,trim={0 0 0 0},clip,width=\linewidth]{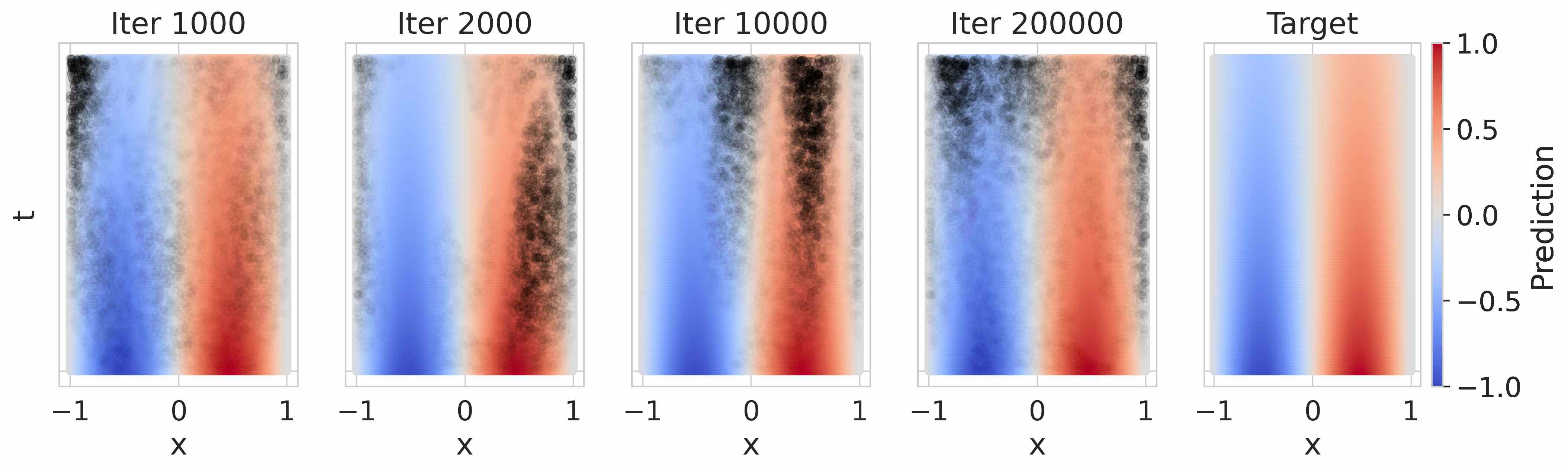} \\
        \addlinespace
        RAR&  \includegraphics[align=c,trim={0 0 0 0},clip,width=\linewidth]{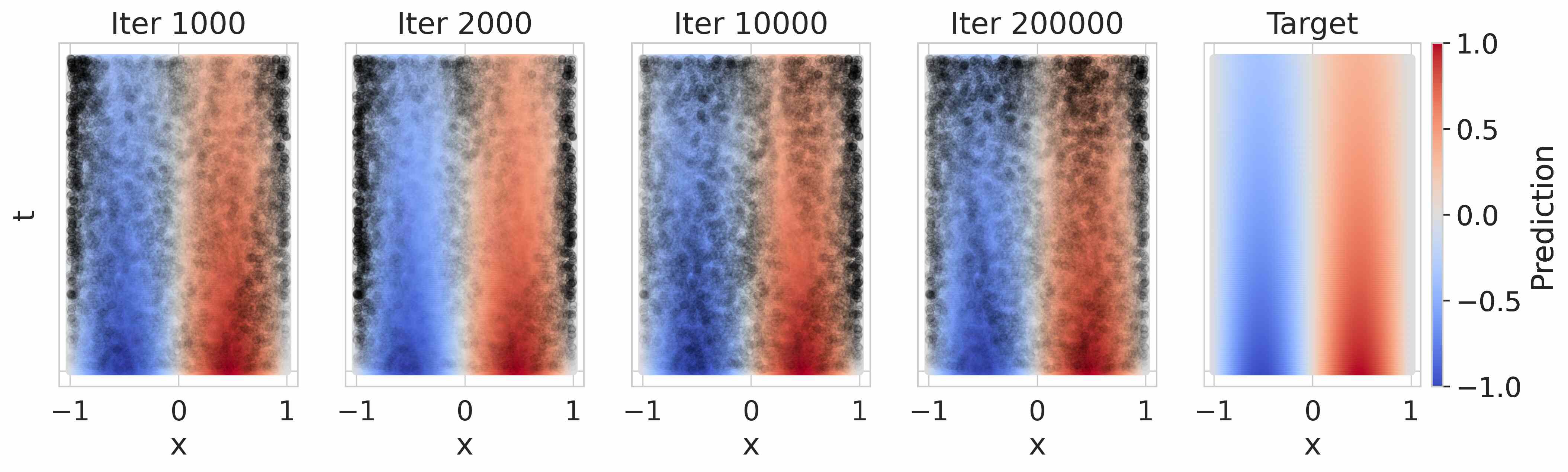} \\
        \addlinespace
        Grad-Dot&  \includegraphics[align=c,trim={0 0 0 0},clip,width=\linewidth]{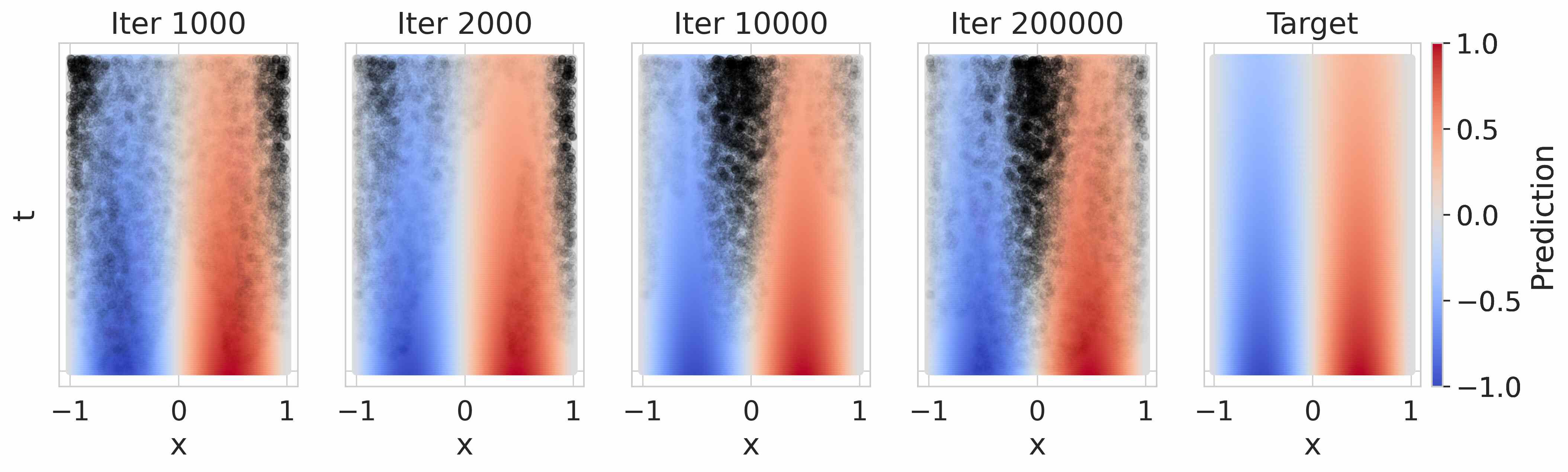} \\
        \addlinespace
        Output-Grad & \includegraphics[align=c,trim={0 0 0 0},clip,width=\linewidth]{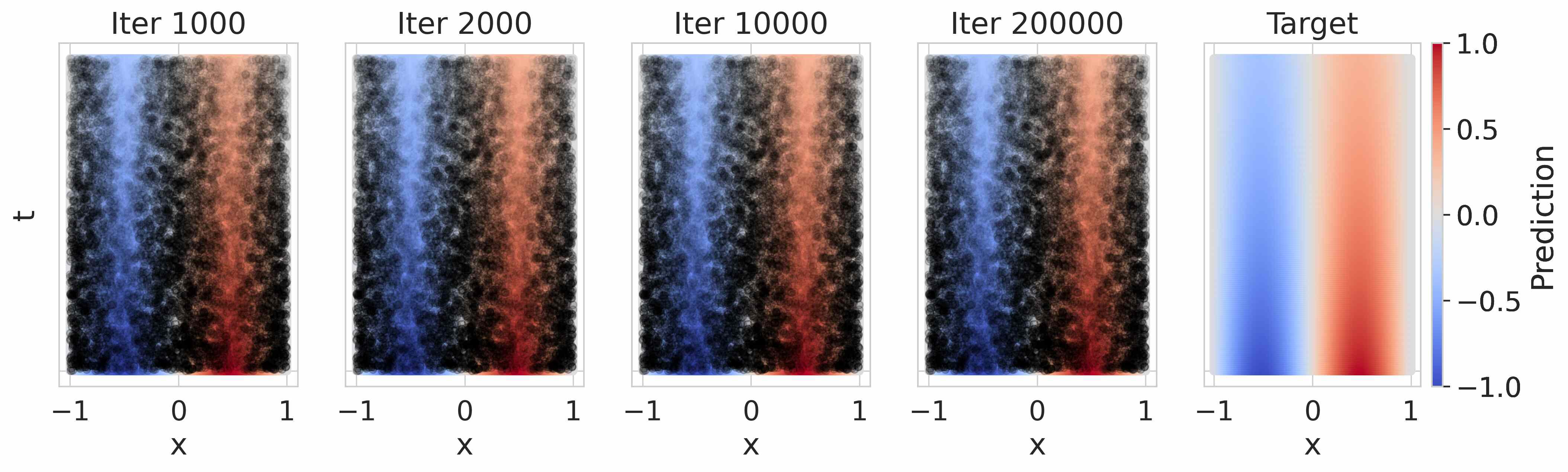} \\
        \addlinespace
        Loss-Grad & \includegraphics[align=c,trim={0 0 0 0},clip,width=\linewidth]{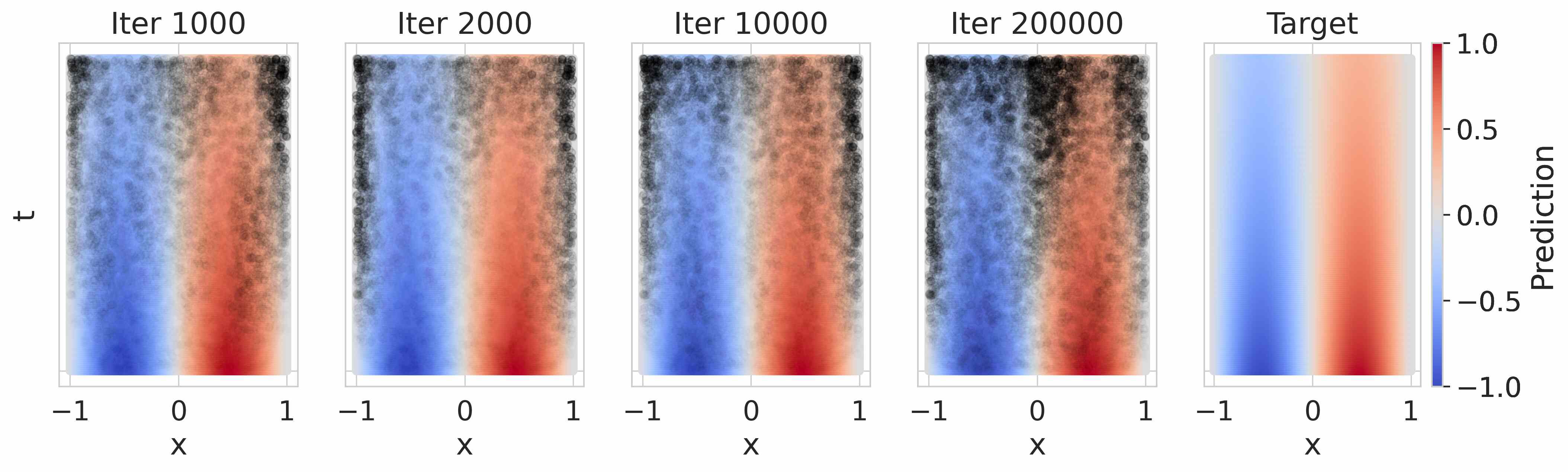} \\
    \end{tabular}
    \caption{\textbf{Diffusion \emph{(Replacing)}}: Training point selection dynamics throughout training across respective scoring methods, averaged over 10 runs. Black points represent sampling probability determined by \cref{eq:res-pmf} with $\alpha=1$ and $c=1$. High transparency indicates low sampling probability. Values are clipped to $[-1,1]$.}
    \label{fig:diffusion_replacing}
\end{figure}

\begin{figure}
    \centering
    \begin{tabular}{>{\centering\arraybackslash}m{\methodcolwidth} >{\centering\arraybackslash}p{0.9\textwidth - \methodcolwidth - 0.5cm}}
        PINN\-fluence&  \includegraphics[align=c,trim={0 0 0 0},clip,width=\linewidth]{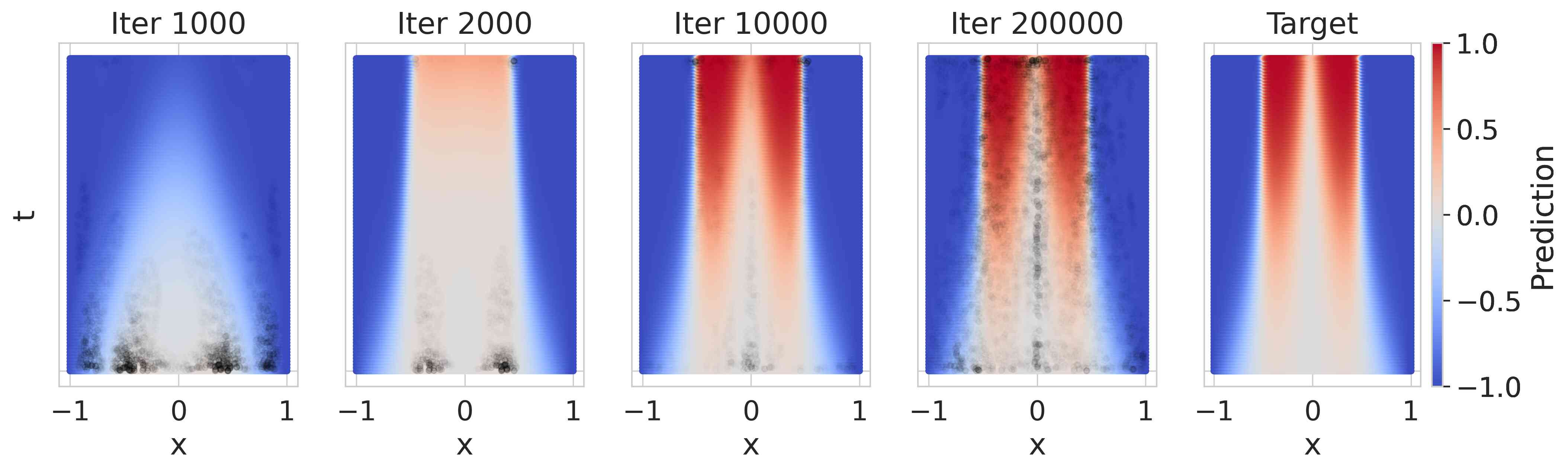} \\
        \addlinespace
        RAR&  \includegraphics[align=c,trim={0 0 0 0},clip,width=\linewidth]{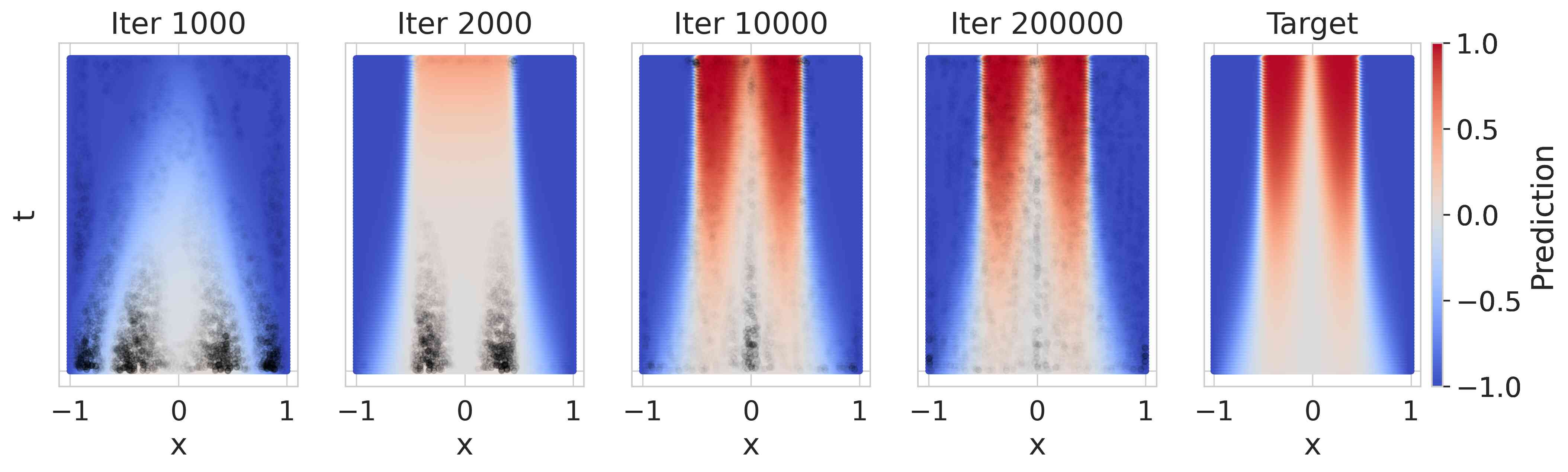} \\
        \addlinespace
        Grad-Dot&  \includegraphics[align=c,trim={0 0 0 0},clip,width=\linewidth]{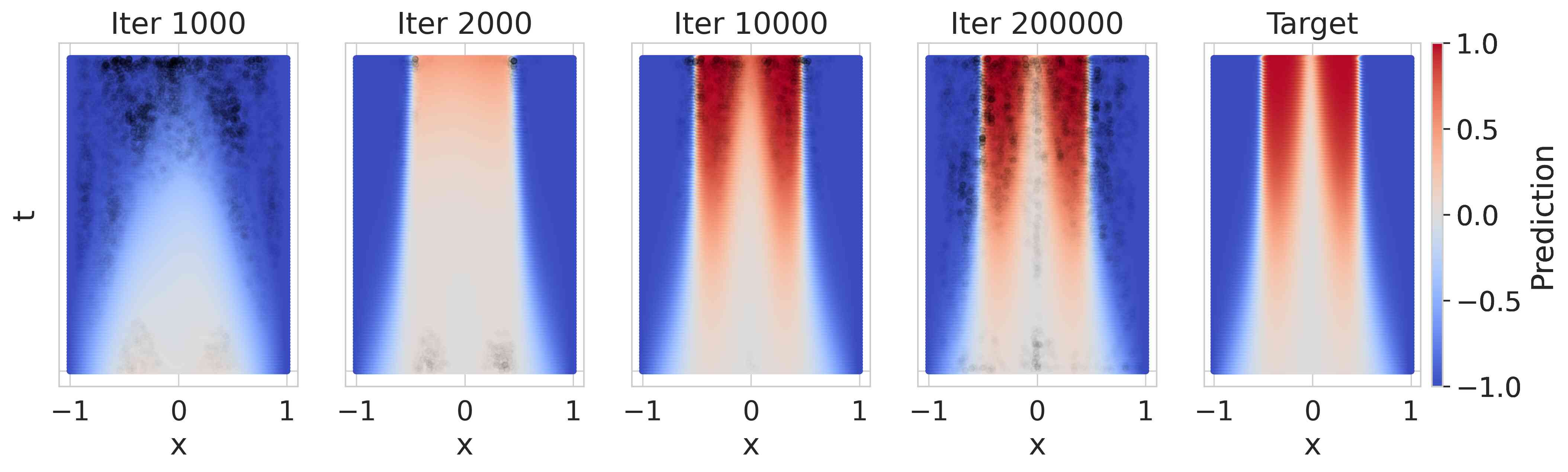} \\
        \addlinespace
        Output-Grad & \includegraphics[align=c,trim={0 0 0 0},clip,width=\linewidth]{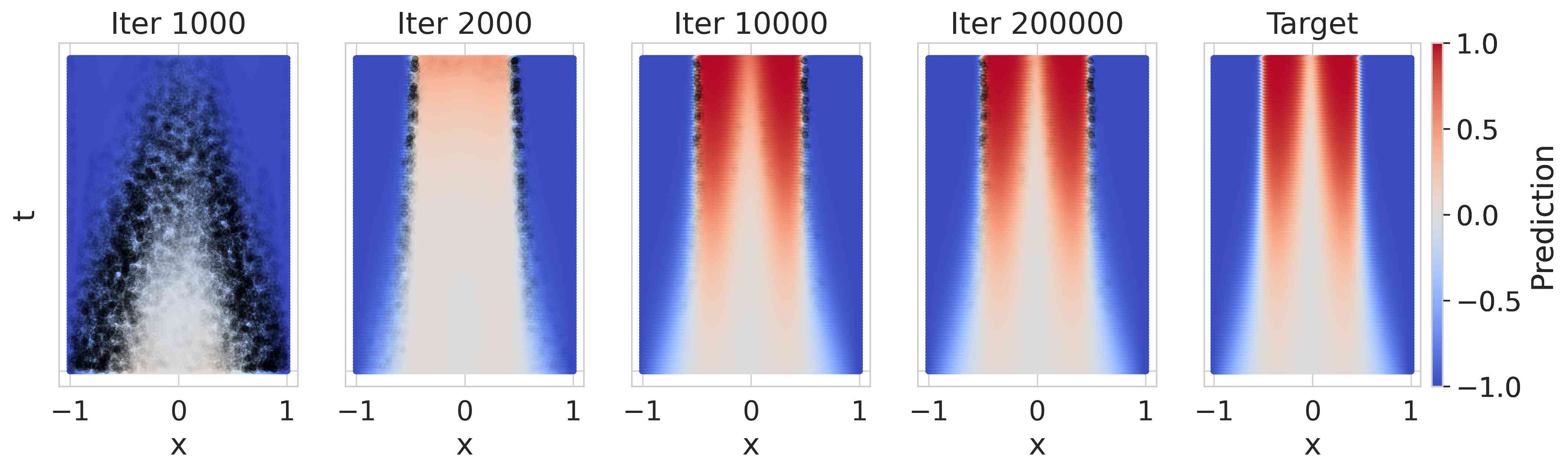} \\
        \addlinespace
        Loss-Grad & \includegraphics[align=c,trim={0 0 0 0},clip,width=\linewidth]{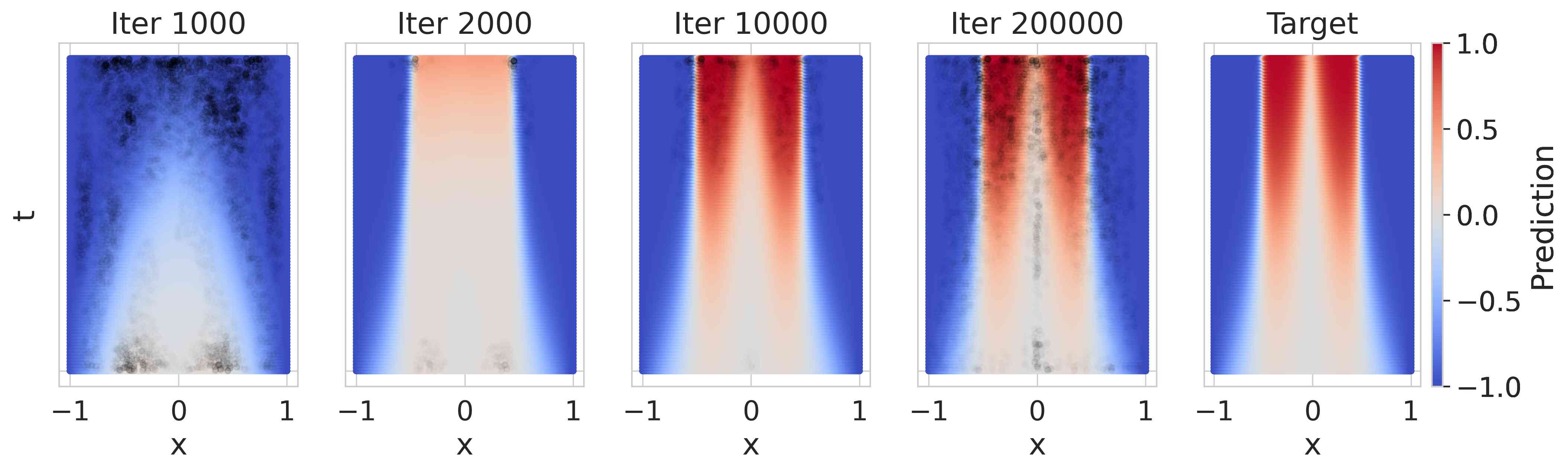} \\
    \end{tabular}
    \caption{\textbf{Allen-Cahn \emph{(Adding)}}: Training point selection dynamics throughout training across respective scoring methods, averaged over 10 runs. Black points represent sampling probability determined by \cref{eq:res-pmf} with $\alpha=2$ and $c=0$. High transparency indicates low sampling probability. Values are clipped to $[-1,1]$.}
    \label{fig:allen_cahn_adding}
\end{figure}

\begin{figure}
    \centering
    \begin{tabular}{>{\centering\arraybackslash}m{\methodcolwidth} >{\centering\arraybackslash}p{0.9\textwidth - \methodcolwidth - 0.5cm}}
        PINN\-fluence & \includegraphics[align=c,trim={0 0 0 0},clip,width=\linewidth]{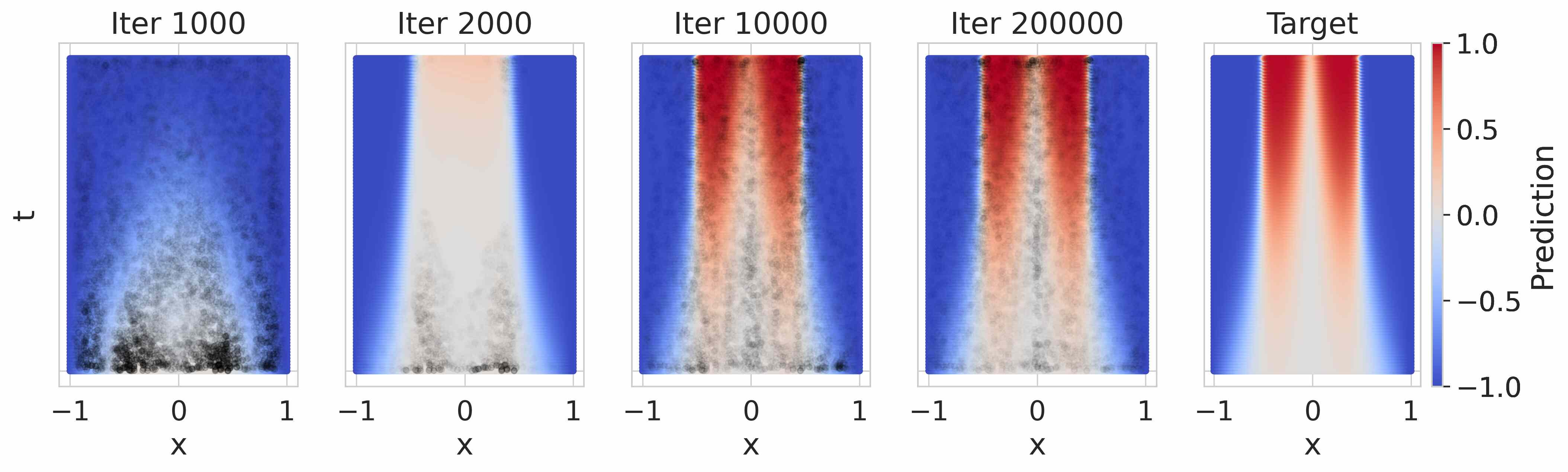} \\
        \addlinespace 
        RAR&  \includegraphics[align=c,trim={0 0 0 0},clip,width=\linewidth]{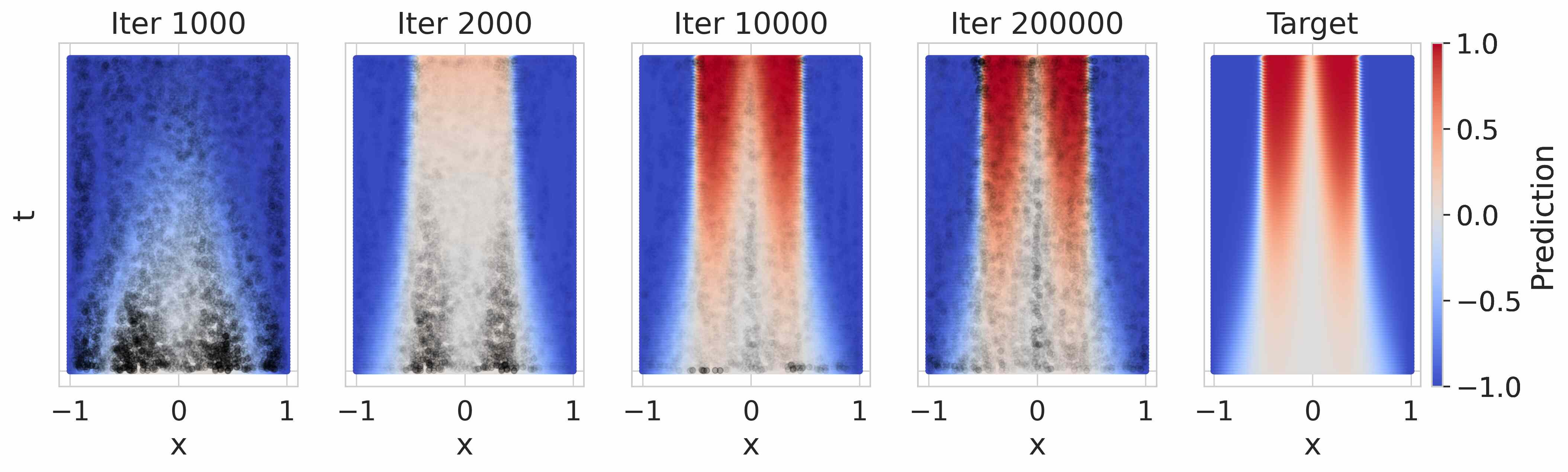} \\
        \addlinespace
        Grad-Dot&  \includegraphics[align=c,trim={0 0 0 0},clip,width=\linewidth]{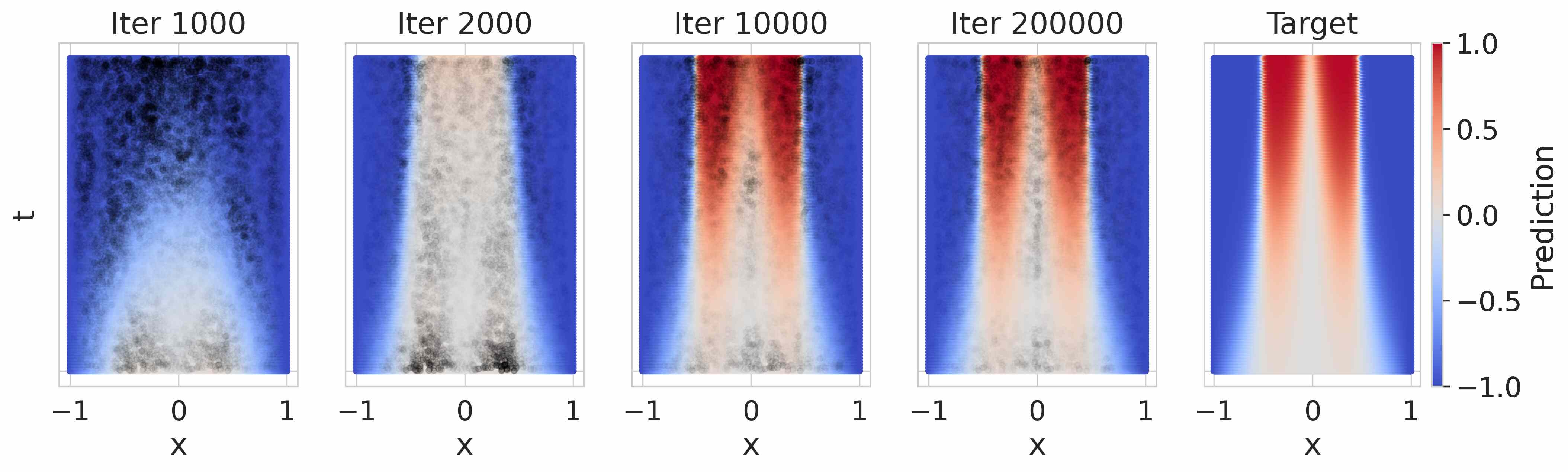} \\
        \addlinespace
        Output-Grad&  \includegraphics[align=c,trim={0 0 0 0},clip,width=\linewidth]{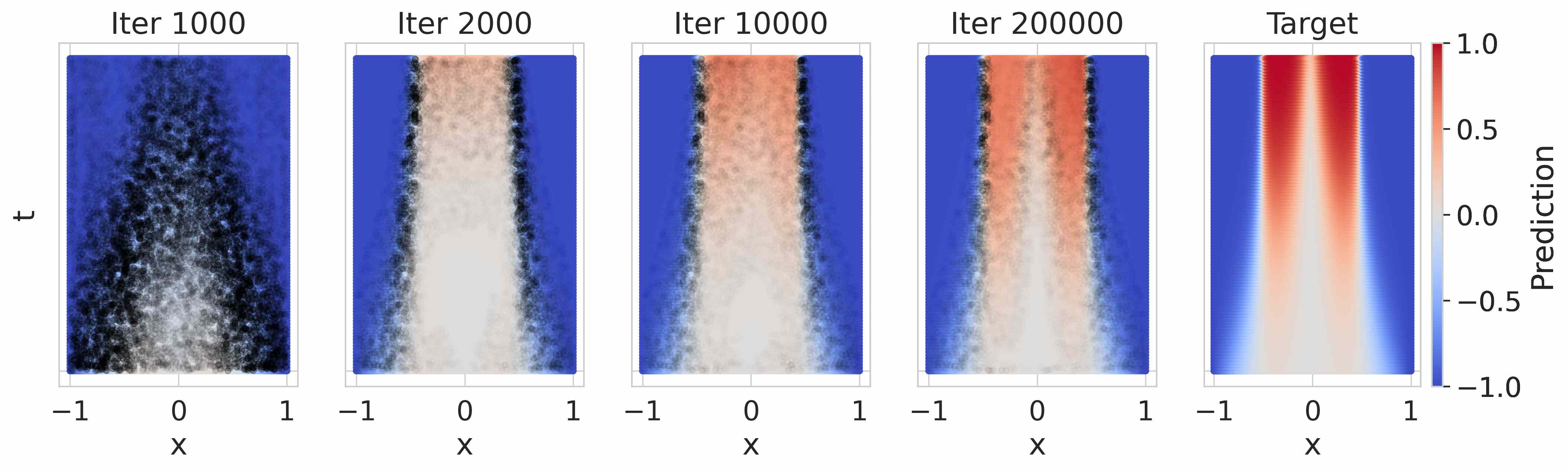} \\
        \addlinespace
        Loss-Grad&  \includegraphics[align=c,trim={0 0 0 0},clip,width=\linewidth]{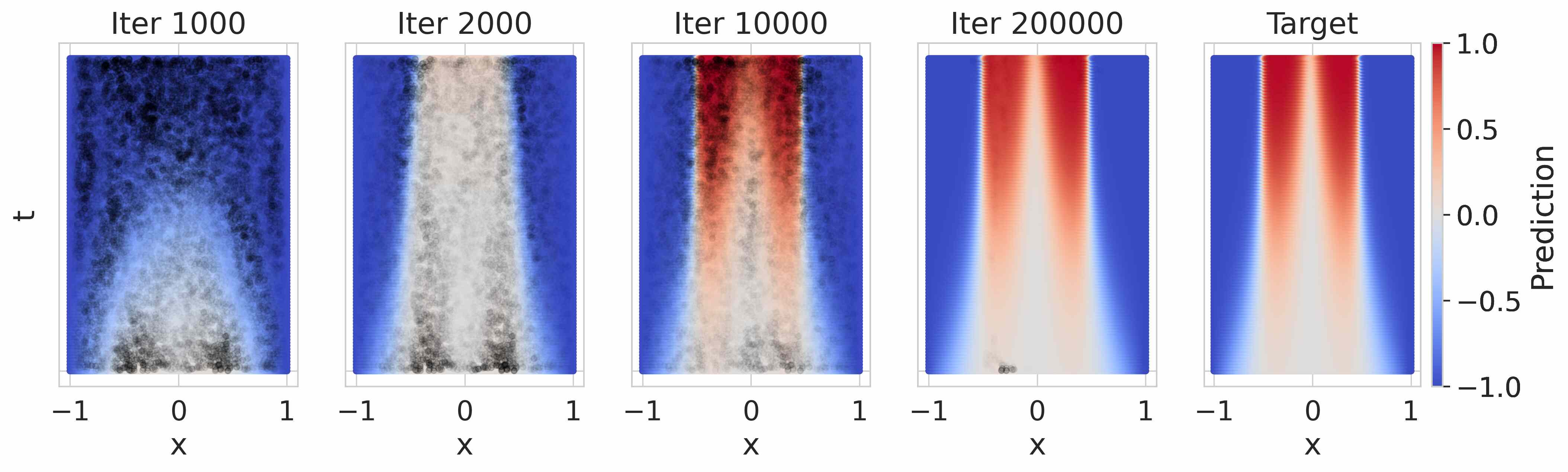} \\
    \end{tabular}
    \caption{\textbf{Allen-Cahn \emph{(Replacing)}}: Training point selection dynamics throughout training across respective scoring methods, averaged over 10 runs. Black points represent sampling probability determined by \cref{eq:res-pmf} with $\alpha=1$ and $c=1$. High transparency indicates low sampling probability. Values are clipped to $[-1,1]$.}
    \label{fig:allen_cahn_replacing}
\end{figure}

\begin{figure}
    \centering
    \begin{tabular}{>{\centering\arraybackslash}m{\methodcolwidth} >{\centering\arraybackslash}p{0.9\textwidth - \methodcolwidth - 0.5cm}}
        PINN\-fluence&  \includegraphics[align=c,trim={0 0 0 0},clip,width=\linewidth]{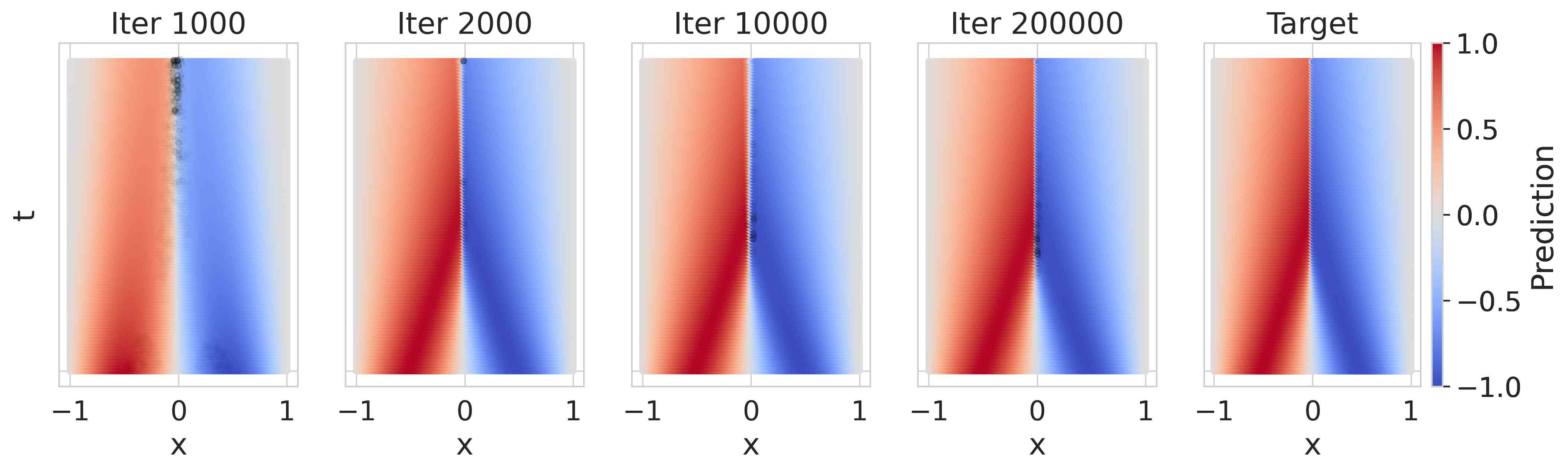} \\
        \addlinespace
        RAR&  \includegraphics[align=c,trim={0 0 0 0},clip,width=\linewidth]{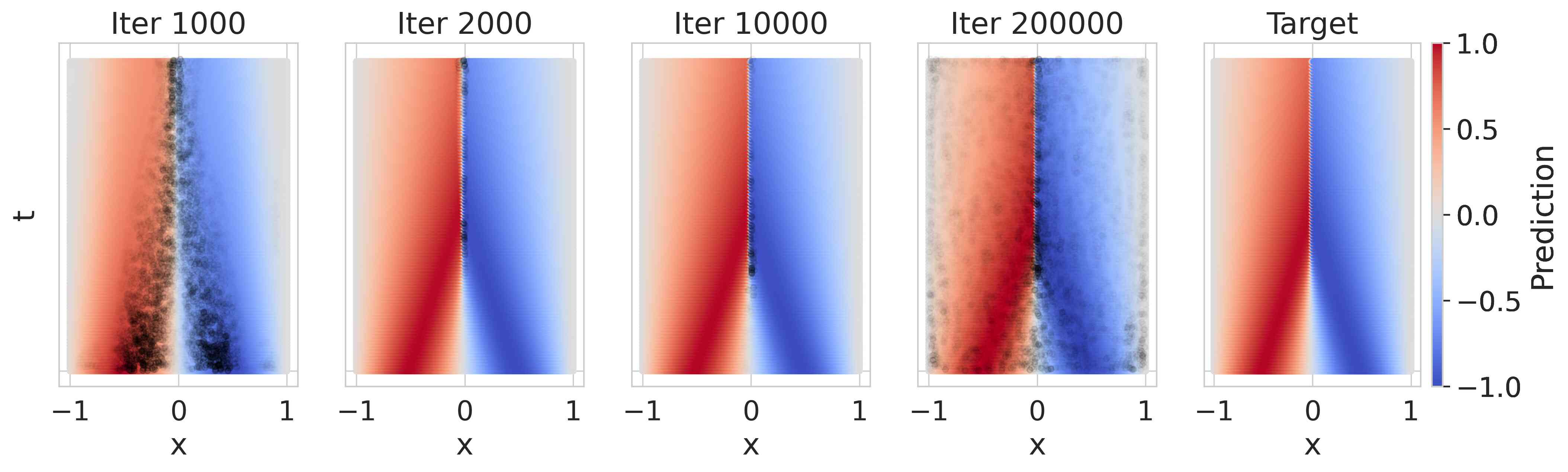} \\
        \addlinespace
        Grad-Dot&  \includegraphics[align=c,trim={0 0 0 0},clip,width=\linewidth]{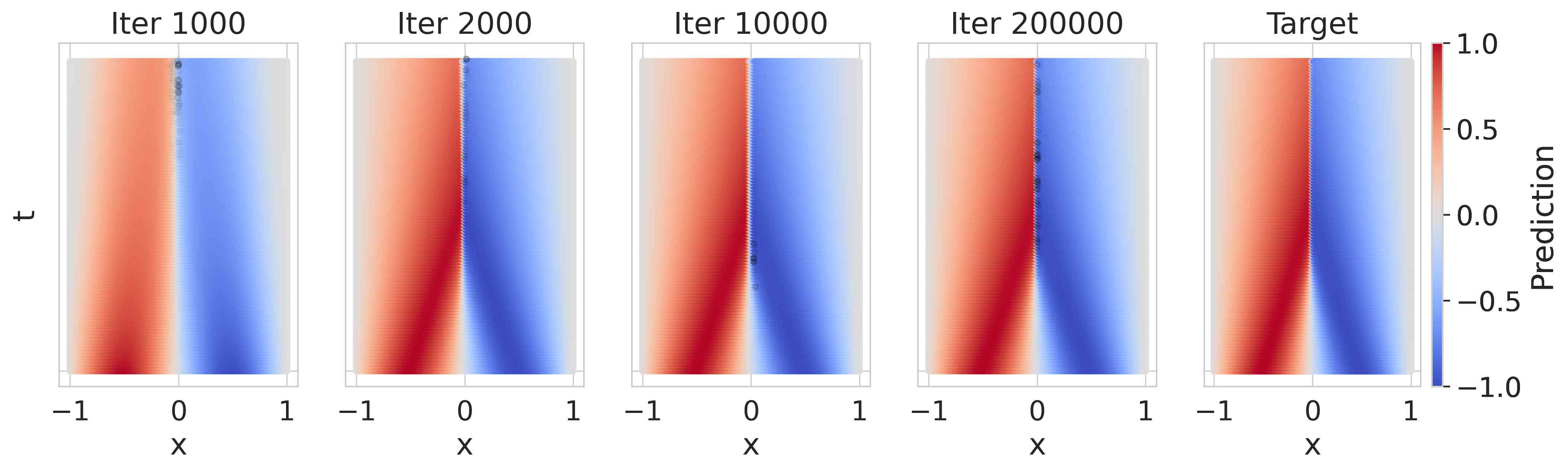} \\
        \addlinespace
        Output-Grad&  \includegraphics[align=c,trim={0 0 0 0},clip,width=\linewidth]{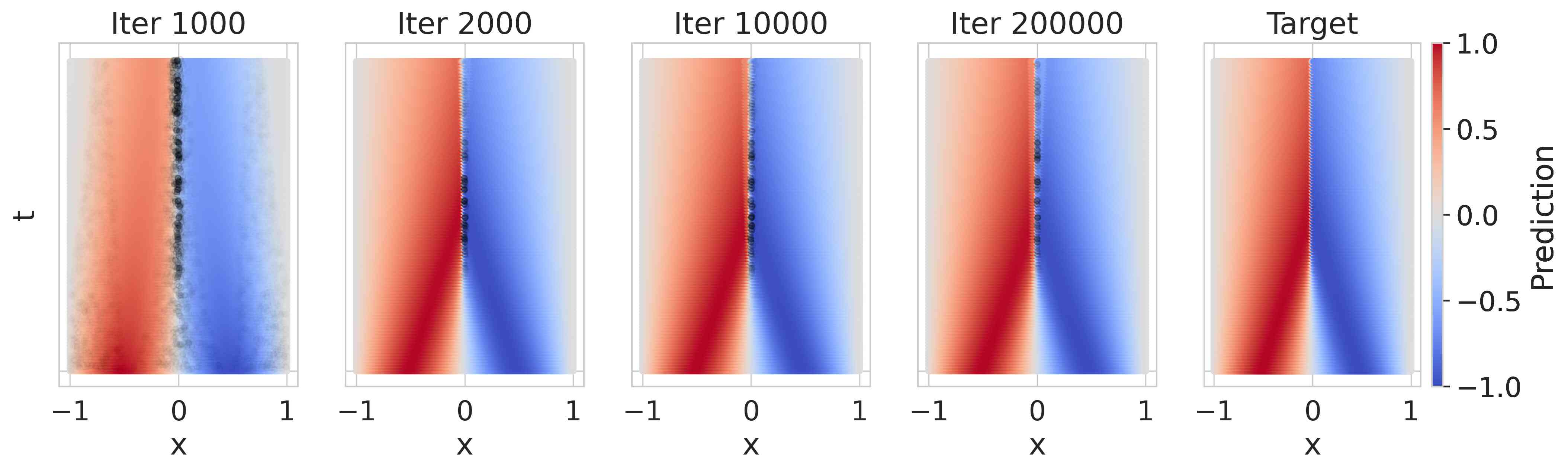} \\
        \addlinespace
        Loss-Grad&  \includegraphics[align=c,trim={0 0 0 0},clip,width=\linewidth]{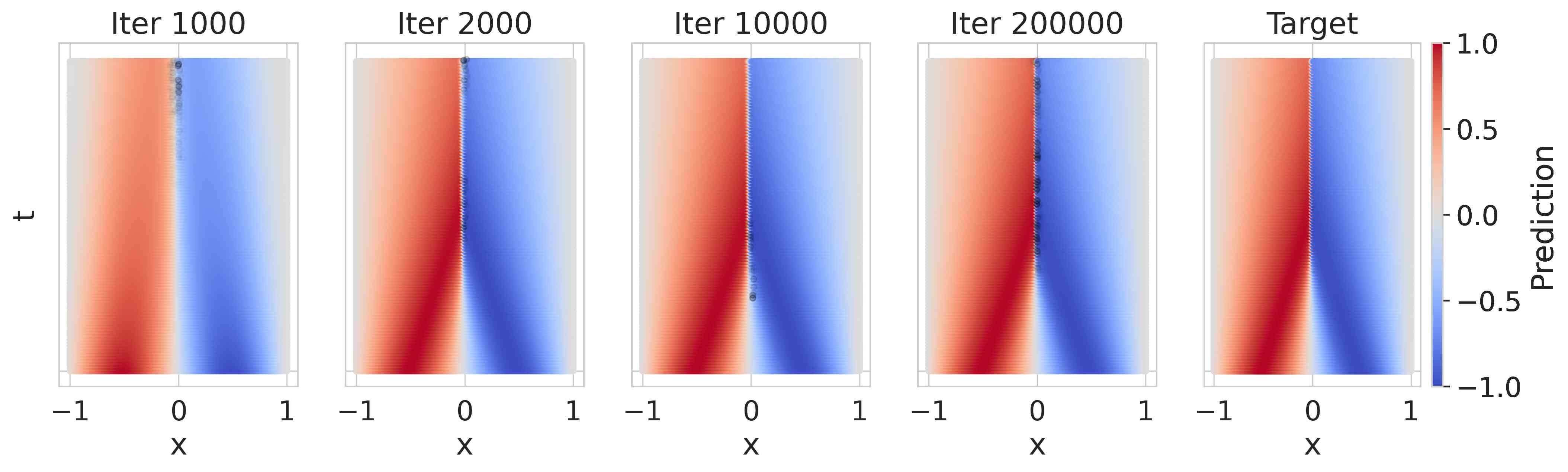} \\
    \end{tabular}
    \caption{\textbf{Burgers' \emph{(Adding)}}: Training point selection dynamics throughout training across respective scoring methods, averaged over 10 runs. Black points represent sampling probability determined by \cref{eq:res-pmf} with $\alpha=2$ and $c=0$. High transparency indicates low sampling probability. Values are clipped to $[-1,1]$.}
    \label{fig:burgers_adding}
\end{figure}

\begin{figure}
    \centering
    \begin{tabular}{>{\centering\arraybackslash}m{\methodcolwidth} >{\centering\arraybackslash}p{0.9\textwidth - \methodcolwidth - 0.5cm}}
        PINN\-fluence&  \includegraphics[align=c,trim={0 0 0 0},clip,width=\linewidth]{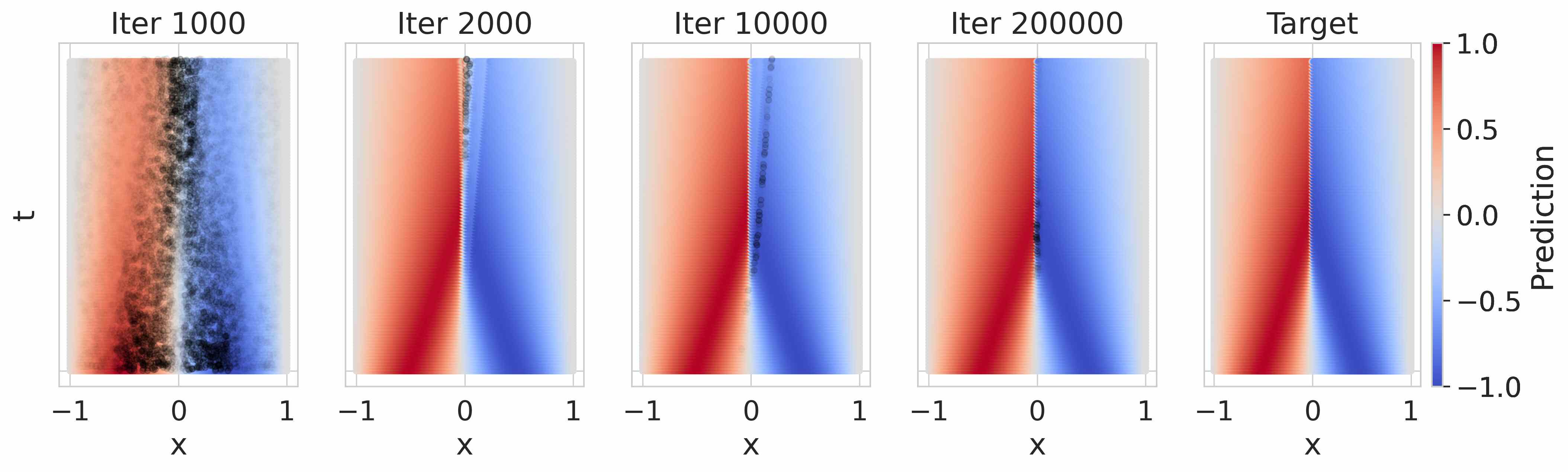} \\
        \addlinespace
        RAR&  \includegraphics[align=c,trim={0 0 0 0},clip,width=\linewidth]{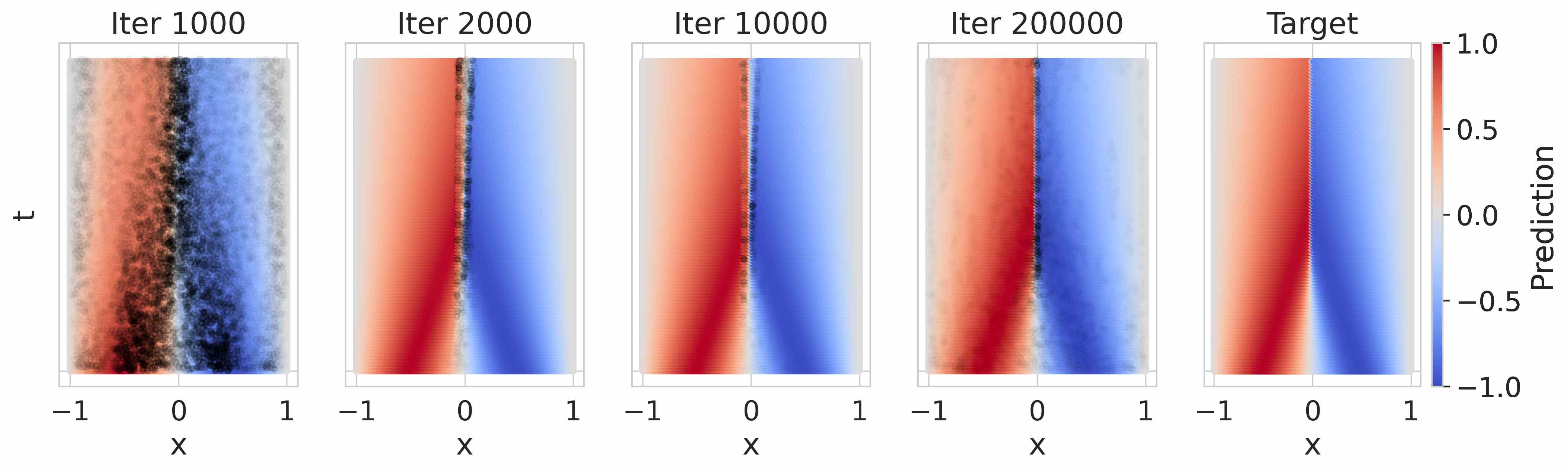} \\
        \addlinespace
        Grad-Dot&  \includegraphics[align=c,trim={0 0 0 0},clip,width=\linewidth]{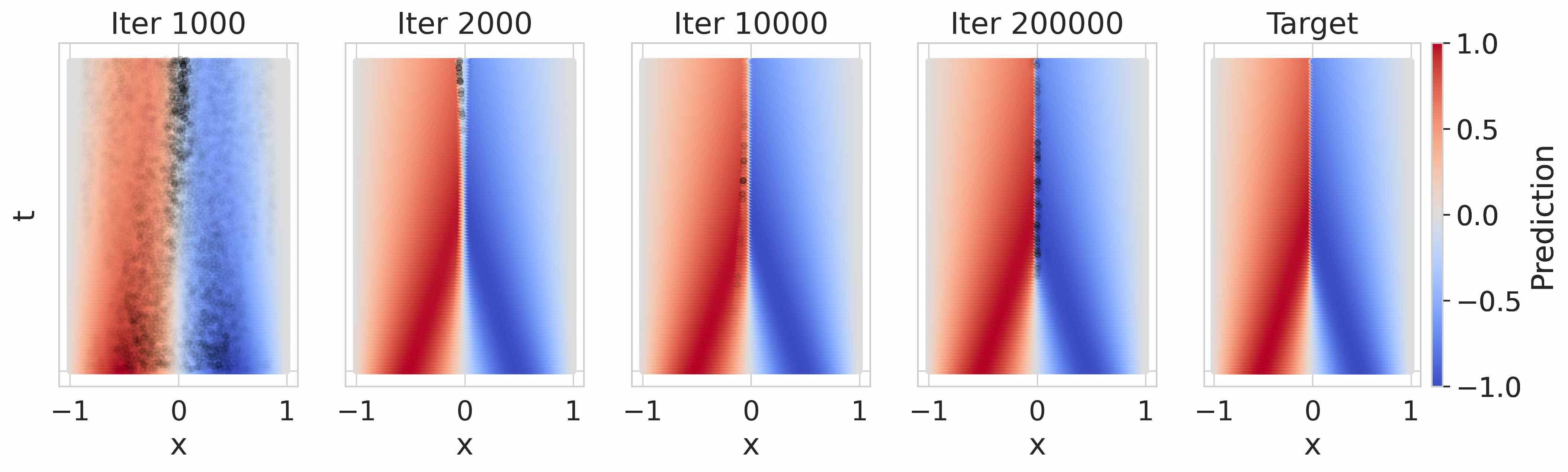} \\
        \addlinespace
        Output-Grad&  \includegraphics[align=c,trim={0 0 0 0},clip,width=\linewidth]{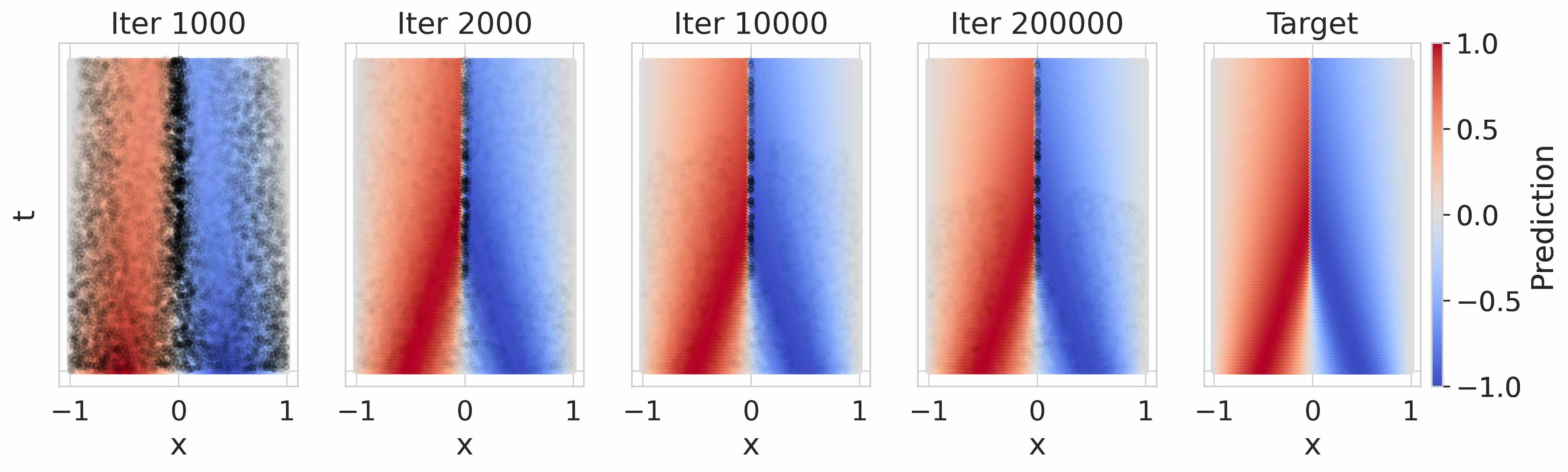} \\
        \addlinespace
        Loss-Grad&  \includegraphics[align=c,trim={0 0 0 0},clip,width=\linewidth]{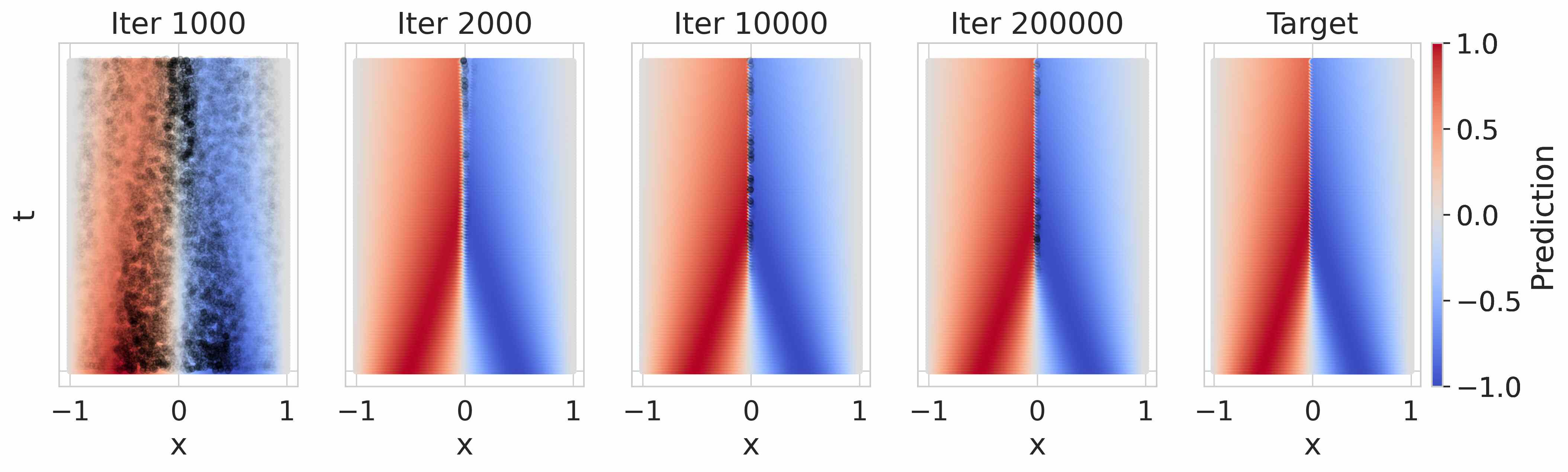} \\
    \end{tabular}
    \caption{\textbf{Burgers' \emph{(Replacing)}}: Training point selection dynamics throughout training across respective scoring methods, averaged over 10 runs. Black points represent sampling probability determined by \cref{eq:res-pmf} with $\alpha=1$ and $c=1$. High transparency indicates low sampling probability. Values are clipped to $[-1,1]$.}
    \label{fig:burgers_replacing}
\end{figure}

\begin{figure}
    \centering
    \begin{tabular}{>{\centering\arraybackslash}m{\methodcolwidth} >{\centering\arraybackslash}p{0.9\textwidth - \methodcolwidth - 0.5cm}}
        PINN\-fluence&  \includegraphics[align=c,trim={0 0 0 0},clip,width=\linewidth]{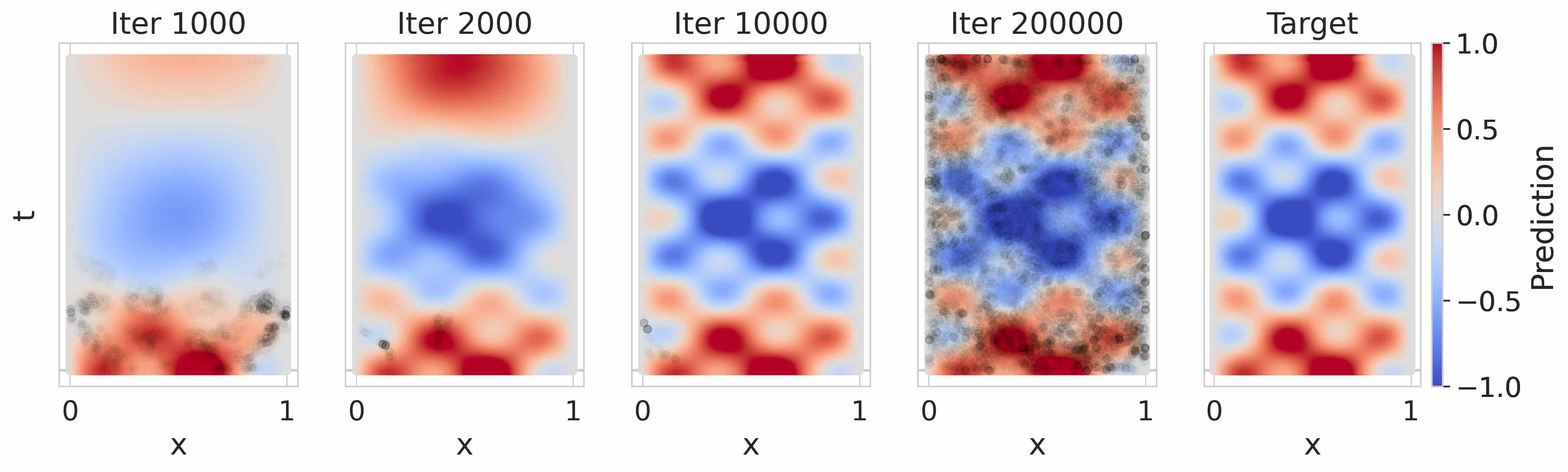} \\
        \addlinespace
        RAR&  \includegraphics[align=c,trim={0 0 0 0},clip,width=\linewidth]{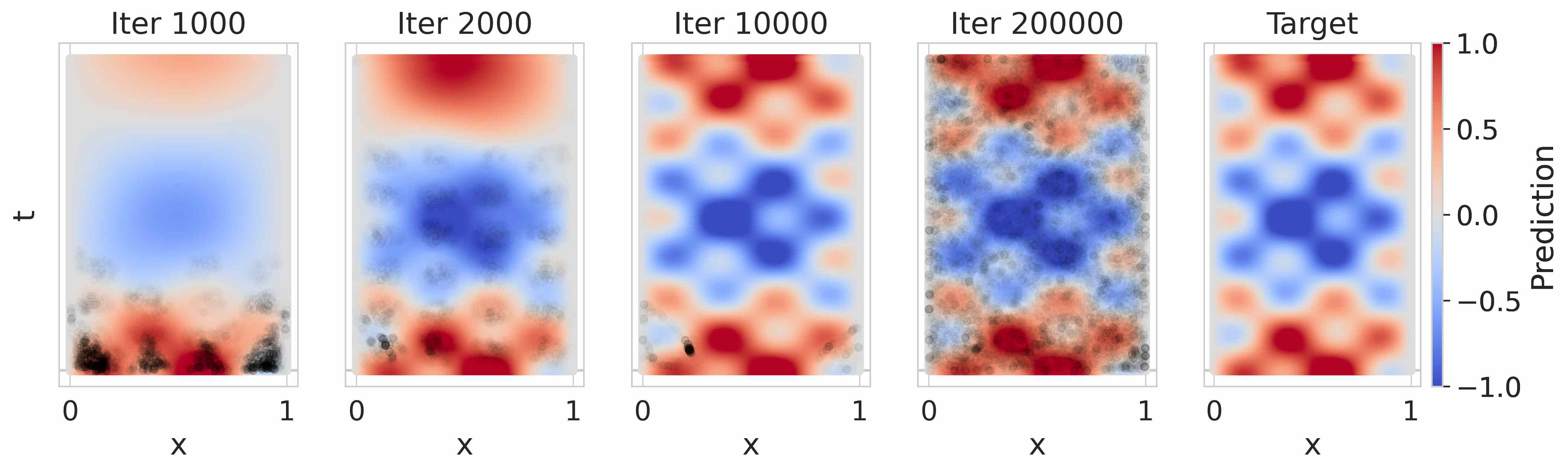} \\
        \addlinespace
        Grad-Dot&  \includegraphics[align=c,trim={0 0 0 0},clip,width=\linewidth]{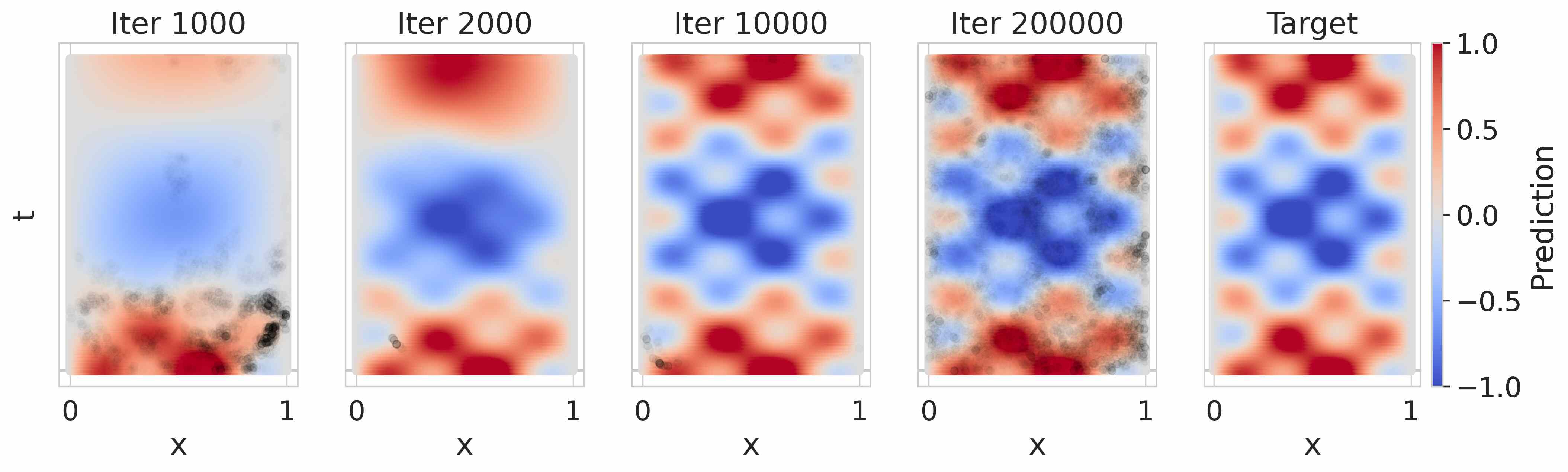} \\
        \addlinespace
        Output-Grad&  \includegraphics[align=c,trim={0 0 0 0},clip,width=\linewidth]{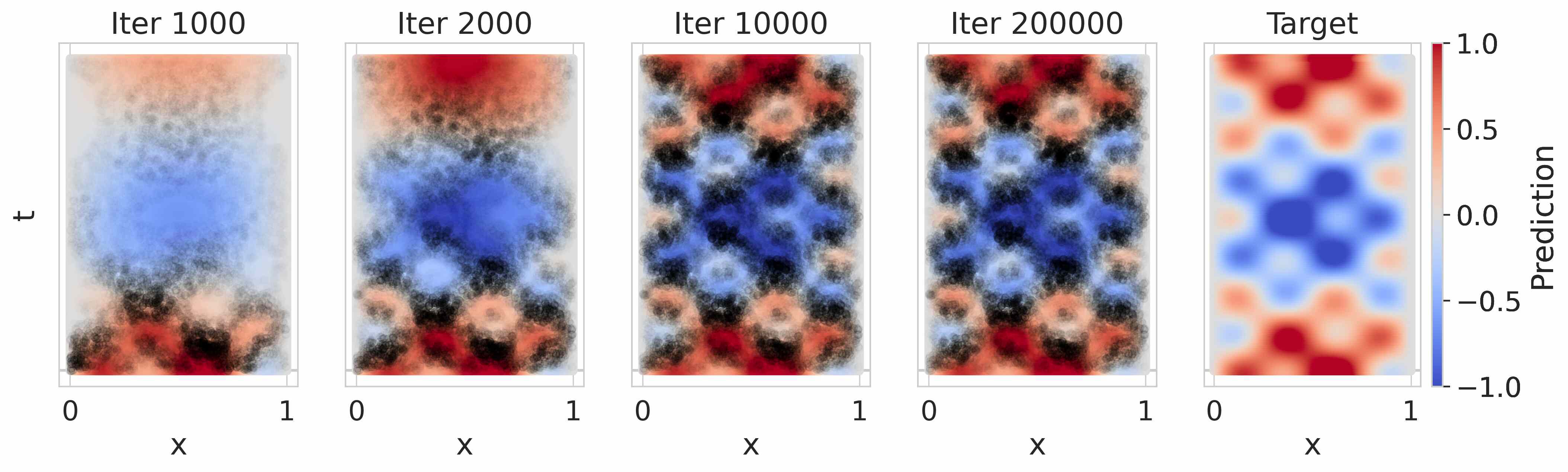} \\
        \addlinespace
        Loss-Grad&  \includegraphics[align=c,trim={0 0 0 0},clip,width=\linewidth]{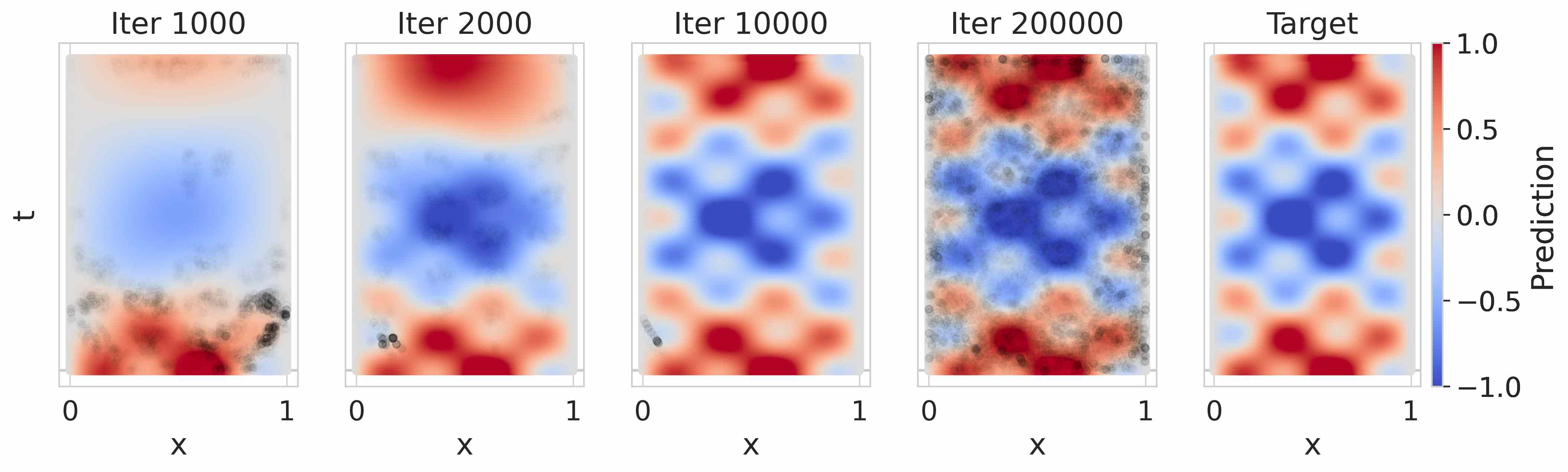} \\
    \end{tabular}
    \caption{\textbf{Wave \emph{(Adding)}}: Training point selection dynamics throughout training across respective scoring methods, averaged over 10 runs. Black points represent sampling probability determined by \cref{eq:res-pmf} with $\alpha=2$ and $c=0$. High transparency indicates low sampling probability. Values are clipped to $[-1,1]$.}
    \label{fig:wave_adding}
\end{figure}

\begin{figure}
    \centering
    \begin{tabular}{>{\centering\arraybackslash}m{\methodcolwidth} >{\centering\arraybackslash}p{0.9\textwidth - \methodcolwidth - 0.5cm}}
        PINN\-fluence&  \includegraphics[align=c,trim={0 0 0 0},clip,width=\linewidth]{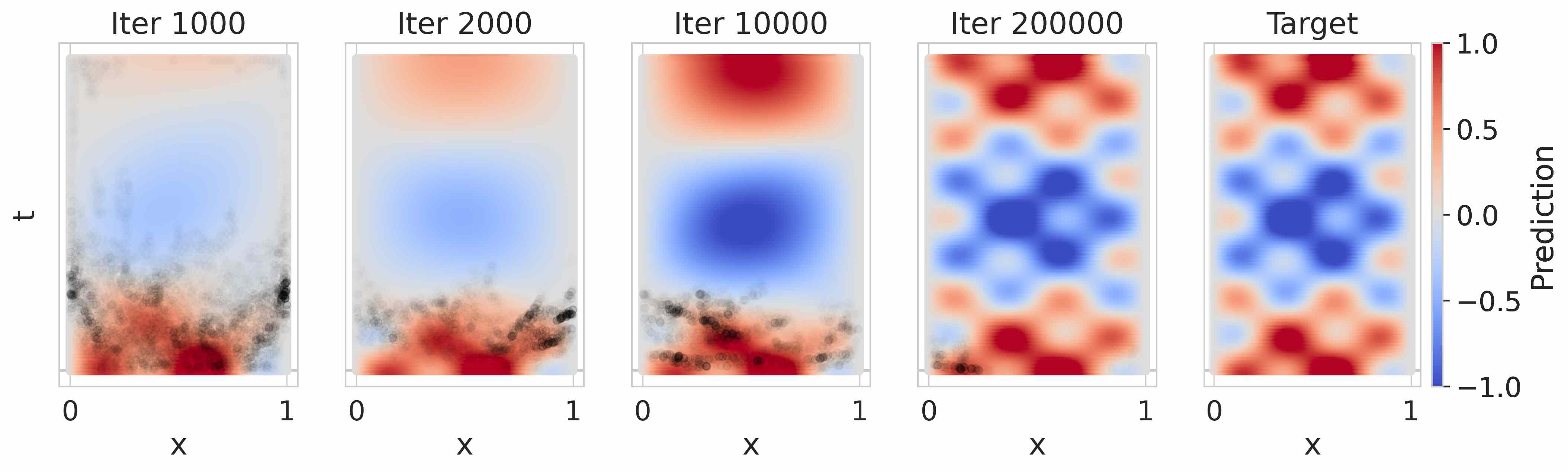} \\
        \addlinespace
        RAR&  \includegraphics[align=c,trim={0 0 0 0},clip,width=\linewidth]{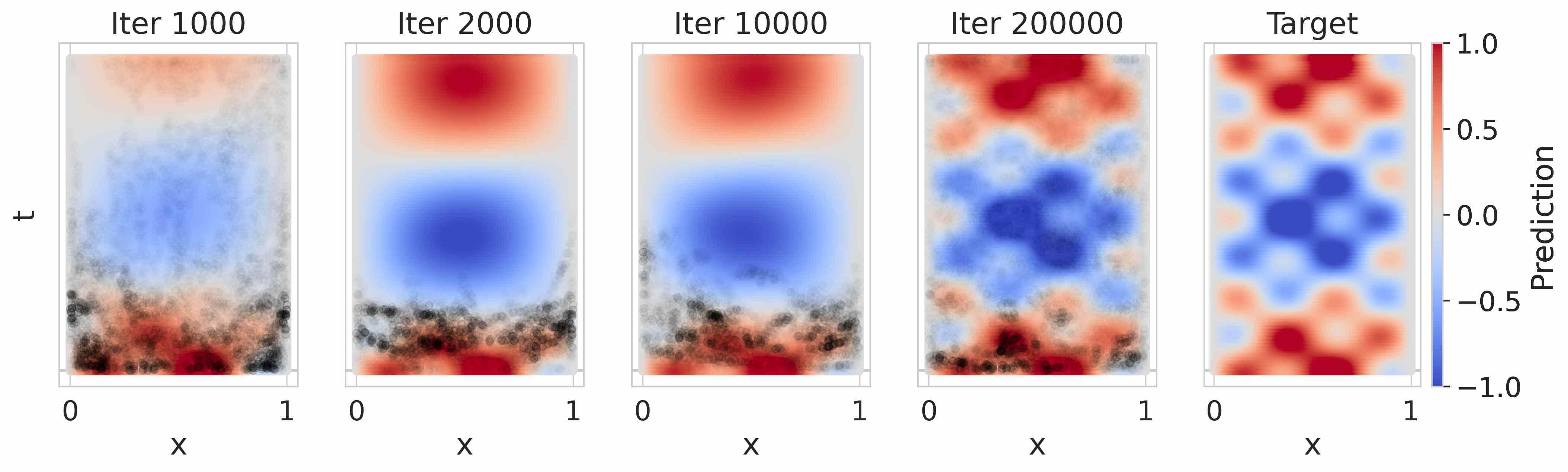} \\
        \addlinespace
        Grad-Dot&  \includegraphics[align=c,trim={0 0 0 0},clip,width=\linewidth]{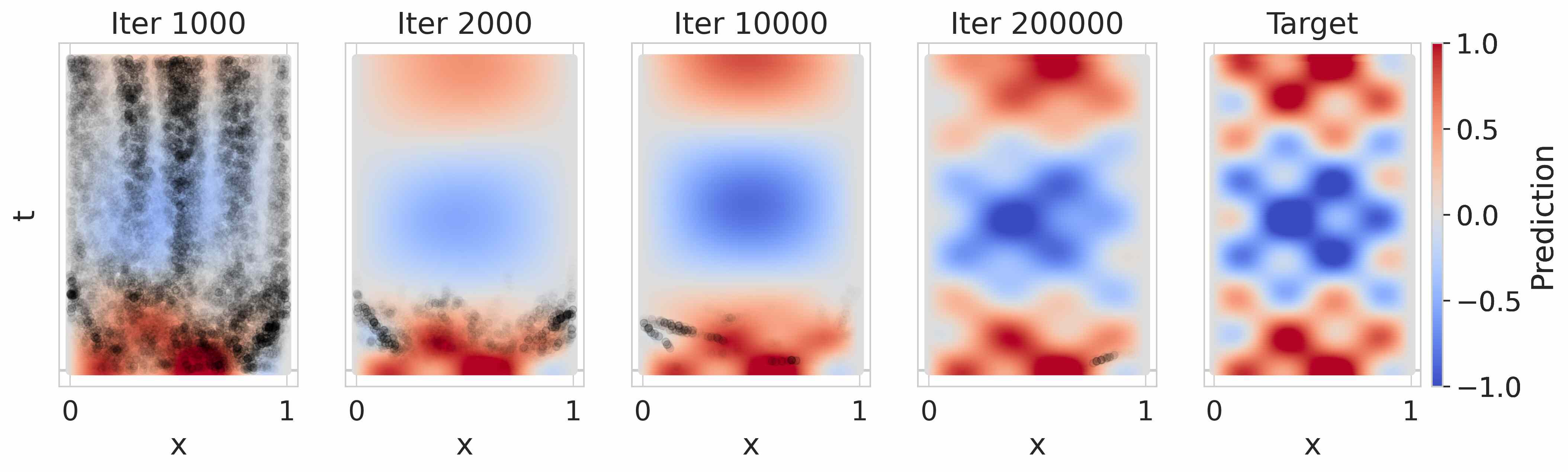} \\
        \addlinespace
        Output-Grad&  \includegraphics[align=c,trim={0 0 0 0},clip,width=\linewidth]{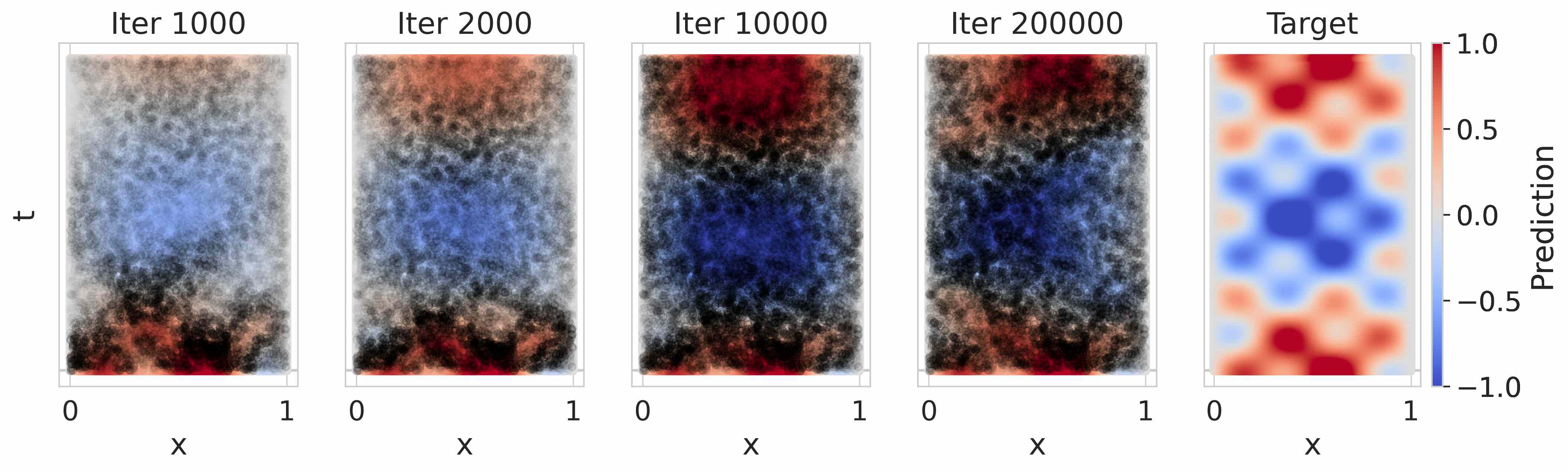} \\
        \addlinespace
        Loss-Grad&  \includegraphics[align=c,trim={0 0 0 0},clip,width=\linewidth]{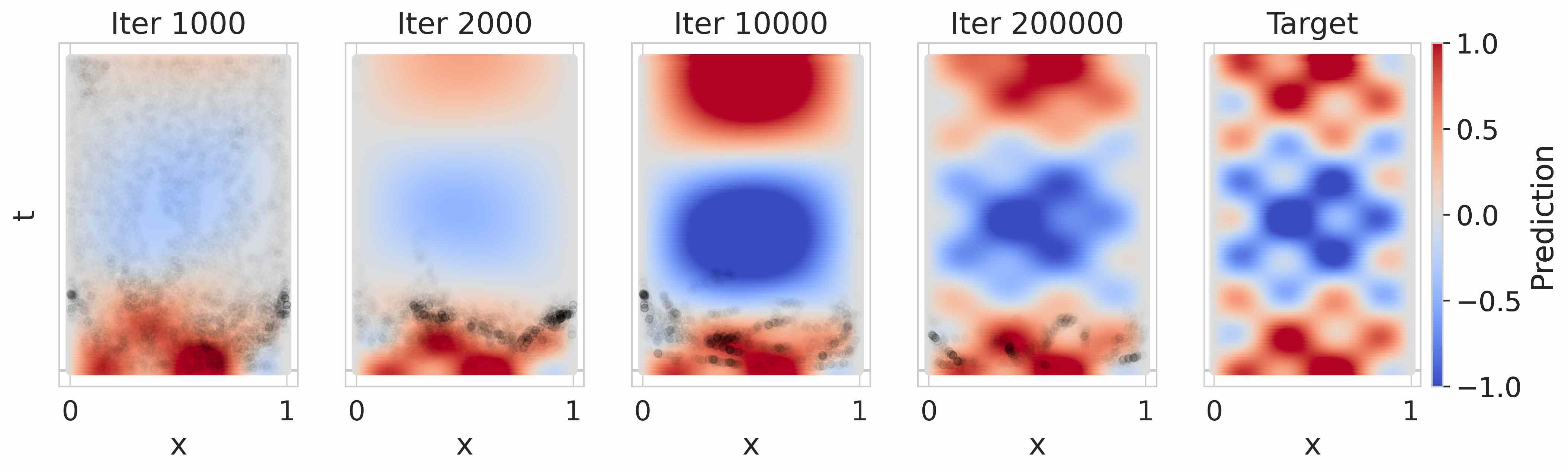} \\
    \end{tabular}
    \caption{\textbf{Wave \emph{(Replacing)}}: Training point selection dynamics throughout training across respective scoring methods, averaged over 10 runs. Black points represent sampling probability determined by \cref{eq:res-pmf} with $\alpha=1$ and $c=1$. High transparency indicates low sampling probability. Values are clipped to $[-1,1]$.}
    \label{fig:wave_replacing}
\end{figure}

\begin{figure}
    \centering
    \begin{tabular}{>{\centering\arraybackslash}m{\methodcolwidth} >{\centering\arraybackslash}p{0.9\textwidth - \methodcolwidth - 0.5cm}}
        PINN\-fluence&  \includegraphics[align=c,trim={0 0 0 0},clip,width=\linewidth]{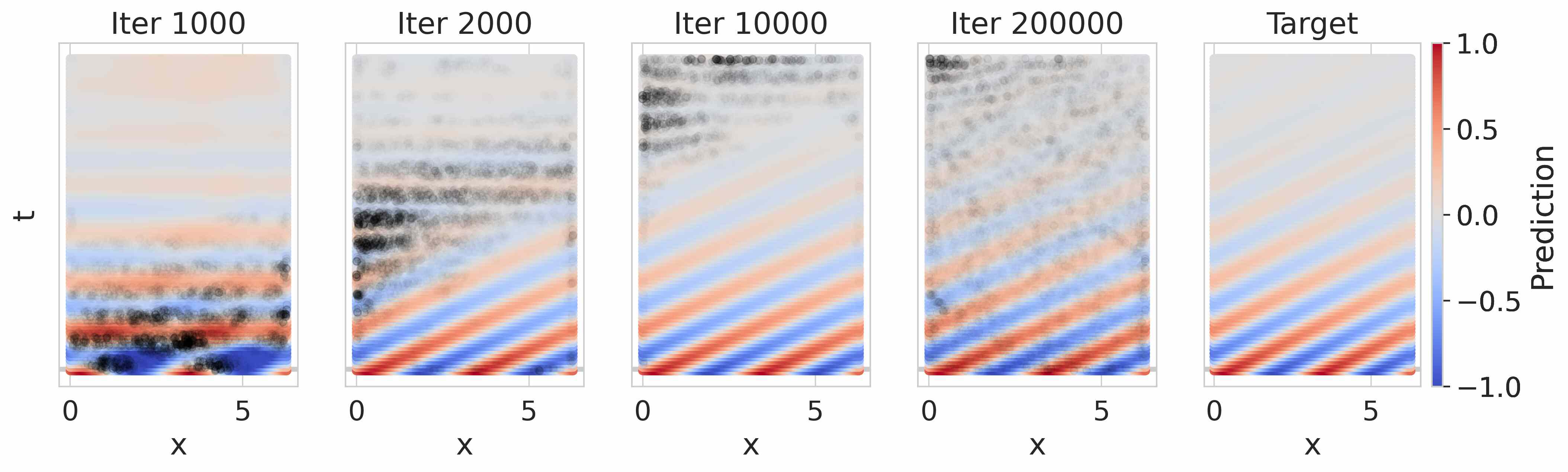} \\
        \addlinespace 
        RAR&  \includegraphics[align=c,trim={0 0 0 0},clip,width=\linewidth]{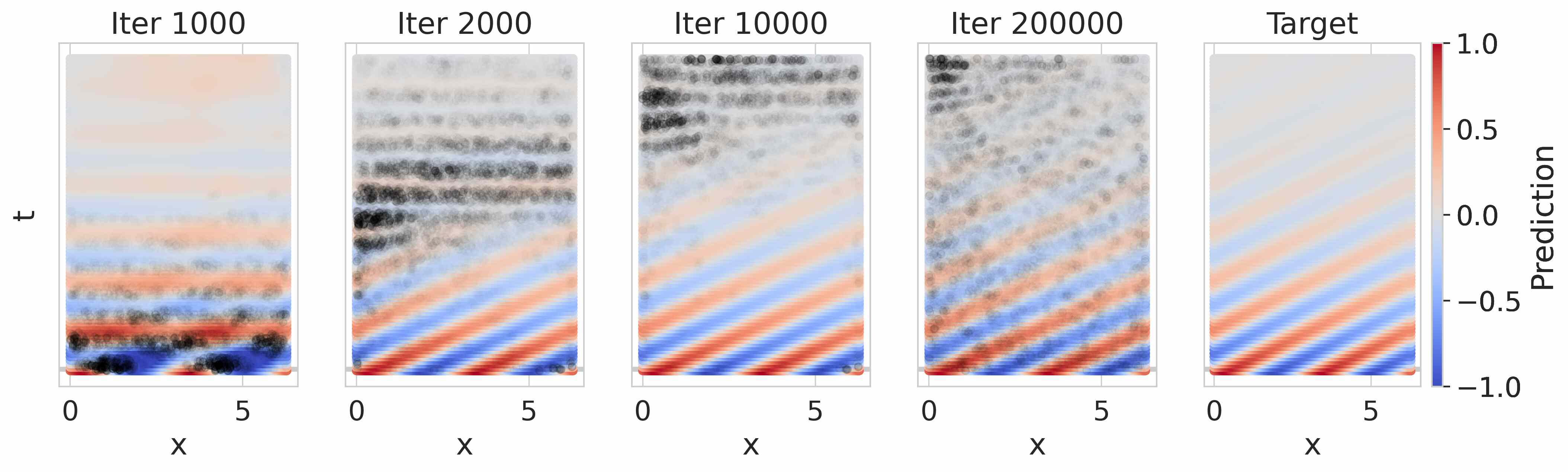} \\
        \addlinespace
        Grad-Dot&  \includegraphics[align=c,trim={0 0 0 0},clip,width=\linewidth]{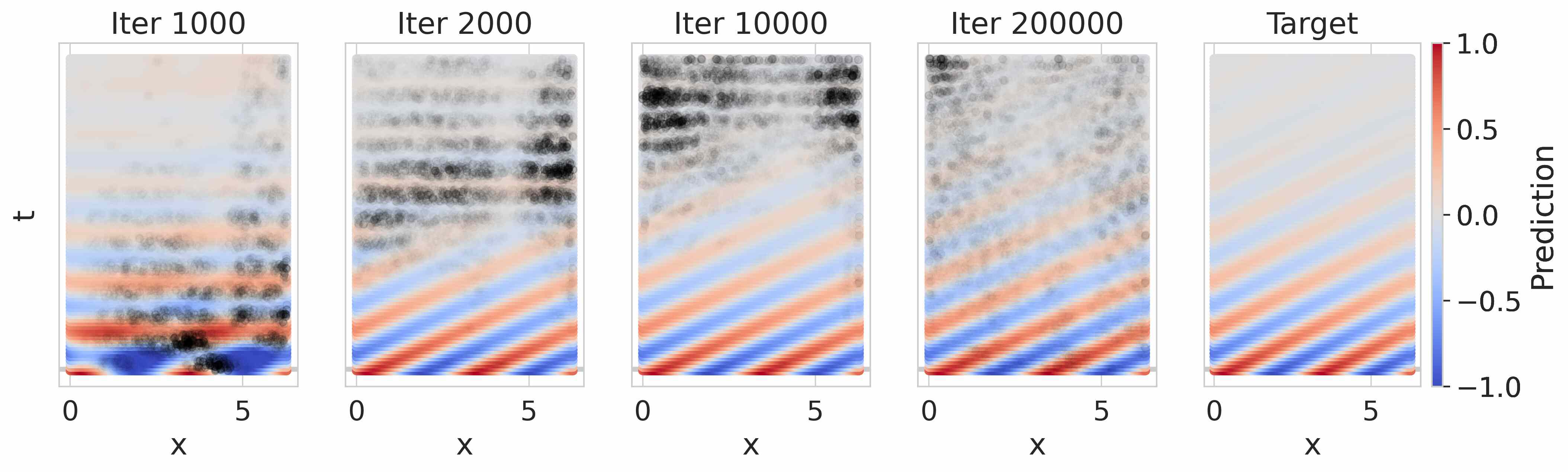} \\
        \addlinespace
        Output-Grad&  \includegraphics[align=c,trim={0 0 0 0},clip,width=\linewidth]{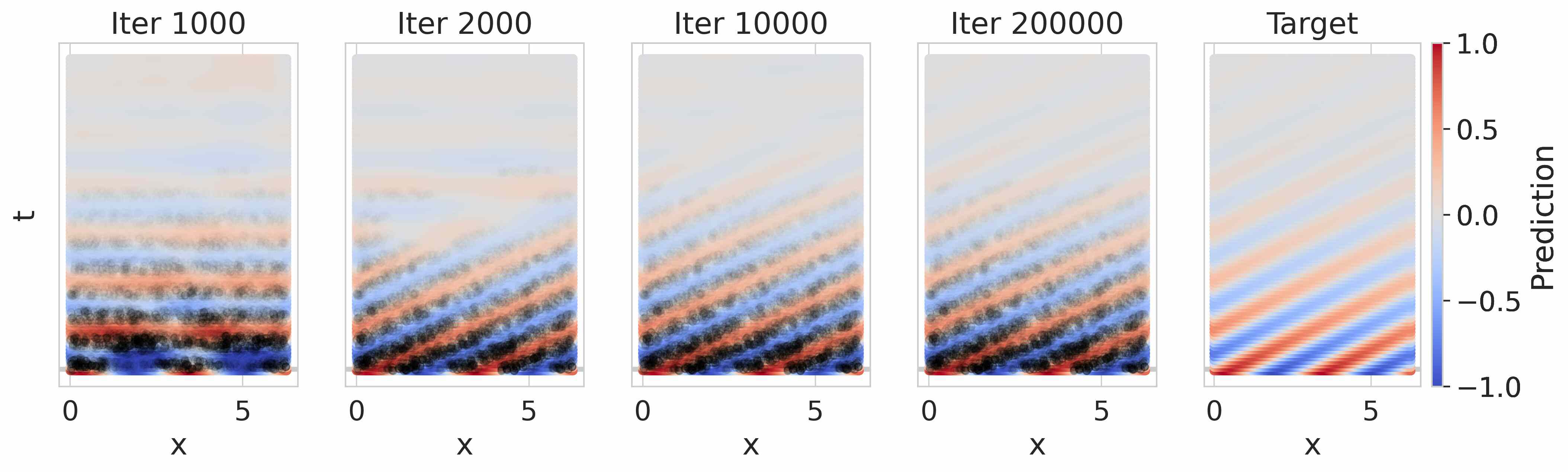} \\
        \addlinespace
        Loss-Grad&  \includegraphics[align=c,trim={0 0 0 0},clip,width=\linewidth]{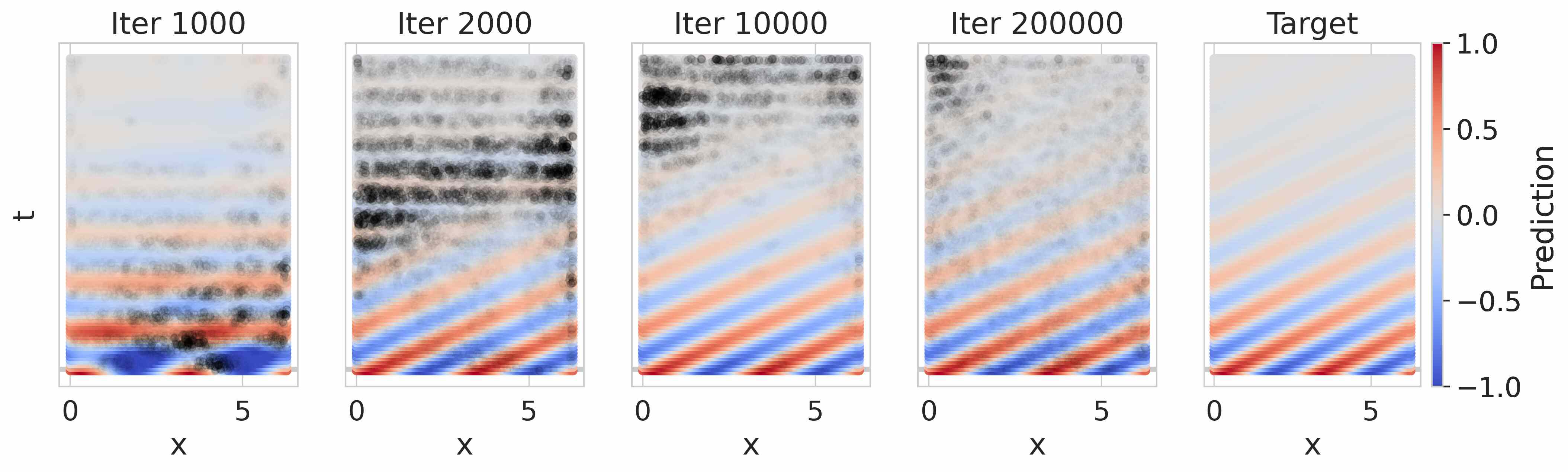} \\
    \end{tabular}
    \caption{\textbf{Drift-Diffusion \emph{(Adding)}}: Training point selection dynamics throughout training across respective scoring methods, averaged over 10 runs. Black points represent sampling probability determined by \cref{eq:res-pmf} with $\alpha=2$ and $c=0$. High transparency indicates low sampling probability. Values are clipped to $[-1,1]$.}
    \label{fig:drift_diffusion_sine_adding}
\end{figure}

\begin{figure}
    \centering
    \begin{tabular}{>{\centering\arraybackslash}m{\methodcolwidth} >{\centering\arraybackslash}p{0.9\textwidth - \methodcolwidth - 0.5cm}}
        PINN\-fluence & \includegraphics[align=c,trim={0 0 0 0},clip,width=\linewidth]{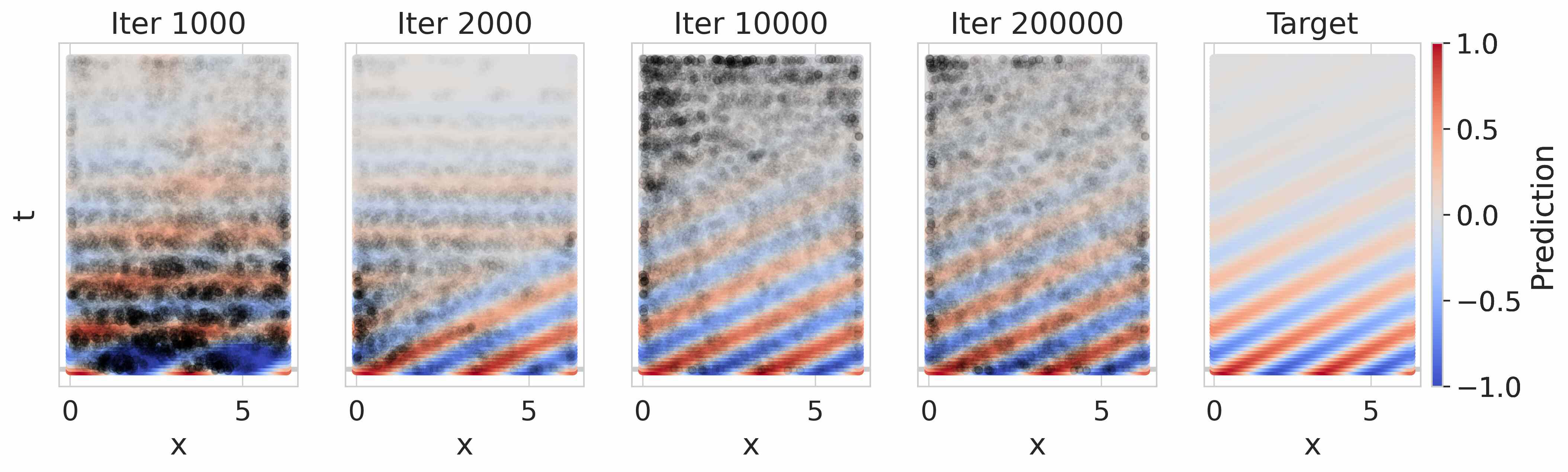} \\ 
        \addlinespace 
        RAR&  \includegraphics[align=c,trim={0 0 0 0},clip,width=\linewidth]{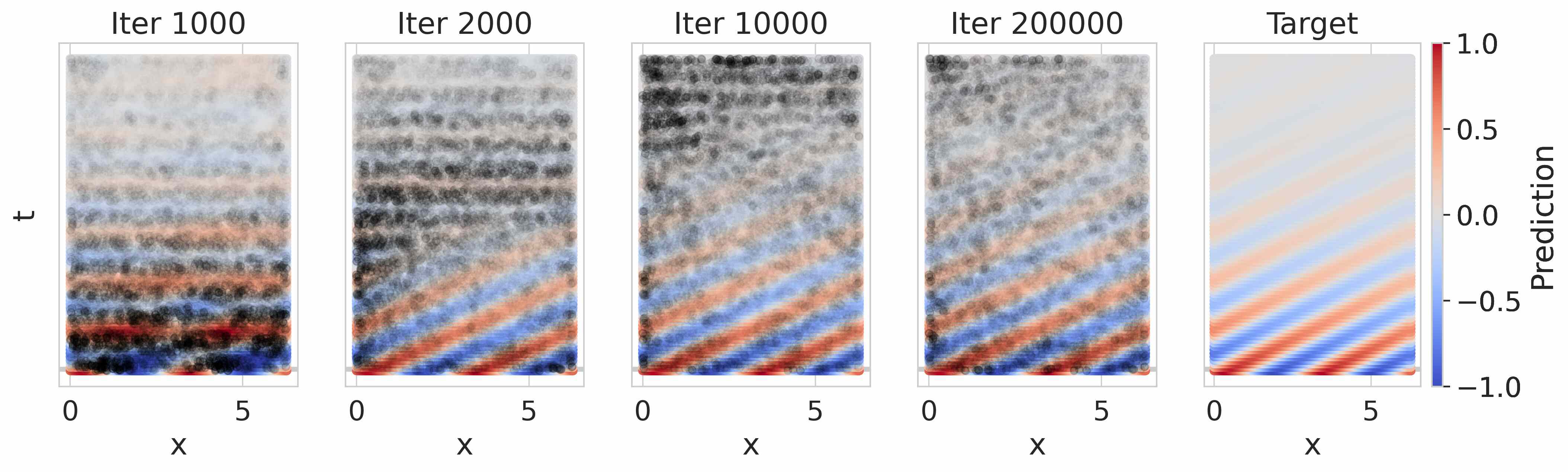} \\
        \addlinespace
        Grad-Dot&  \includegraphics[align=c,trim={0 0 0 0},clip,width=\linewidth]{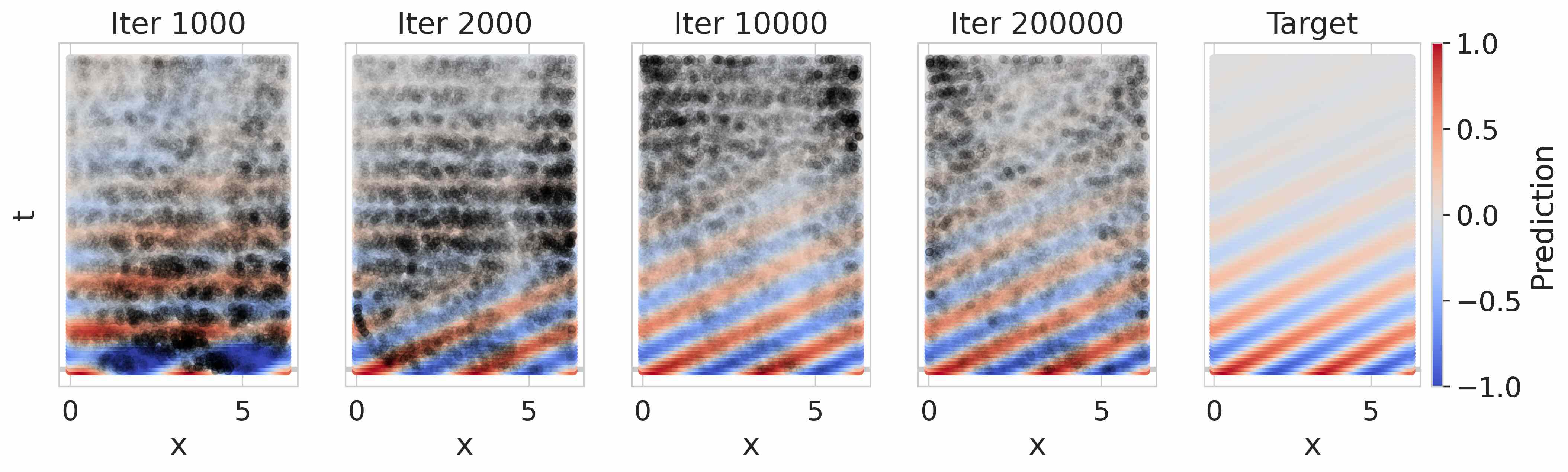} \\
        \addlinespace
        Output-Grad&  \includegraphics[align=c,trim={0 0 0 0},clip,width=\linewidth]{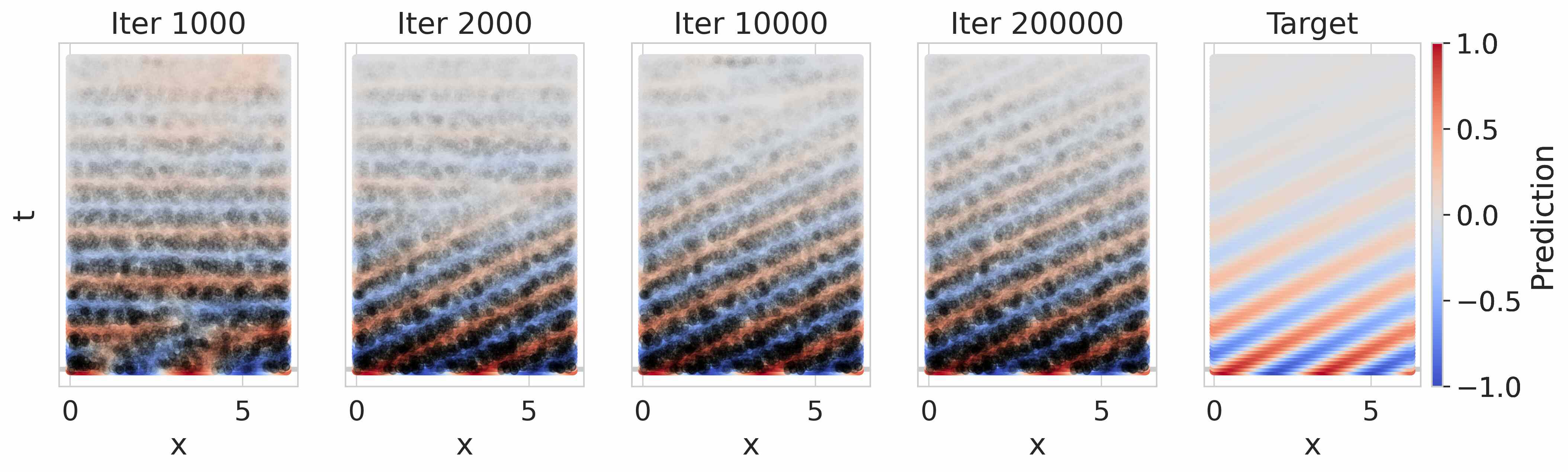} \\
        \addlinespace
        Loss-Grad&  \includegraphics[align=c,trim={0 0 0 0},clip,width=\linewidth]{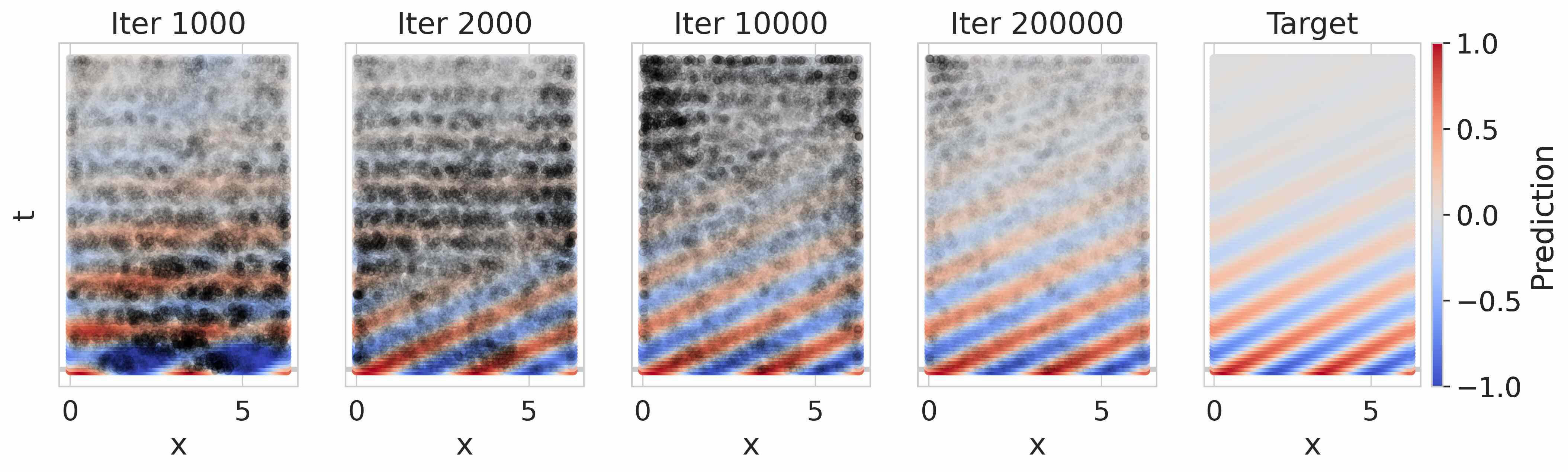} \\
    \end{tabular}
    \caption{\textbf{Drift-Diffusion \emph{(Replacing)}}: Training point selection dynamics throughout training across respective scoring methods, averaged over 10 runs. Black points represent sampling probability determined by \cref{eq:res-pmf} with $\alpha=1$ and $c=1$. High transparency indicates low sampling probability. Values are clipped to $[-1,1]$.}
    \label{fig:drift_diffusion_sine_replacing}
\end{figure}

\clearpage

\subsection{Boxplots}\label{app:boxplots}

\def\mysfigwidth{.455\textwidth}
\def\myfigheight{.18\textheight}
\def\mysubfigskip{\vspace{-6pt}}

\vspace{-10pt}

\begin{figure}[h!]
    \centering

    \begin{subfigure}{\mysfigwidth}
        \centering
        \includegraphics[width=\linewidth]{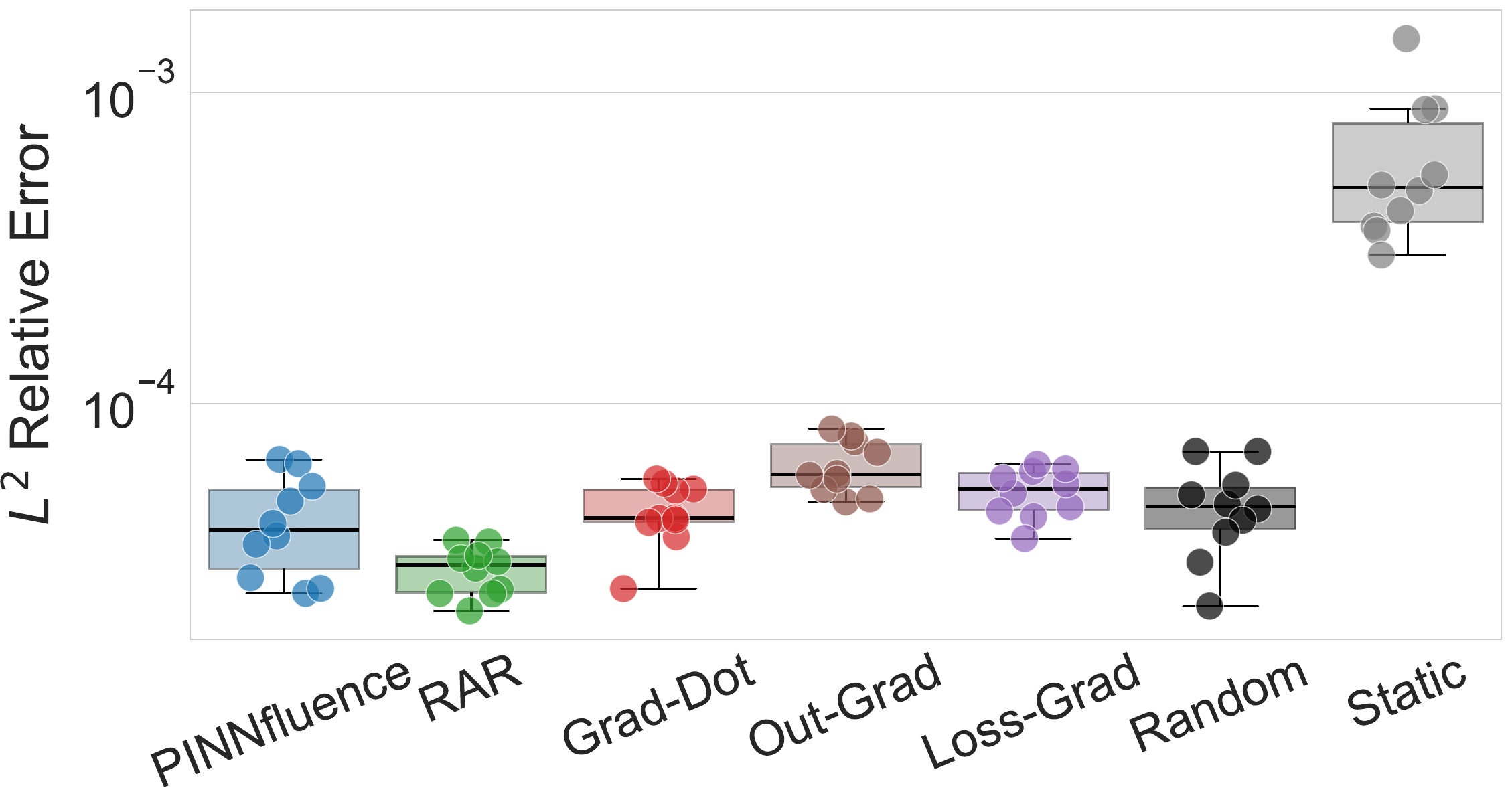}\mysubfigskip
        \caption{Diffusion (\emph{Adding})}
    \end{subfigure}
    \begin{subfigure}{\mysfigwidth}\mysubfigskip
        \centering
        \includegraphics[width=\linewidth]{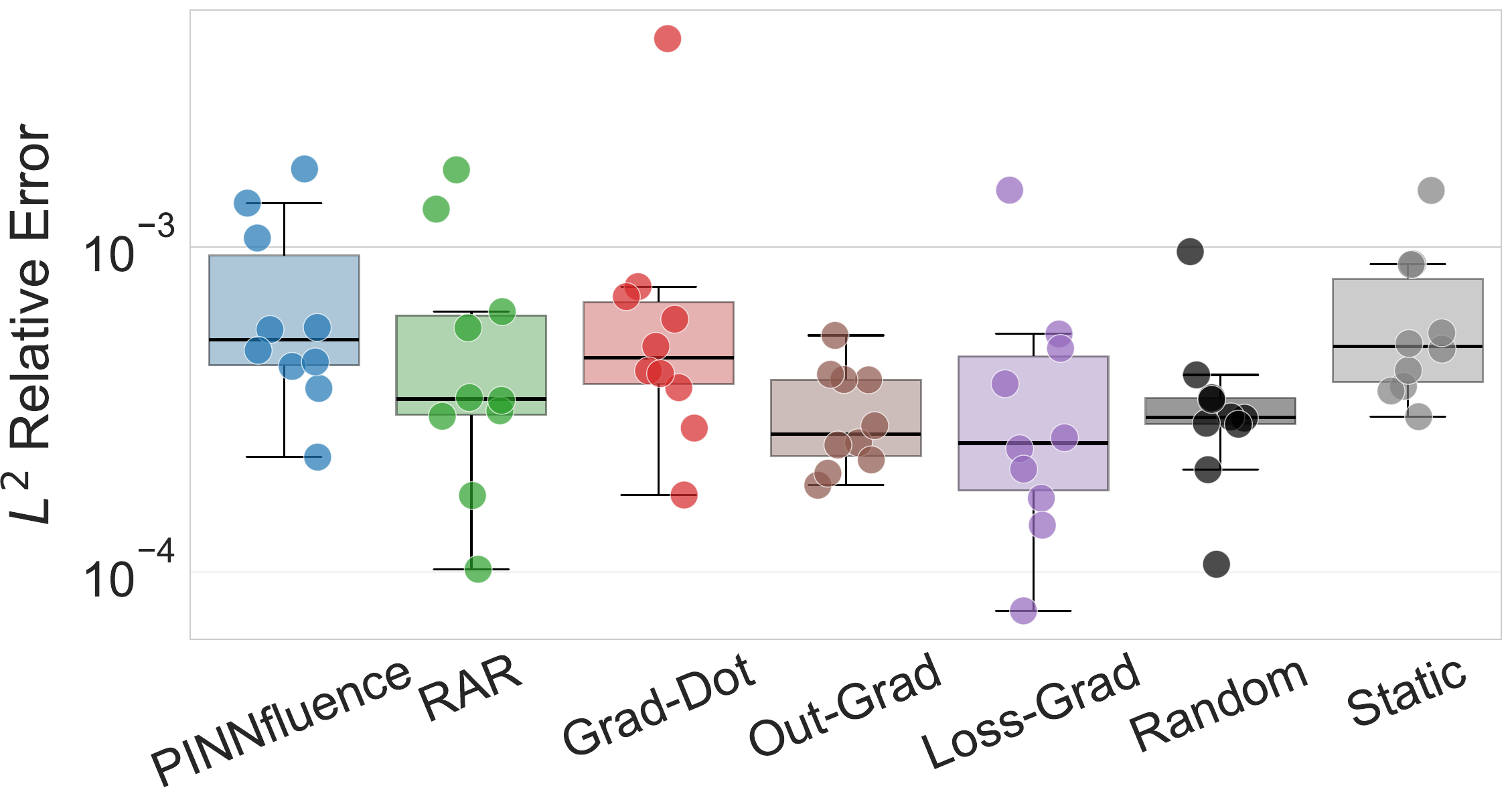}\mysubfigskip
        \caption{Diffusion (\emph{Replacing})}
    \end{subfigure}

    \begin{subfigure}{\mysfigwidth}
        \centering
        \includegraphics[width=\linewidth]{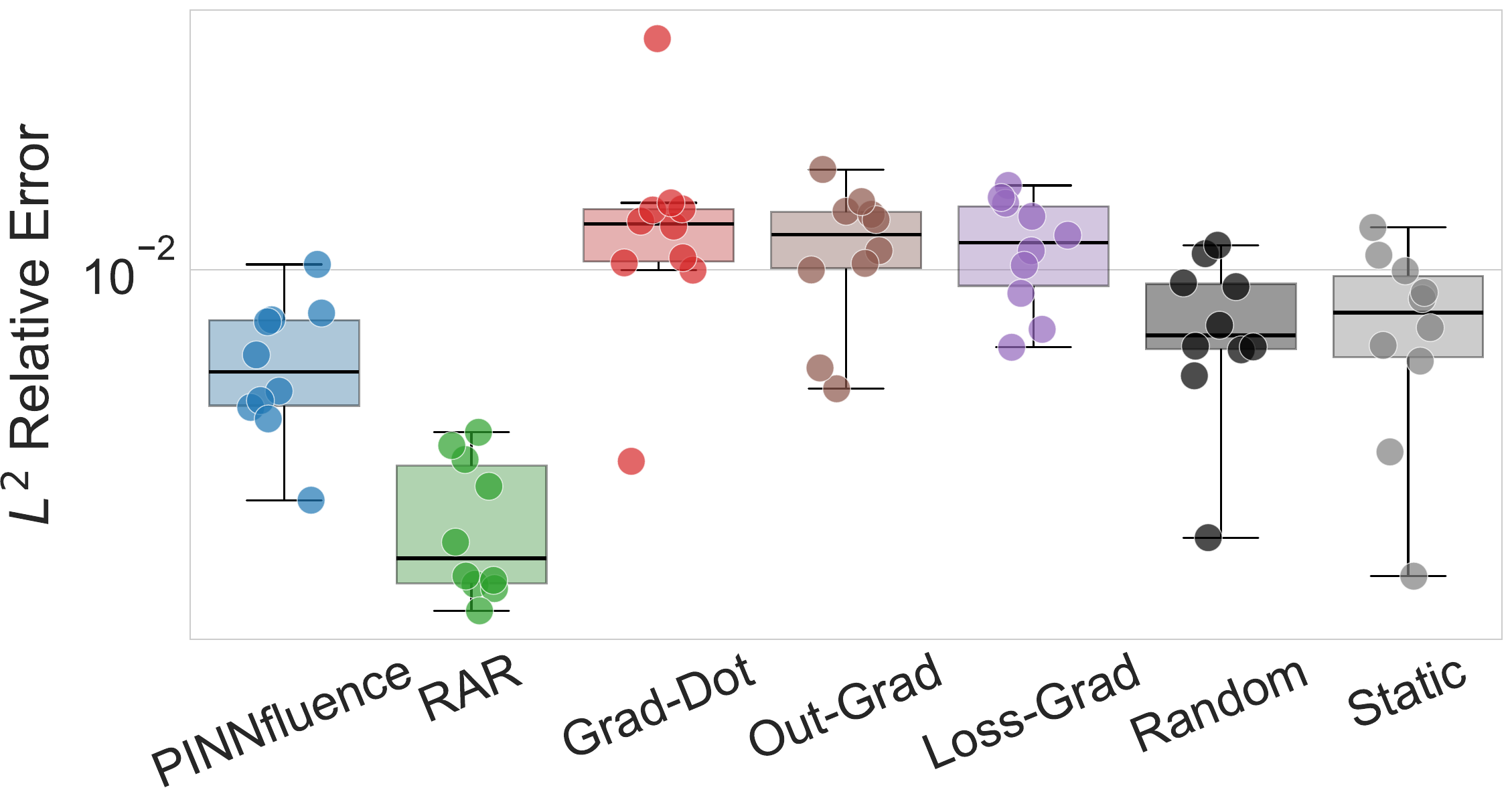}\mysubfigskip
        \caption{Allen-Cahn (\emph{Adding})}
    \end{subfigure}
    \begin{subfigure}{\mysfigwidth}\mysubfigskip
        \centering
        \includegraphics[width=\linewidth]{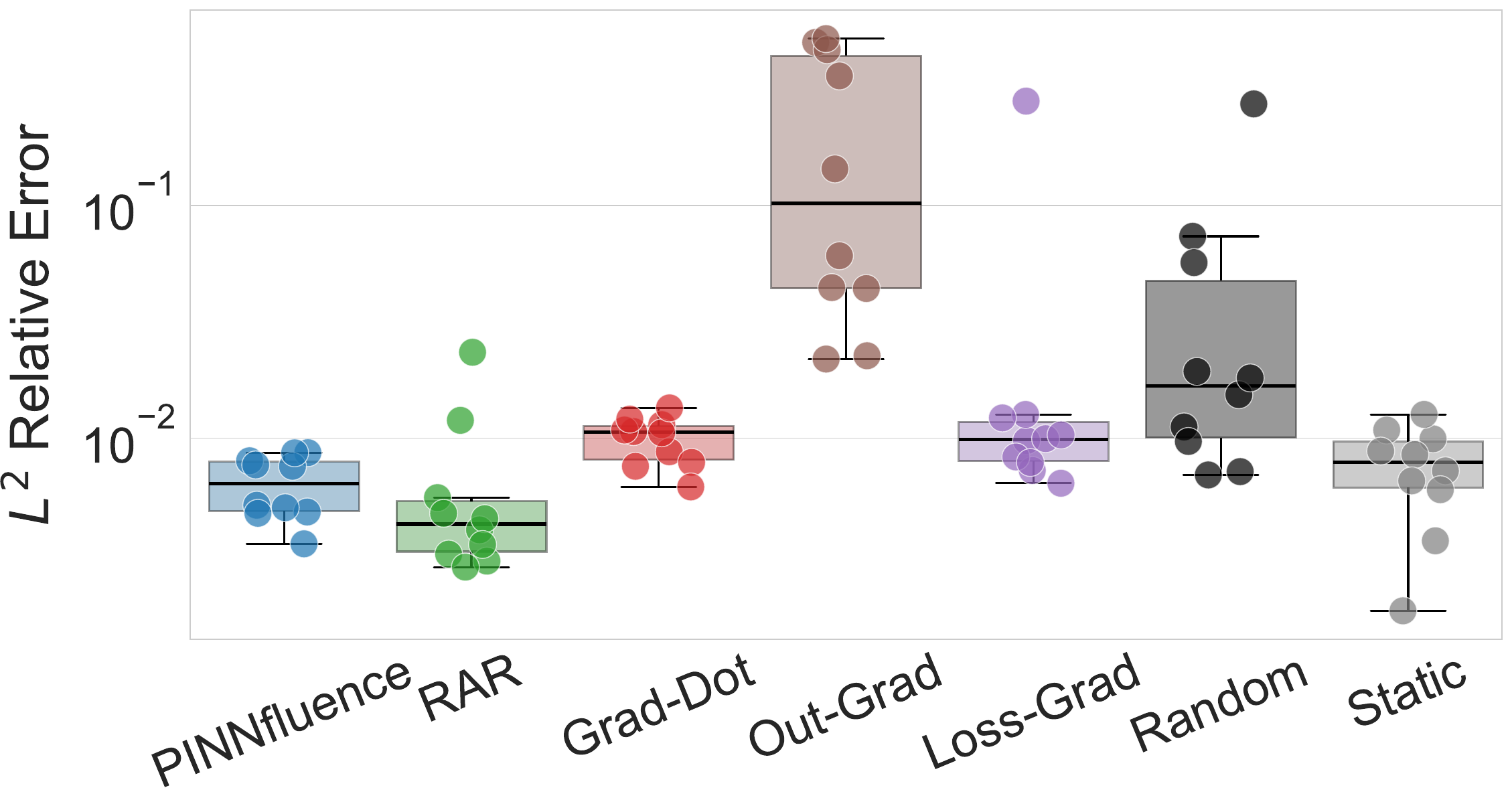}\mysubfigskip
        \caption{Allen-Cahn (\emph{Replacing})}
    \end{subfigure}

    \begin{subfigure}{\mysfigwidth}
        \centering
        \includegraphics[width=\linewidth]{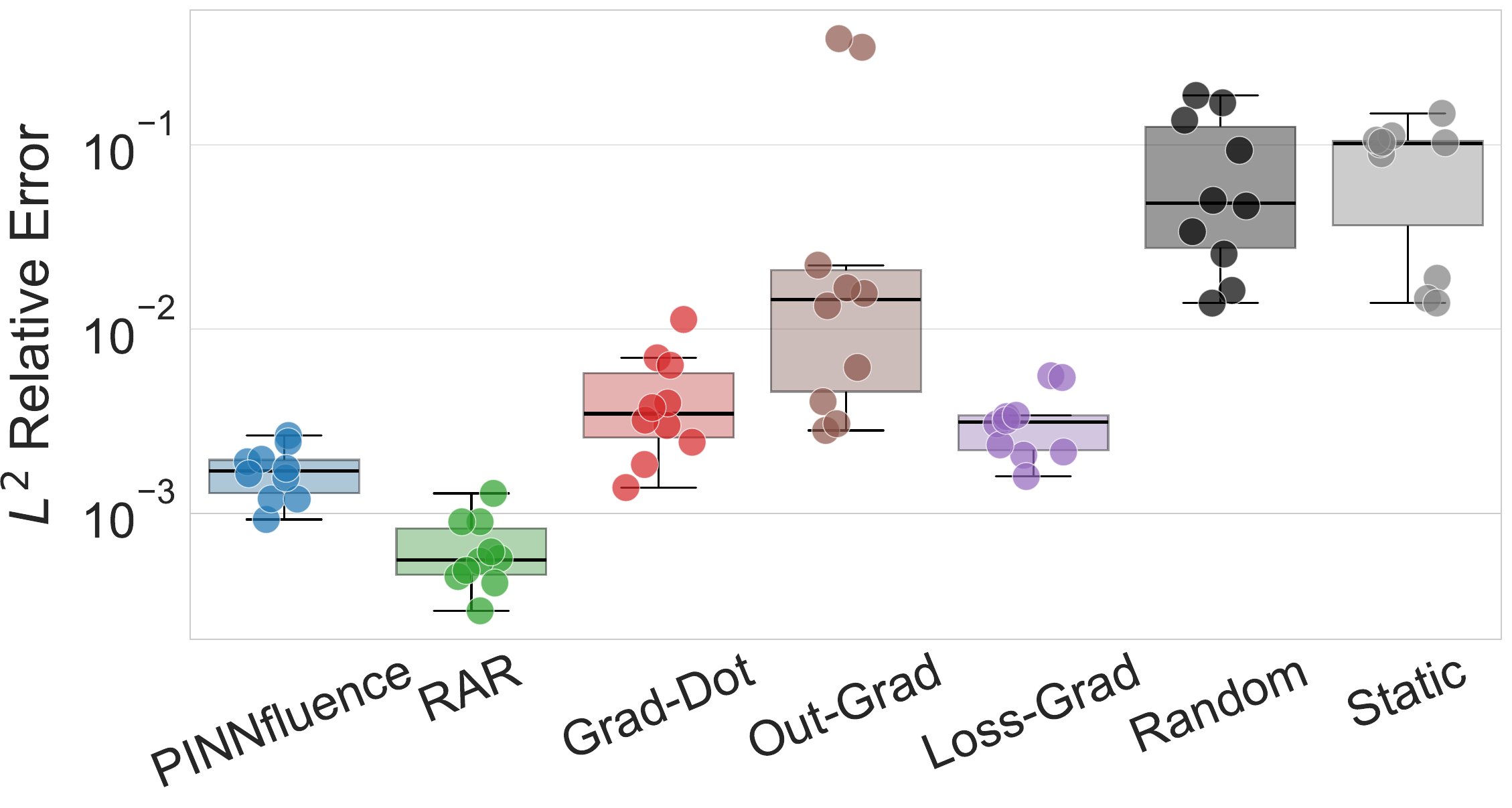}\mysubfigskip
        \caption{Burgers' (\emph{Adding})}
    \end{subfigure}
    \begin{subfigure}{\mysfigwidth}\mysubfigskip
        \centering
        \includegraphics[width=\linewidth]{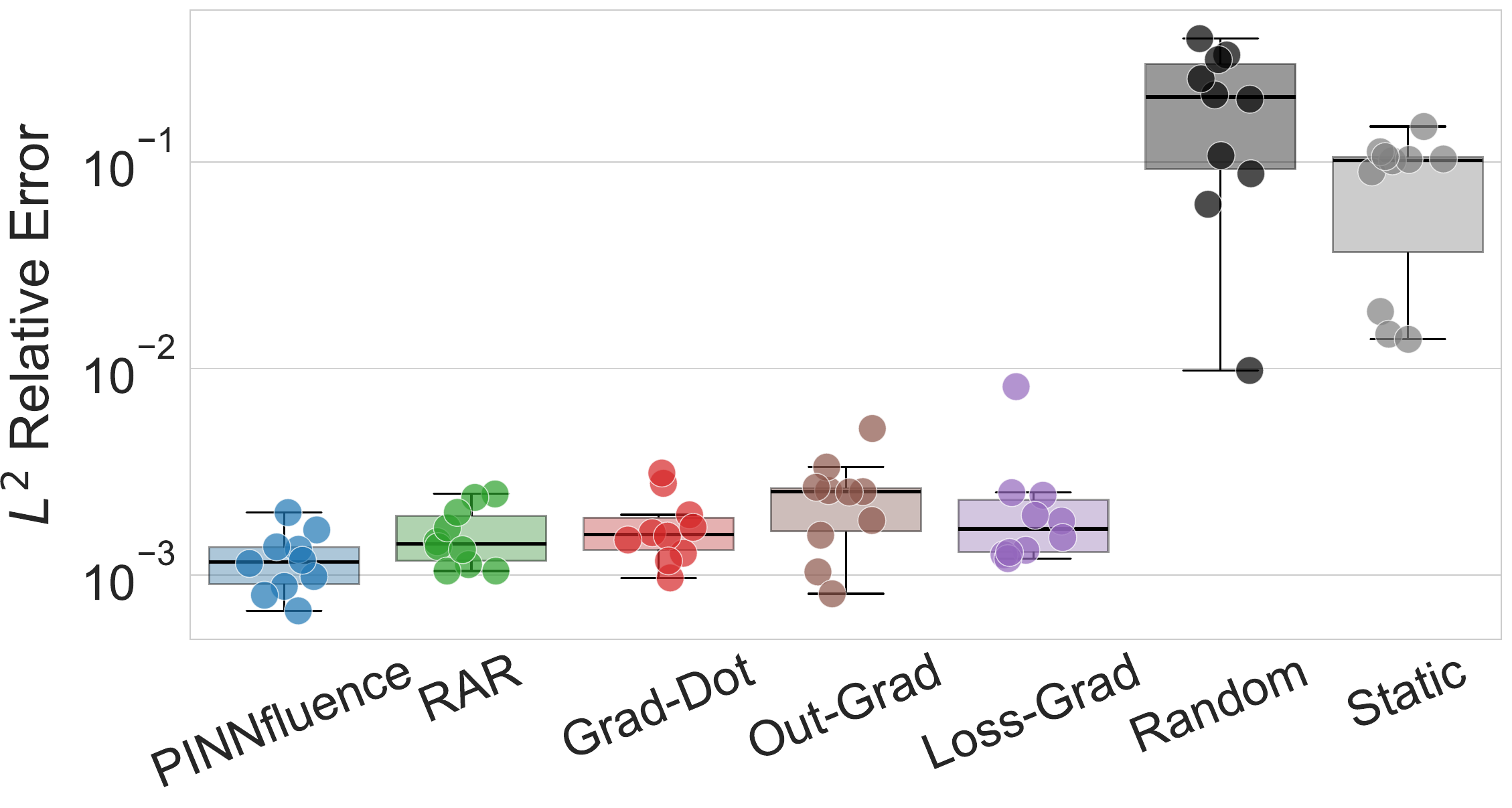}\mysubfigskip
        \caption{Burgers' (\emph{Replacing})}
    \end{subfigure}

    \begin{subfigure}{\mysfigwidth}
        \centering
        \includegraphics[width=\linewidth]{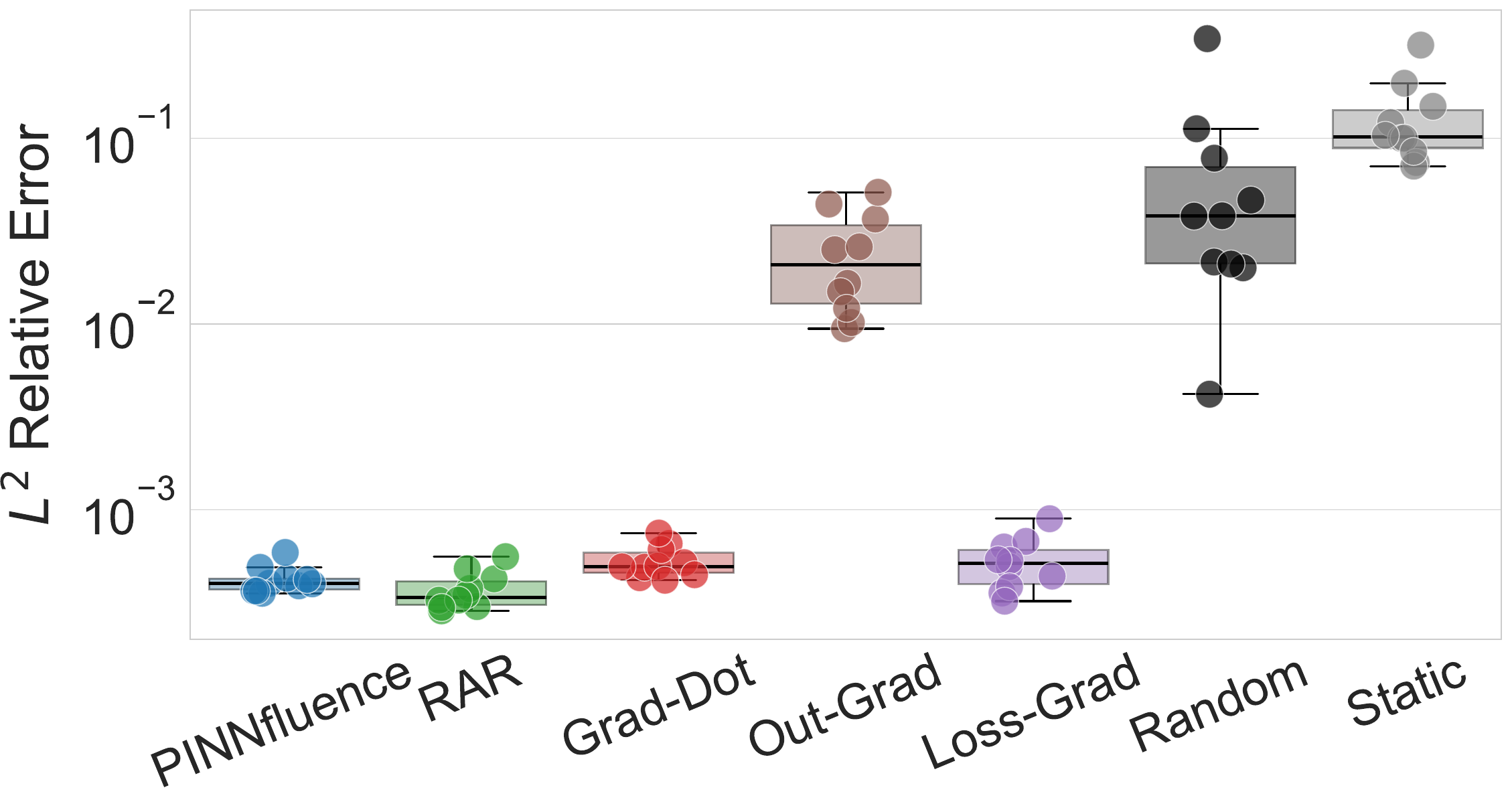}\mysubfigskip
        \caption{Wave (\emph{Adding})}
    \end{subfigure}
    \begin{subfigure}{\mysfigwidth}\mysubfigskip
        \centering
        \includegraphics[width=\linewidth]{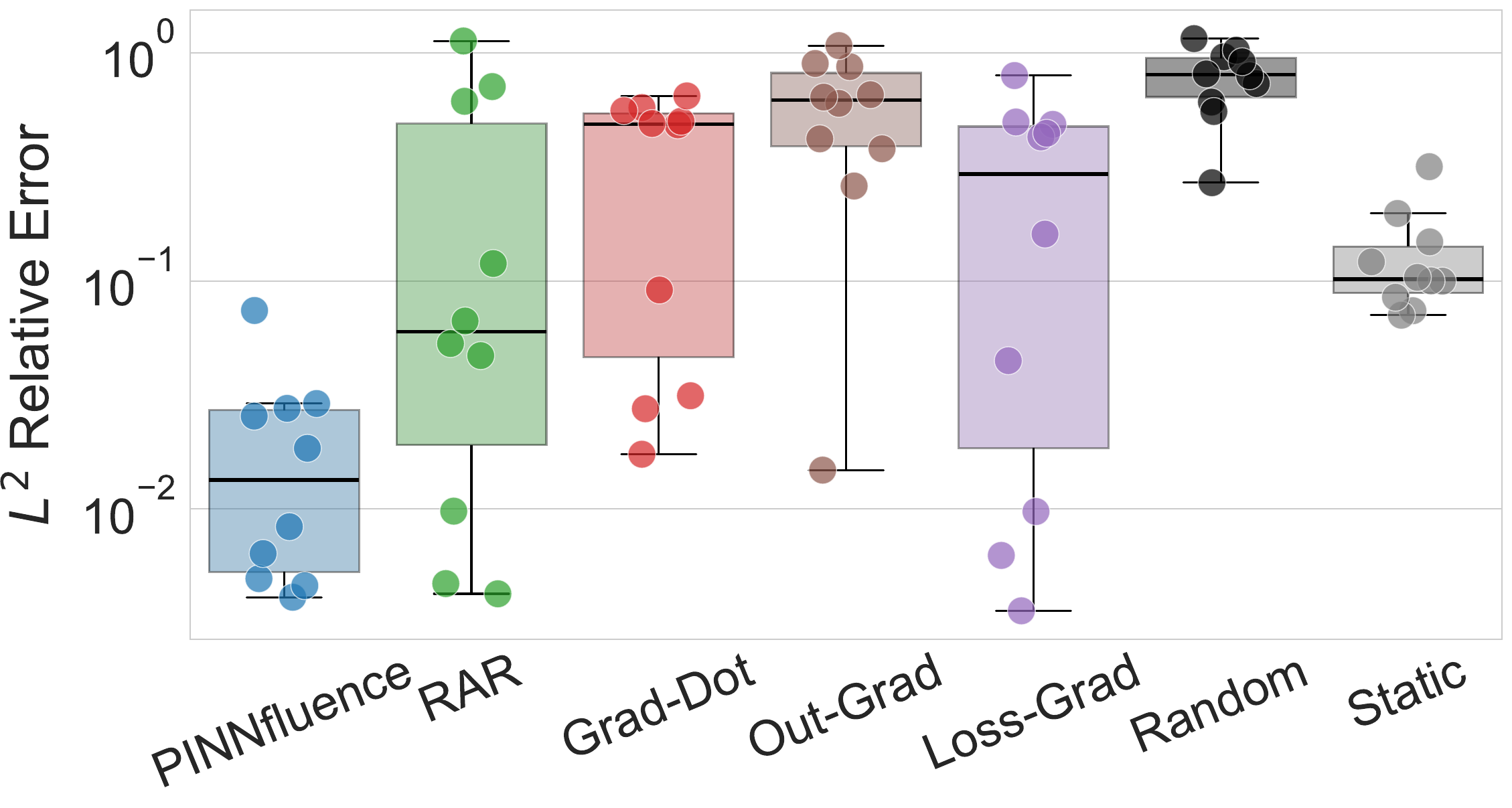}\mysubfigskip
        \caption{Wave (\emph{Replacing})}
    \end{subfigure}

    \begin{subfigure}{\mysfigwidth}
        \centering
        \includegraphics[width=\linewidth]{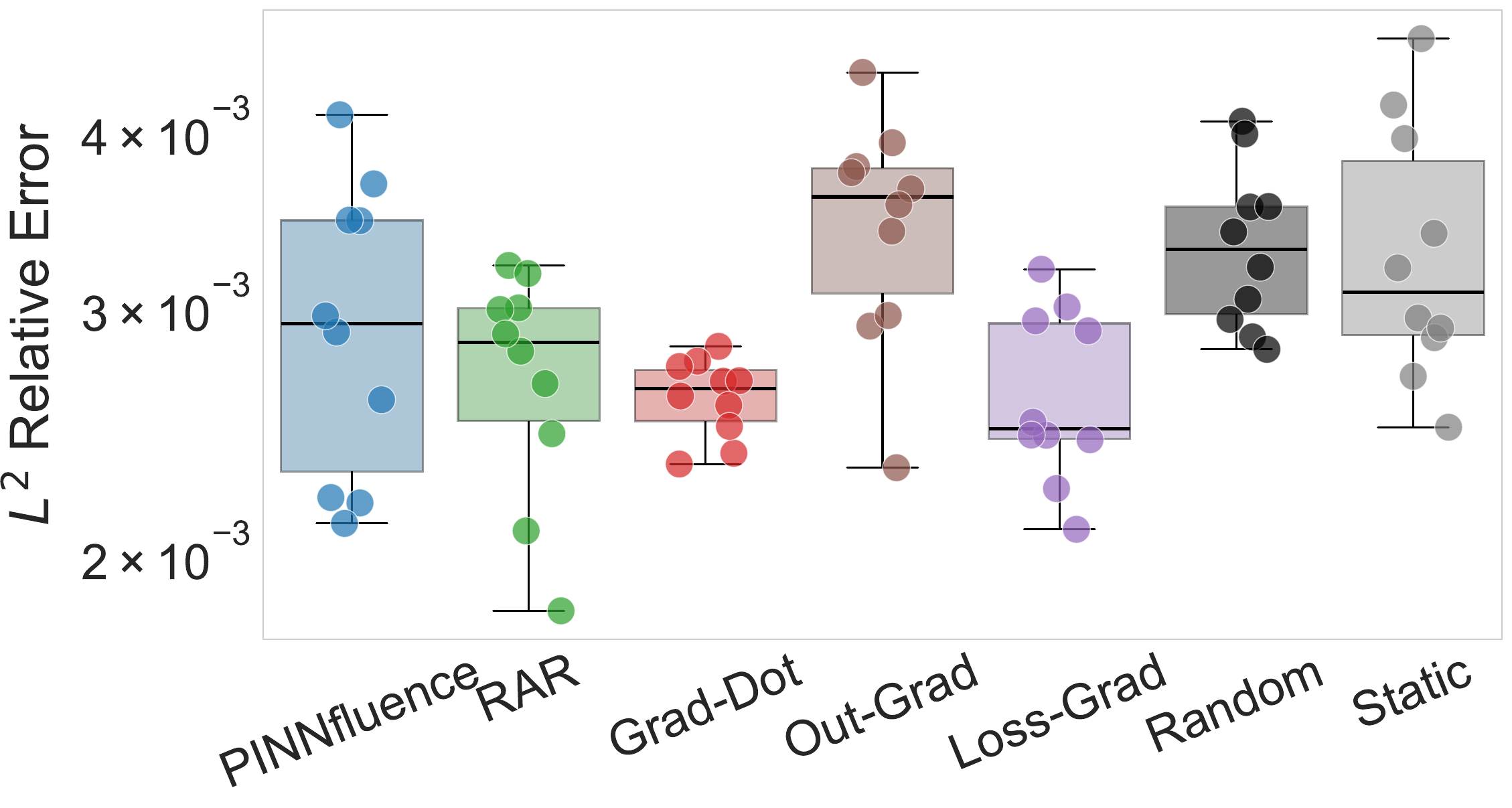}\mysubfigskip
        \caption{Drift-Diffusion (\emph{Adding})}
    \end{subfigure}
    \begin{subfigure}{\mysfigwidth}\mysubfigskip
        \centering
        \includegraphics[width=\linewidth]{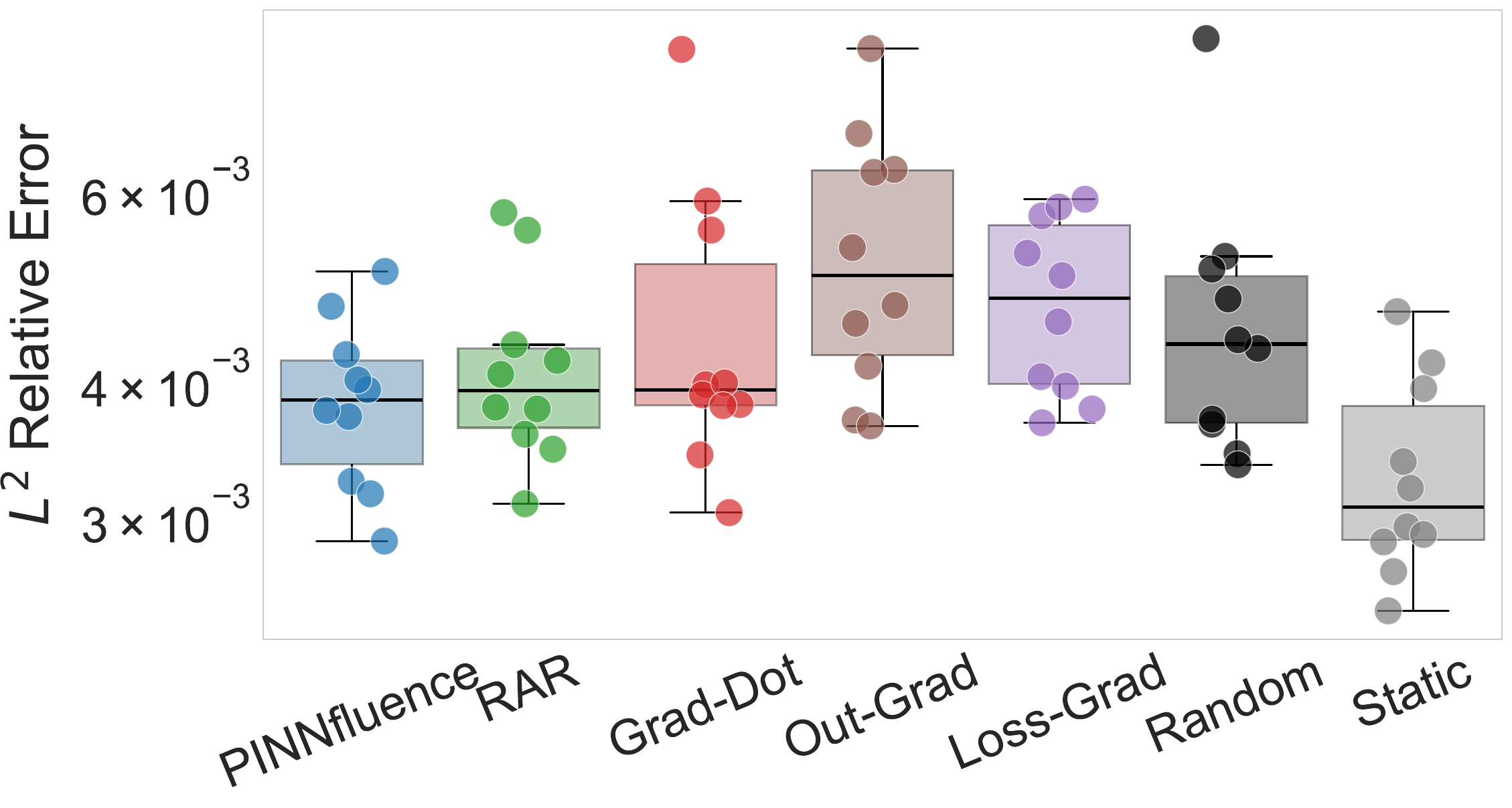}\mysubfigskip
        \caption{Drift-Diffusion (\emph{Replacing})}
    \end{subfigure}
    
    \caption{Boxplot comparing relative errors ($L^2$) across all sampling strategies. Each dot represents one run. The y-axis is log scaled.}
    \label{fig:boxplots_l2}
\end{figure}

\clearpage

\begin{figure}
    \centering

    \begin{subfigure}{\mysfigwidth}
        \centering
        \includegraphics[width=\linewidth]{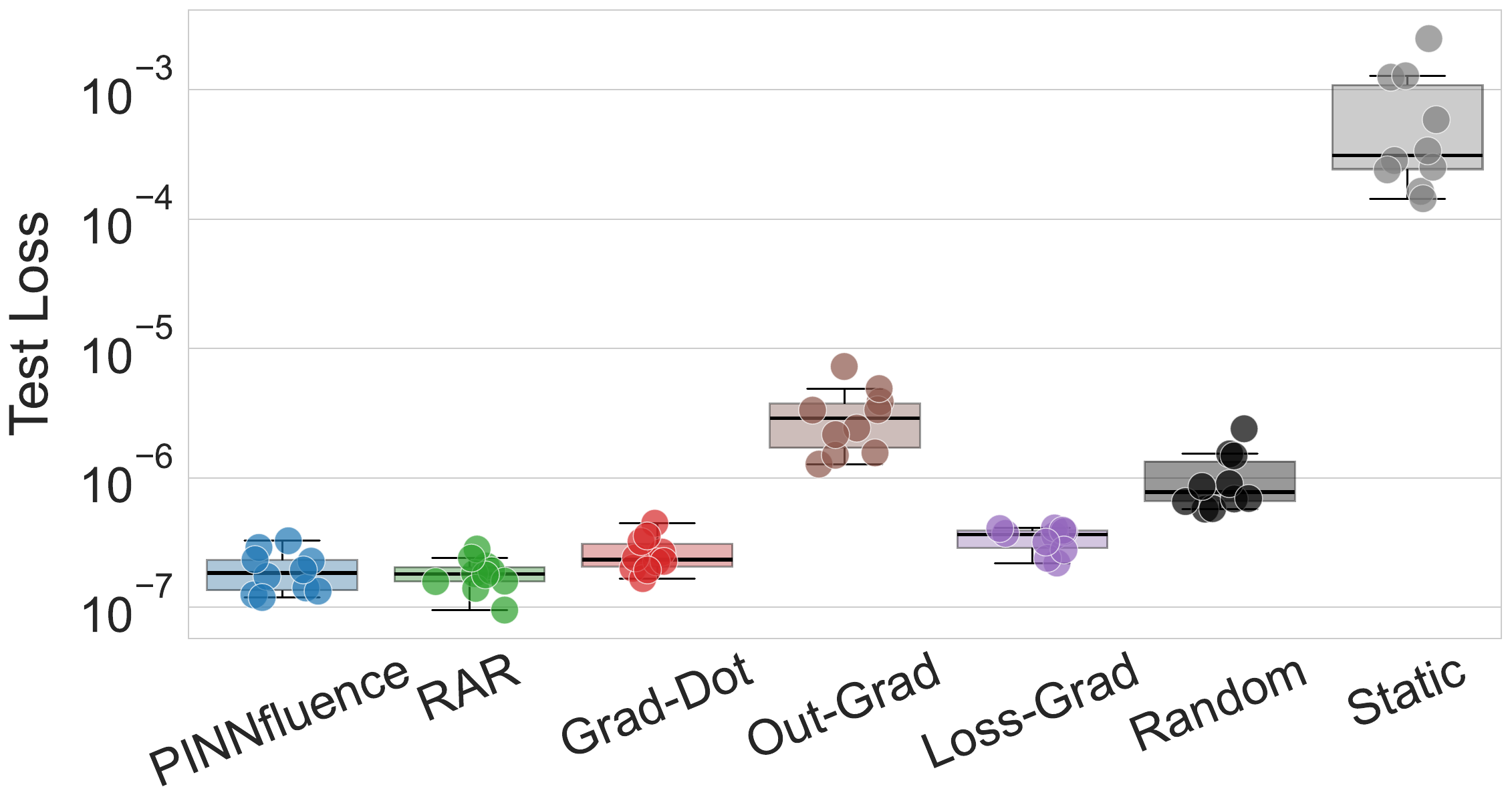}\mysubfigskip
        \caption{Diffusion (\emph{Adding})}
    \end{subfigure}
    \begin{subfigure}{\mysfigwidth}\mysubfigskip
        \centering
        \includegraphics[width=\linewidth]{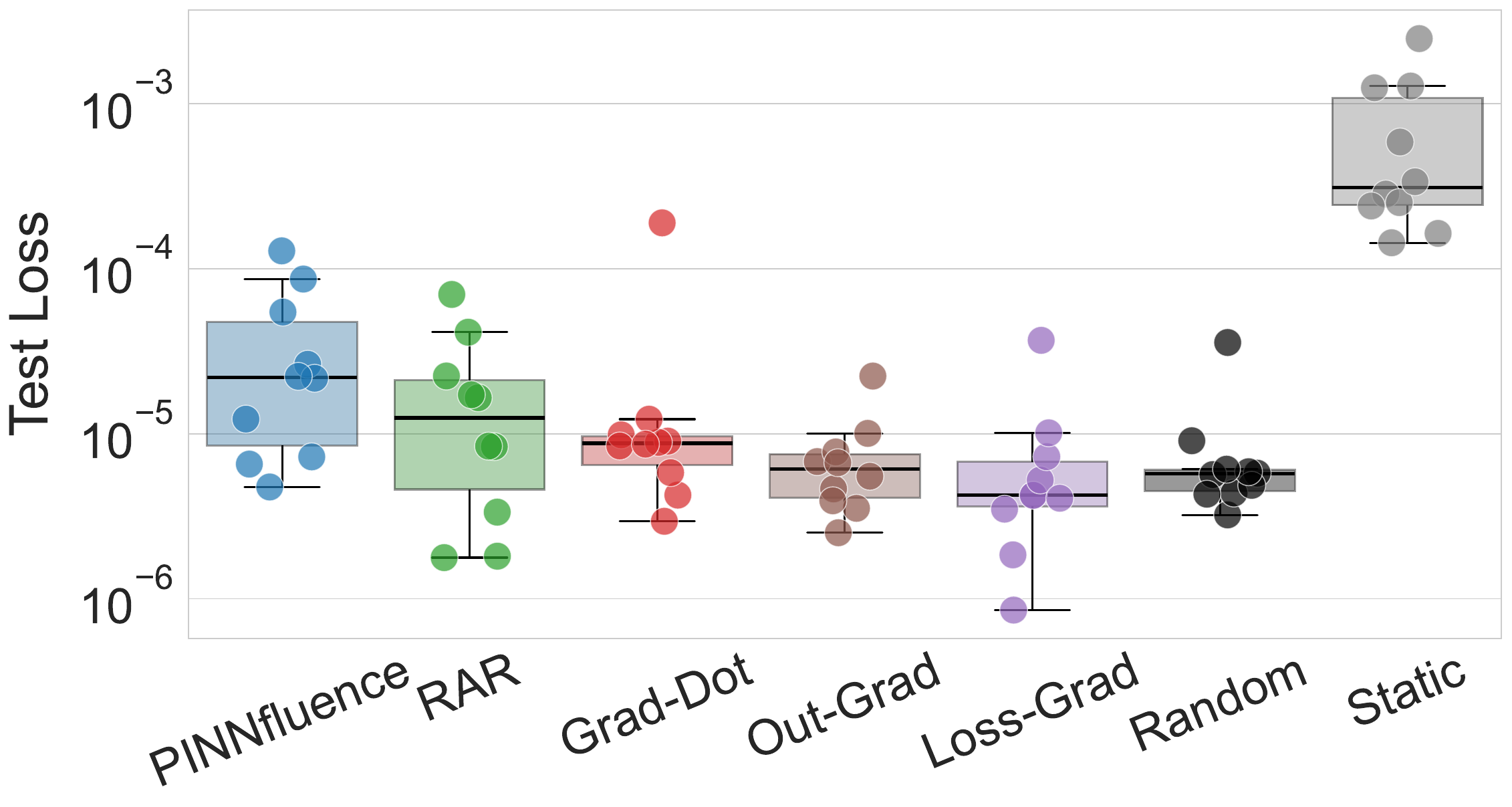}\mysubfigskip
        \caption{Diffusion (\emph{Replacing})}
    \end{subfigure}

    \begin{subfigure}{\mysfigwidth}
        \centering
        \includegraphics[width=\linewidth]{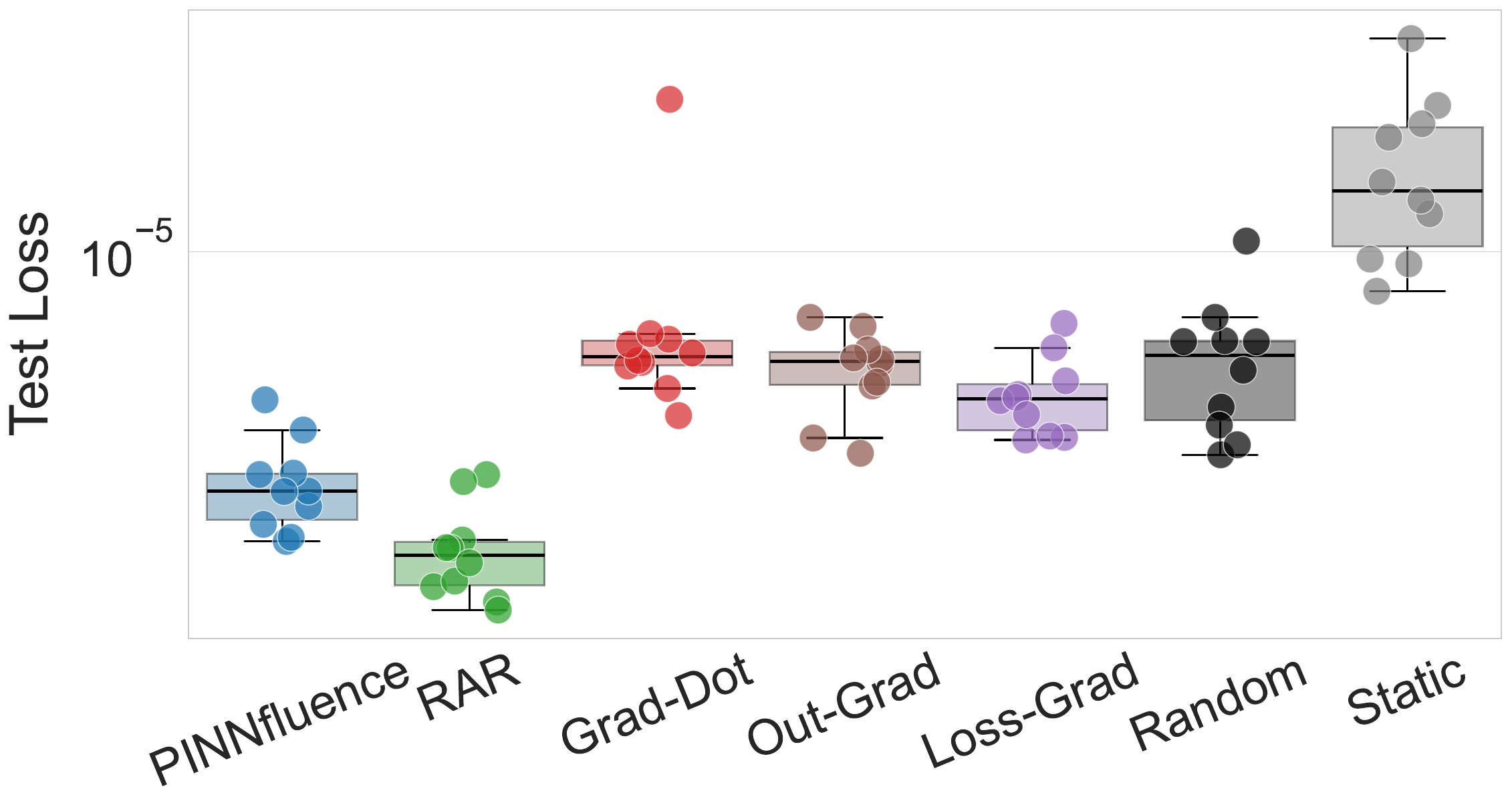}\mysubfigskip
        \caption{Allen-Cahn (\emph{Adding})}
    \end{subfigure}
    \begin{subfigure}{\mysfigwidth}\mysubfigskip
        \centering
        \includegraphics[width=\linewidth]{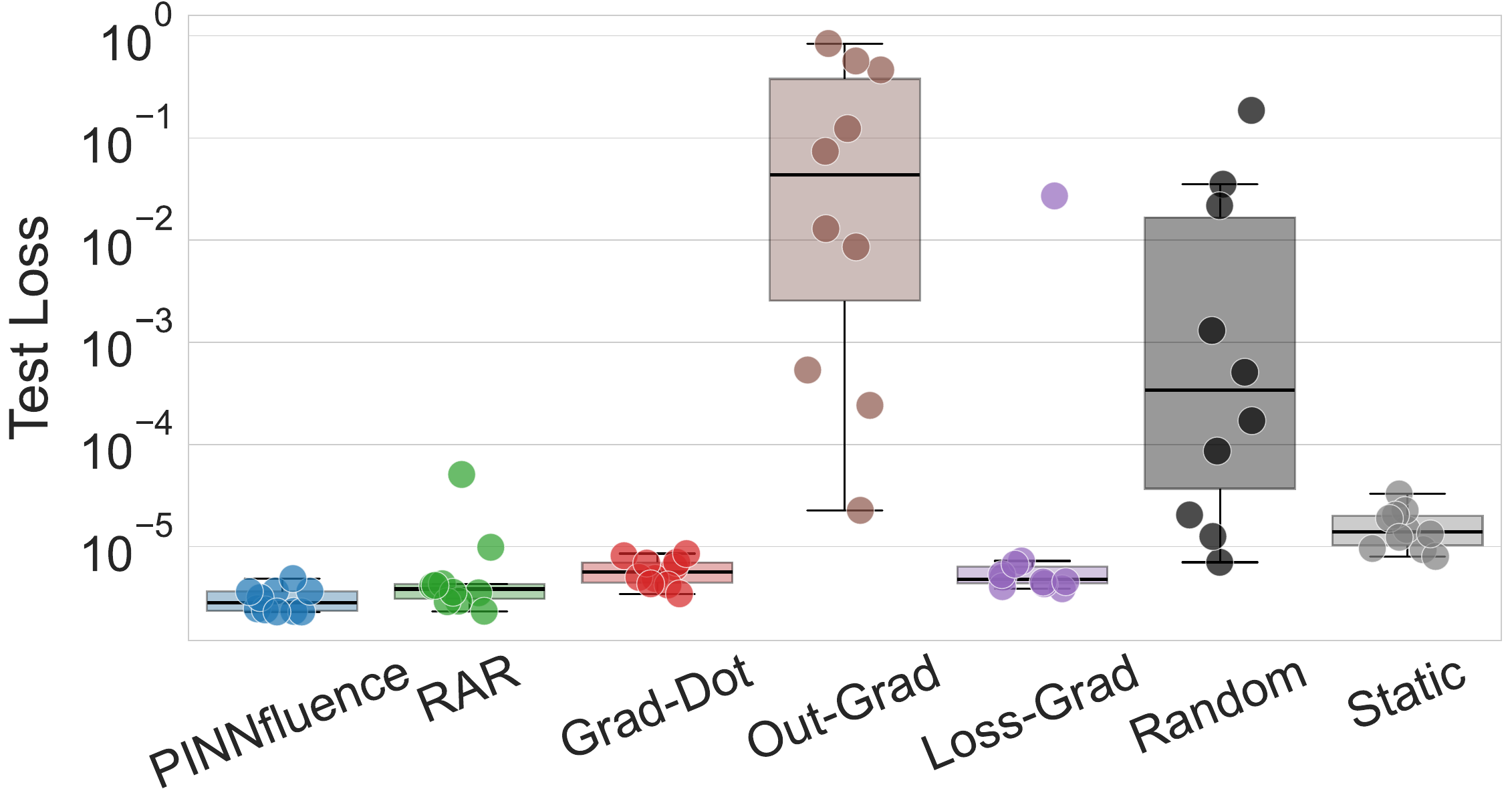}\mysubfigskip
        \caption{Allen-Cahn (\emph{Replacing})}
    \end{subfigure}

    \begin{subfigure}{\mysfigwidth}
        \centering
        \includegraphics[width=\linewidth]{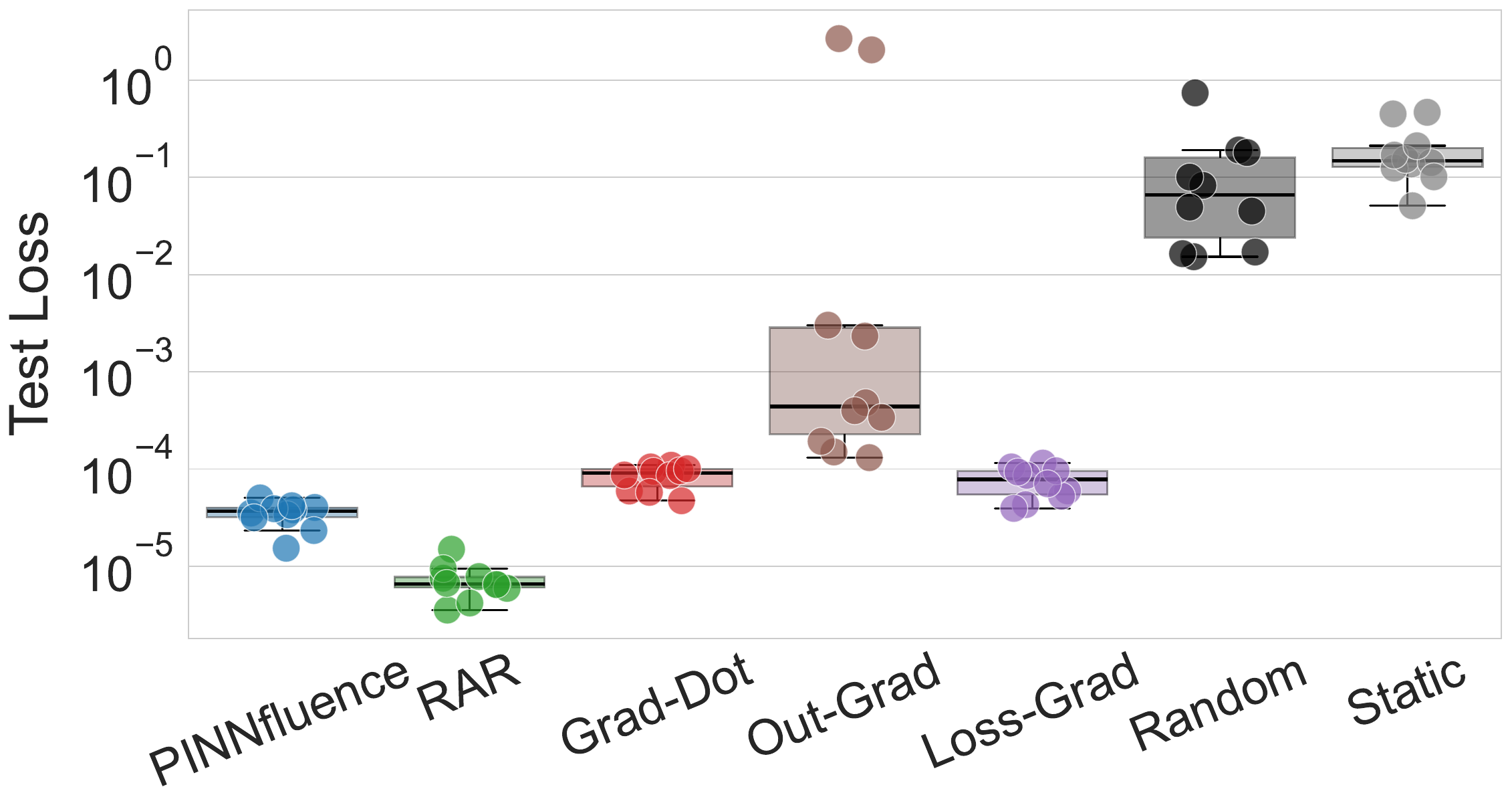}\mysubfigskip
        \caption{Burgers' (\emph{Adding})}
    \end{subfigure}
    \begin{subfigure}{\mysfigwidth}\mysubfigskip
        \centering
        \includegraphics[width=\linewidth]{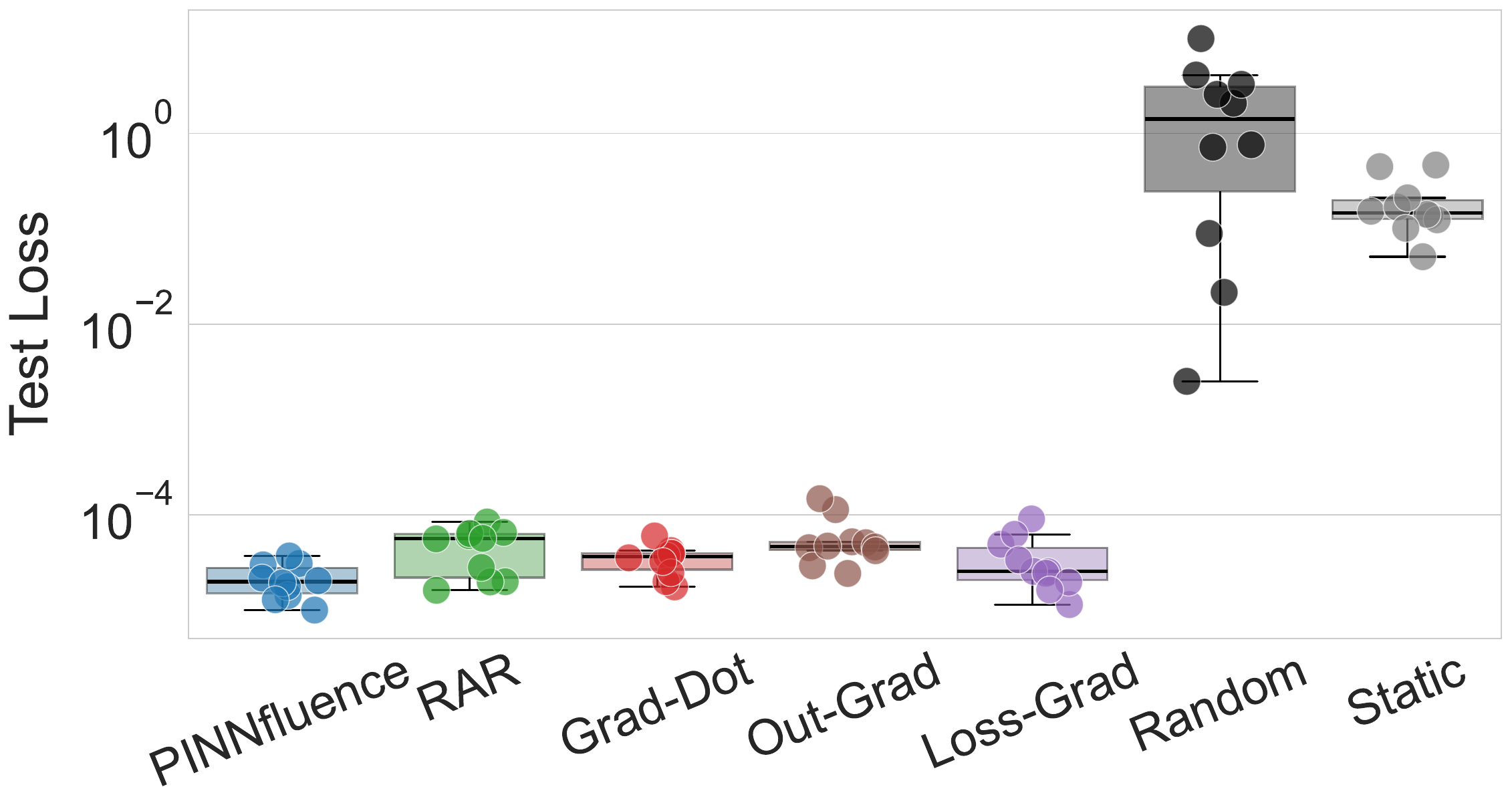}\mysubfigskip
        \caption{Burgers' (\emph{Replacing})}
    \end{subfigure}

    \begin{subfigure}{\mysfigwidth}
        \centering
        \includegraphics[width=\linewidth]{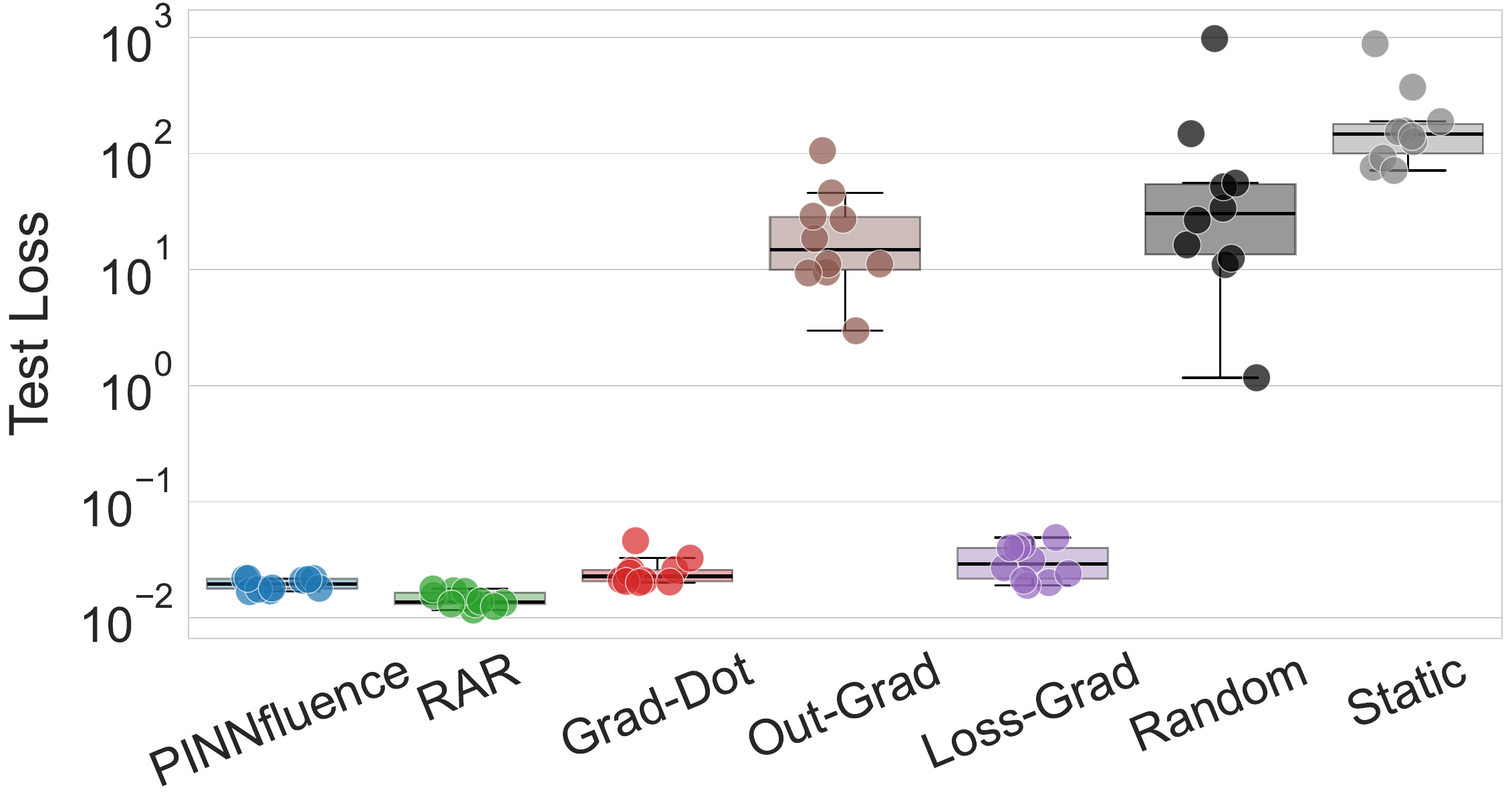}\mysubfigskip
        \caption{Wave (\emph{Adding})}
    \end{subfigure}
    \begin{subfigure}{\mysfigwidth}\mysubfigskip
        \centering
        \includegraphics[width=\linewidth]{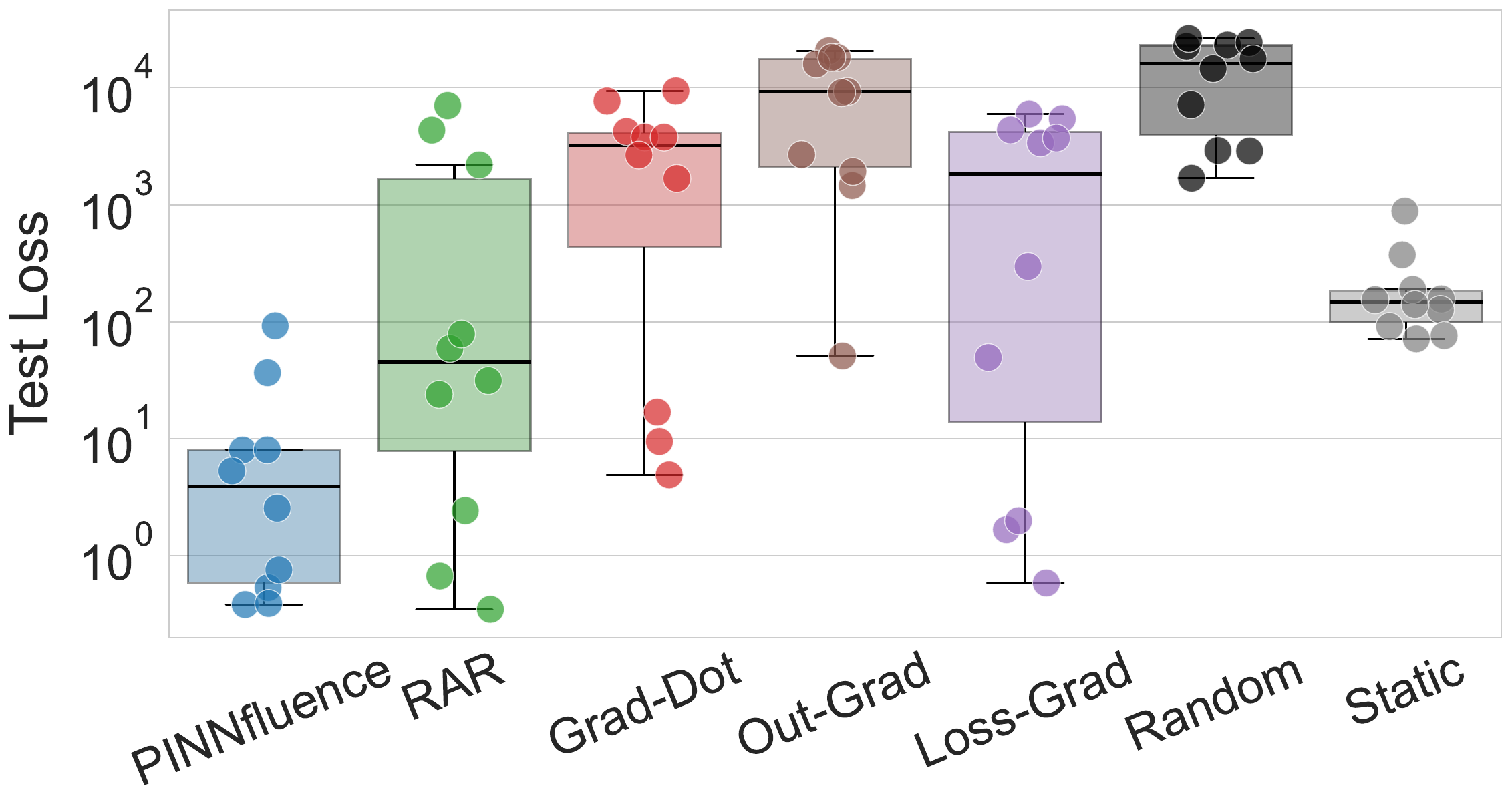}\mysubfigskip
        \caption{Wave (\emph{Replacing})}
    \end{subfigure}

    \begin{subfigure}{\mysfigwidth}
        \centering
        \includegraphics[width=\linewidth]{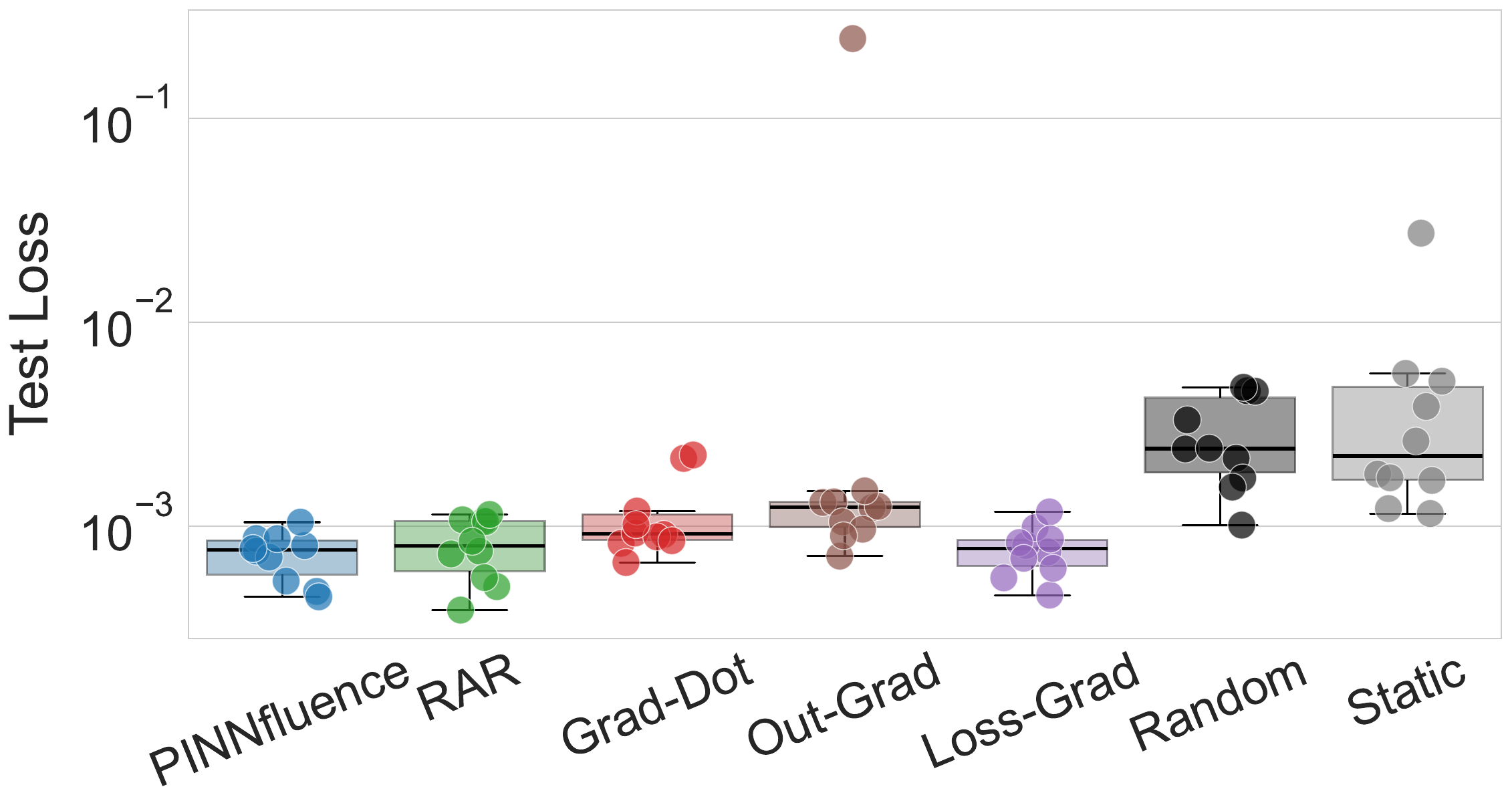}\mysubfigskip
        \caption{Drift-Diffusion (\emph{Adding})}
    \end{subfigure}
    \begin{subfigure}{\mysfigwidth}\mysubfigskip
        \centering
        \includegraphics[width=\linewidth]{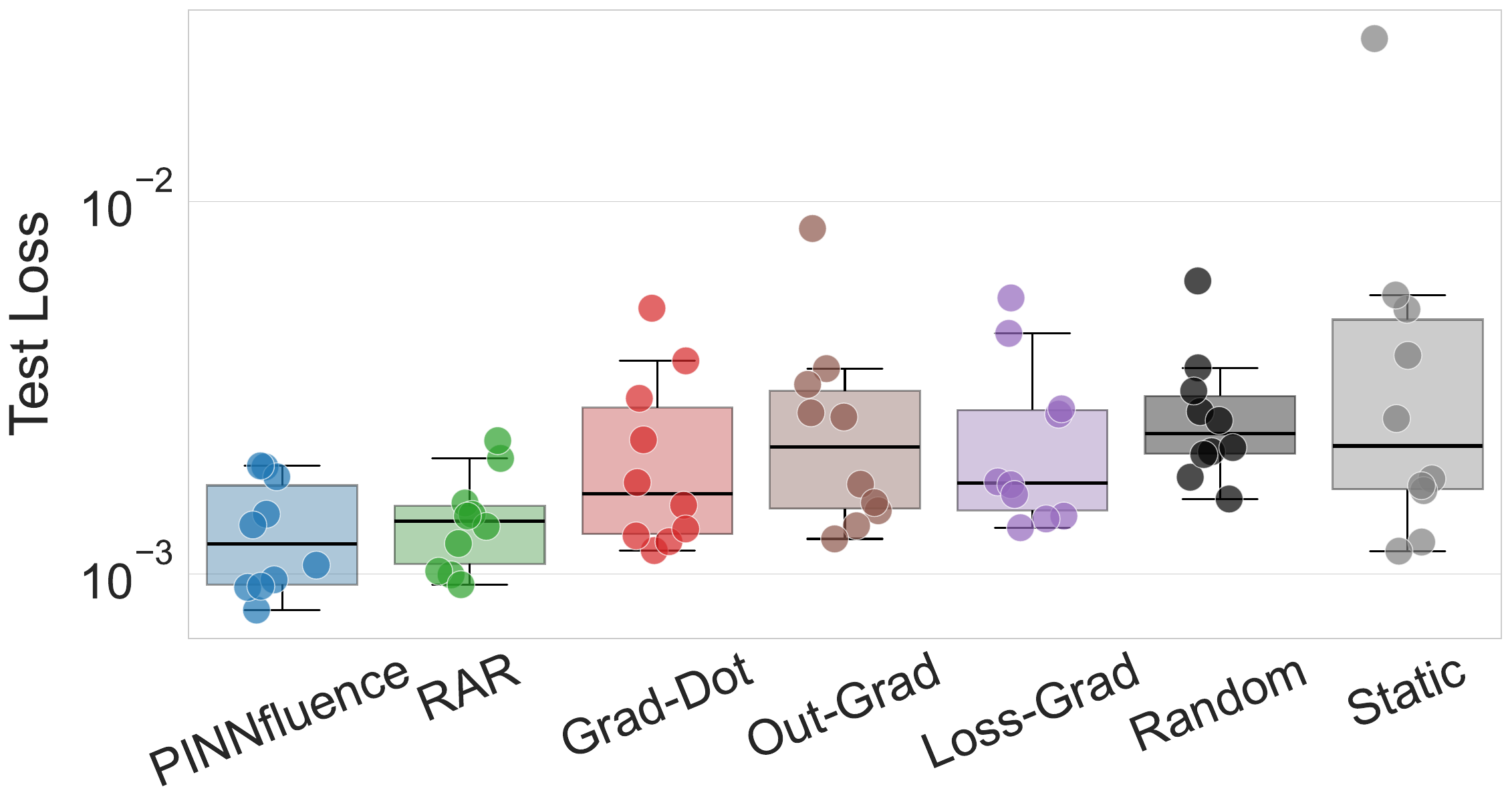}\mysubfigskip
        \caption{Drift-Diffusion (\emph{Replacing})}
    \end{subfigure}
    
    \caption{Boxplot comparing test losses across all sampling strategies. Each dot represents one run. The y-axis is log scaled.}
    \label{fig:boxplots_test_loss}
\end{figure}

\end{document}